\documentclass[10pt,twocolumn,letterpaper]{article}

\usepackage[pagenumbers]{cvpr} 

\usepackage[utf8]{inputenc} 
\usepackage[T1]{fontenc}    
\usepackage{hyperref}       
\usepackage{url}            
\usepackage{booktabs}       
\usepackage{amsfonts}       
\usepackage{nicefrac}       
\usepackage{microtype}      
\usepackage{amsmath}
\usepackage{graphicx}
\usepackage{caption}  
\usepackage{enumitem}
\usepackage{makecell}
\usepackage{graphicx}
\usepackage{float}
\usepackage{amsmath}
\usepackage{multirow}
\usepackage{colortbl}
\usepackage[dvipsnames]{xcolor} 
\usepackage{fontawesome}

\usepackage{algorithm}
\usepackage{algpseudocode} 
\usepackage{overpic}
\usepackage[utf8]{inputenc} 
\usepackage[T1]{fontenc}    
\usepackage{hyperref}       
\usepackage{url}            
\usepackage{booktabs}       
\usepackage{amsfonts}       
\usepackage{nicefrac}       
\usepackage{microtype}      
\usepackage{amsmath}
\usepackage{amssymb}
\usepackage{enumitem}
\usepackage{makecell}
\usepackage{tikz}  
\usetikzlibrary{shapes.geometric, arrows.meta, calc} 
\usepackage{subcaption}
\usepackage{multirow}
\usepackage{rotating}
\usepackage{comment}
\usepackage{wrapfig}
\usepackage{makecell}

\usepackage{tcolorbox}
\tcbuselibrary{skins}
\tcbuselibrary{breakable} 
\usepackage{lipsum} 
\definecolor{cvprblue}{rgb}{0.21,0.49,0.74}

\def\paperID{3517} 
\def\confName{CVPR}
\def\confYear{2026}

\title{InstantViR: Real-Time Video Inverse Problem Solver with \\Distilled Diffusion Prior}

\author{Weimin Bai$^{1*}$, \:Suzhe Xu$^{2*}$, \:Yiwei Ren$^1$, \:Jinhua Hao$^3$, \:Ming Sun$^3$, \:Wenzheng Chen$^1$, \:He Sun$^1$\textsuperscript{\dag} \\[2pt]
1. Peking University \quad
2. Huaqiao University \quad
3. Kuaishou Technology \\ [2pt]
    \vspace{0.05em}  
    \url{https://ai4scientificimaging.org/instantvir} 
}
\vspace{-0.15em}

\begin{document}
\maketitle

\begingroup
\renewcommand\thefootnote{$*$}
\footnotetext{These authors contributed equally to this work.}
\endgroup

\begingroup
\renewcommand\thefootnote{\dag}  
\footnotetext{Corresponding author.}
\endgroup

\begin{abstract}

\textcolor{black}{Video inverse problems are fundamental to streaming, telepresence, and AR/VR, where high perceptual quality must coexist with tight latency constraints. Diffusion-based priors currently deliver state-of-the-art reconstructions, but existing approaches either adapt image diffusion models with ad hoc temporal regularizers—leading to temporal artifacts—or rely on native video diffusion models whose iterative posterior sampling is far too slow for real-time use.}
\textcolor{black}{We introduce \textbf{InstantViR}, an amortized inference framework for ultra-fast video reconstruction powered by a pre-trained video diffusion prior. We distill a powerful bidirectional video diffusion model (teacher) into a causal autoregressive student that maps a degraded video directly to its restored version in a single forward pass, inheriting the teacher’s strong temporal modeling while completely removing iterative test-time optimization. The distillation is prior-driven: it only requires the teacher diffusion model and known degradation operators, and does not rely on externally paired clean/noisy video data. To further boost throughput, we replace the video-diffusion backbone VAE with a high-efficiency LeanVAE via an innovative teacher-space regularized distillation scheme, enabling low-latency latent-space processing.}
\textcolor{black}{Across streaming random inpainting, Gaussian deblurring and super-resolution, InstantViR matches or surpasses the reconstruction quality of diffusion-based baselines while running at over \textbf{35 FPS} on NVIDIA A100 GPUs, achieving up to 100$\times$ speedups over iterative video diffusion solvers. These results show that diffusion-based video reconstruction is compatible with real-time, interactive, editable, streaming scenarios, turning high-quality video restoration into a practical component of modern vision systems.}
\end{abstract}    
\vspace{-0.5em}
\section{Introduction}
\label{sec:introduction}
Videos have become the dominant medium for communication, entertainment, and information in the digital age.
Consequently, the task of solving video inverse problems—reconstructing a high-quality, clean video $\boldsymbol{x}$ from degraded measurements $\boldsymbol{y}$—is of pivotal importance, with applications ranging from enhancing consumer footage and restoring historical films to improving medical imaging and autonomous driving systems~\cite{chang2025warped, daras2024warped, kwon2025solving, kwon2024visionxlhighdefinitionvideo, zhang2025step, zou2025flair}.
This reconstruction task is inherently ill-posed, as a single measurement $\boldsymbol{y}$ can correspond to many possible clean videos $\boldsymbol{x}$.
To solve this ambiguity, a standard approach is to cast video reconstruction as a Bayesian inverse problem, where the goal is to sample from the posterior distribution $p(\boldsymbol{x}|\boldsymbol{y})$:
\begin{equation}
p(\boldsymbol{x}|\boldsymbol{y}) \propto p(\boldsymbol{y}|\boldsymbol{x})p(\boldsymbol{x}).
\label{equ:baye}
\end{equation}
This formulation elegantly decouples the problem into two components: (1) the likelihood $p(\boldsymbol{y}|\boldsymbol{x})$, governed by the physics of the degradation (e.g., blur, noise, or masking); and (2) the prior $p(\boldsymbol{x})$, which captures the high-dimensional spatiotemporal statistics of natural videos. The central challenge in video inverse problems is therefore how to represent and compute with this video prior $p(\boldsymbol{x})$ in a way that is both expressive and computationally tractable.

\begin{figure*}[tbp]
\centering
\includegraphics[width=0.99\textwidth]{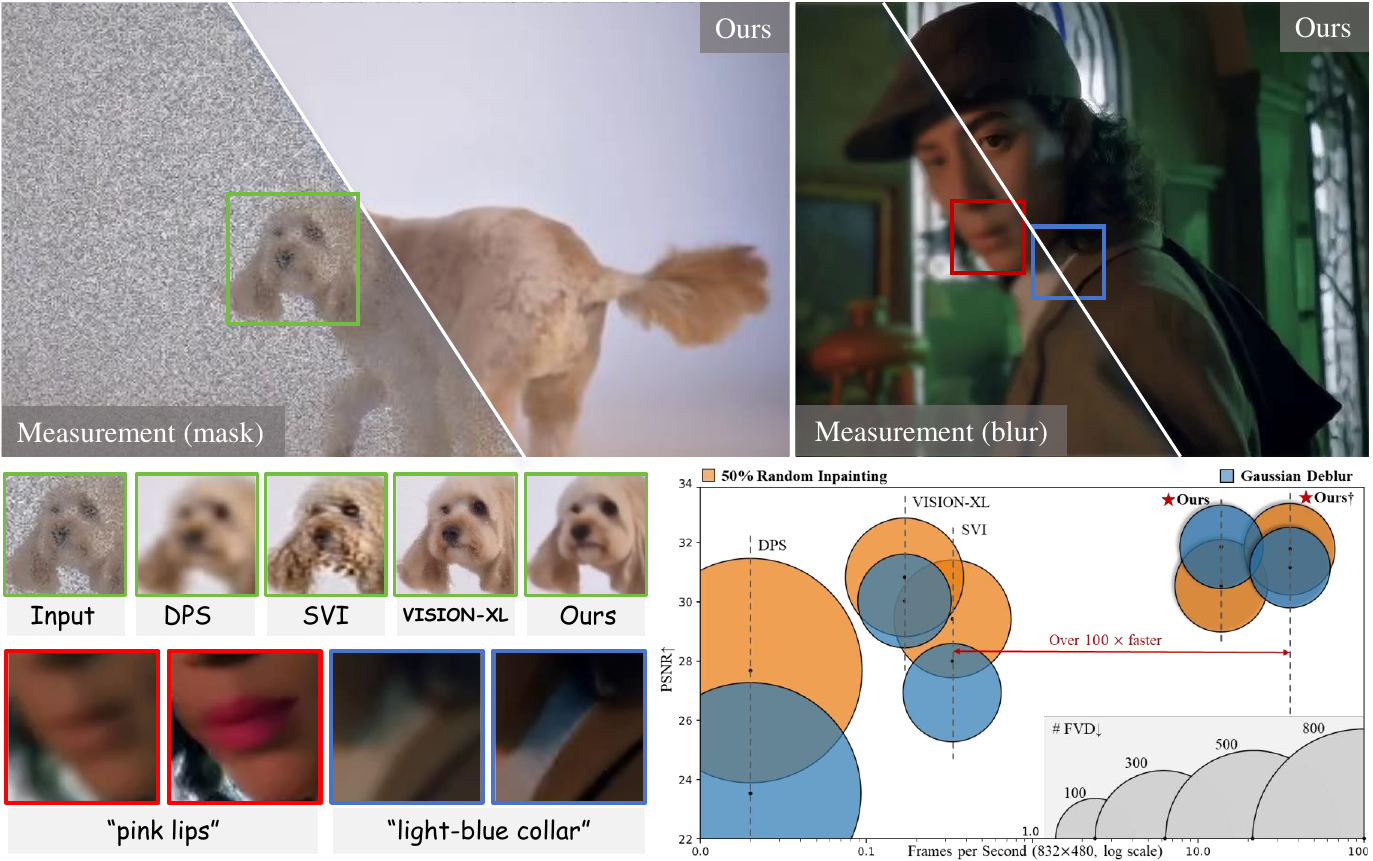}
\vspace{-0.2em}
\caption{We introduce InstantViR, a real-time video inverse problem solver that drastically outperforms slow sampling-based methods in both speed and quality. \textbf{Bottom-right}: At 832$\times$480 resolution, our amortized framework is over \textbf{100}$\times$ faster than sampling-based baselines like SVI~\cite{kwon2025solving}, achieving over \textbf{35} FPS and the excellent quality. \textbf{Left and Bottom-left}: Qualitative examples demonstrate versatile, high-fidelity reconstruction for inpainting and deblurring, along with optional text-guided control (e.g., "pink lips", "light-blue collar").}
\label{fig:teaser}
\vspace{-0.8em}
\end{figure*}

Recently, diffusion models~\cite{ho2020denoising, song2020score, sohl2015deep} have become the dominant tool for learning powerful priors, yielding state-of-the-art results for image inverse problems~\cite{chung2022diffusion, mardani2023variational, rout2023solving, bai2024blind, lee2024diffusion}. A natural first step for videos is to repurpose pre-trained image diffusion models~\cite{ho2020denoising,rombach2022high} for video restoration. 
Methods along this line~\cite{daras2024warped, kwon2025solving, kwon2024visionxlhighdefinitionvideo, zhang2025step} process videos frame-by-frame and enforce consistency only through empirical temporal regularizers (e.g., optical flow constraints or batched noise). While convenient, this image-native strategy has two key drawbacks: (1) it uses priors that ignore rich spatiotemporal dynamics, often leading to flicker and temporal inconsistency; and (2) it still relies on slow, iterative sampling to solve Eq.~\ref{equ:baye}, which is ill-suited for real-time deployment.

A more principled alternative is to adopt native video diffusion models~\cite{wan2.1, wan2025}, which are trained directly on videos and provide much stronger temporal priors, producing visually coherent sequences. 
However, when used for inverse problems, these models inherit and amplify the computational burden of diffusion: posterior sampling~\cite{chung2022diffusion} in Eq.~\ref{equ:baye} requires long iterative trajectories in a much higher-dimensional video space, making inference even more expensive. 
As a result, current diffusion-based approaches face a stark trade-off: either use image diffusion priors that are temporally weak and slow, or video diffusion priors that are temporally strong but prohibitively slower and incompatible with real-time, streaming applications.

In this work, we argue that this trade-off is not fundamental. We show that one can retain the rich, temporally consistent prior of a strong video diffusion model without inheriting its prohibitive sampling cost. 
To this end, we introduce InstantViR, a variational framework that recasts video reconstruction as amortized inference via video diffusion distillation~\cite{luo2023diff,huang2024flow,wang2024integrating,yin2024one,yin2024slow,zhou2024score, lee2024diffusion}. 
Instead of solving a separate, slow optimization for $p(\boldsymbol{x}|\boldsymbol{y})$ at test time, InstantViR learns a universal one-step solver that maps a degraded video $\boldsymbol{y}$ directly to its reconstruction $\boldsymbol{x}$. 

Our method uses an asymmetric teacher–student design. A slow bidirectional video diffusion model (e.g., Wan2.1~\cite{wan2.1}) serves as the teacher, which together with the measurement operator (blur, masking, downsampling, text-guided edits, etc.) defines the target posterior. 
A fast causal autoregressive student is then trained to approximate this posterior in a single forward pass. 
Training is paired-data-free: conditioned on the degraded measurement $\boldsymbol{y}$, we align the reconstruction with the teacher's prior while enforcing physics-based data fidelity, yielding a self-supervised approximation to the video posterior.
As a result, the same framework seamlessly adapts to streaming random inpainting, Gaussian deblurring, super-resolution, and language-guided restoration/editing by simply changing the measurement model and the conditioning prompt of the video diffusion prior.

To make the system truly real-time, we also tackle the often-overlooked VAE bottleneck. We introduce a teacher-space regularized latent adaptation scheme that allows us to replace the original video VAE with LeanVAE~\cite{cheng2025leanvae}—an ultra-efficient spatiotemporal tokenizer with a lightweight Neighborhood-Aware Feedforward backbone and wavelet-based channel compression—while keeping the distilled solver consistent with the teacher prior. Combined with our one-step causal solver, this design allows InstantViR to deliver diffusion-level reconstructions with up to 100× speedup over iterative video diffusion priors, while maintaining high fidelity and temporal coherence.

Concretely, InstantViR contributes: 
(i) a paired-data-free variational distillation framework that converts native video diffusion priors into one-step amortized solvers for diverse video inverse problems; 
(ii) a streaming causal inverse architecture with block-wise attention and KV caching for low-latency, high-fidelity reconstruction; 
and (iii) a teacher-space regularized LeanVAE integration that overcomes latent distribution shift and, to our knowledge, yields the first real-time, interactive pipeline for streaming video reconstruction with diffusion-level quality at over 35 FPS.

\section{Related Works}
\label{sec:relatedworks}


\subsection{Video Diffusion Models as Priors}
\label{sec:video_diffusion_priors}


Diffusion models~\cite{ho2020denoising, song2020score, sohl2015deep} have emerged as the state-of-the-art to serve as powerful priors $p(\boldsymbol{x})$ that can accurately model the complex statistics of natural videos. 
These models define the prior $p(\boldsymbol{x})$ through a learned reverse-time stochastic differential equation (SDE) that denoises a simple Gaussian distribution $p(\boldsymbol{x}_T)$ back to the data distribution $p(\boldsymbol{x}_0)$. 
Concretely, for a clean video $\boldsymbol{x}_0$, a fixed forward process defines how noise is added:
\begin{equation}
d\boldsymbol{x}_t = f(t)\boldsymbol{x}_t\,dt + g(t)\,d\mathbf{w}_t,
\label{eq:forward_sde}
\end{equation}
where $t \in [0, T]$, $f(t)$ and $g(t)$ are drift and diffusion coefficients, and $\mathbf{w}_t$ is a standard Wiener process. 
The generative prior $p(\boldsymbol{x})$ is defined by the corresponding reverse-time SDE:
\begin{equation}
d\boldsymbol{x}_t = \bigl[f(t)\boldsymbol{x}_t - g(t)^2 \nabla_{\boldsymbol{x}_t}\log p_t(\boldsymbol{x}_t)\bigr]\,dt + g(t)\,d\bar{\mathbf{w}}_t,
\label{eq:reverse_sde}
\end{equation}
where the key component, the score function $\nabla_{\boldsymbol{x}_t}\log p_t(\boldsymbol{x}_t)$, is approximated by a neural network $s_\theta(\boldsymbol{x}_t, t)$. This network $s_\theta$ is trained by optimizing the denoising score matching (DSM) objective~\cite{song2019generative} with $\boldsymbol{x}_t = \alpha_t \boldsymbol{x}_0 + \sigma_t \boldsymbol{\epsilon}$:
\begin{equation}
\mathcal{L}_{\mathrm{DSM}} = \mathbb{E}_{t, \boldsymbol{x}_0, \boldsymbol{\epsilon}}\!\Bigl[\lambda(t)\,\bigl\|s_\theta(\boldsymbol{x}_t, t) + \tfrac{\boldsymbol{\epsilon}}{\sigma_t}\bigr\|^2\Bigr].
\label{eq:dsm_loss}
\end{equation}

Recently, powerful video generative priors based on Diffusion Transformers (DiTs)~\cite{peebles2023scalable}, including Open-Sora~\cite{zheng2024open}, Wan2.1~\cite{wan2.1, wan2025}, MovieGen~\cite{polyak2024movie} have demonstrated unprecedented ability to capture complex spatiotemporal dynamics, paving the way toward world simulators~\cite{brooks2024video}.
These models achieve exceptional temporal coherence by employing {bidirectional attention} across all video frames. 
However, this architectural strength is also a critical bottleneck: the generation of a single frame requires processing the entire sequence, including future frames. This non-causal dependency introduces prohibitive latency and makes them fundamentally incompatible with real-time or streaming applications.

\subsection{Diffusion Models for Inverse Problems}
\label{sec:diffusion_for_inverse_problems}

Using the diffusion priors defined above to solve inverse problems (Eq.~\ref{equ:baye}) requires combining the pre-trained prior score $s_\theta$ with the likelihood $p(\boldsymbol{y}|\boldsymbol{x})$. While this Bayesian framework has achieved great success in static image inverse problems, the field is now moving towards the significantly more challenging video domain, which introduces the critical demand for spatiotemporal consistency.

\vspace{-0.5em}
\paragraph{Diffusion Models for Image Inverse Problems.}
In the image domain, sampling-based methods~\cite{chung2022diffusion, rout2023solving, song2023solving} work by injecting an approximation of the likelihood gradient, $\nabla_{\mathbf{x}} \log p(\boldsymbol{y}|\boldsymbol{x})$, into the reverse sampling process (Eq.~\ref{eq:reverse_sde}). 
This is often achieved by approximating the intractable likelihood term $\nabla_{\mathbf{x}_t} \log p(y|\mathbf{x}_t)$ using the posterior mean $\hat{\mathbf{x}}_0$. 
Particle-based optimization methods~\cite{mardani2023variational, zilberstein2024repulsive} avoid this approximation via variational inference.
While effective for producing high-fidelity images, these methods are inherently slow, requiring hundreds or thousands of function evaluations to generate a single result.

\paragraph{Diffusion Models for Video Inverse Problems.}
Extending this framework to video inverse problems introduces a significant new challenge: maintaining temporal consistency. 
The field has primiarily focused on adapting the {image-based solvers} by adding temporal heuristics. 
For example, Warped Diffusion~\cite{daras2024warped} uses optical flow to warp noise, SVI~\cite{kwon2025solving} and Vision-XL~\cite{kwon2024visionxlhighdefinitionvideo} treat time as a batch dimension, while STEP~\cite{zhang2025step} tuned image diffusion model to capture temporal consistency.
These methods are constrained by a weak image-native prior that struggles with complex dynamics and remain prohibitively slow.
To address this speed bottleneck, one-step distillation methods~\cite{wang2025seedvr2, zhuang2025flashvsr} have been proposed for specific tasks like super-resolution. However, they are highly task-specific and relies on massive, fully-paired $(\boldsymbol{x}, \boldsymbol{y})$ datasets (e.g., 10M video pairs), limiting its generality and scalability.

This leaves a clear research gap: existing methods force a trade-off between weak-but-slow image priors, or fast-but-task-specific models that require massive paired data. 
This motivates our work, {InstantViR}, which introduces an amortized paradigm to distill powerful, general-purpose video priors ({e.g.}, Wan2.1~\cite{wan2.1}) into a one-step, feed-forward solver that operates {without paired data}, enabling real-time, high-fidelity reconstruction for the first time.

\begin{figure*}[t]
    \centering
    \includegraphics[width=0.999\textwidth]{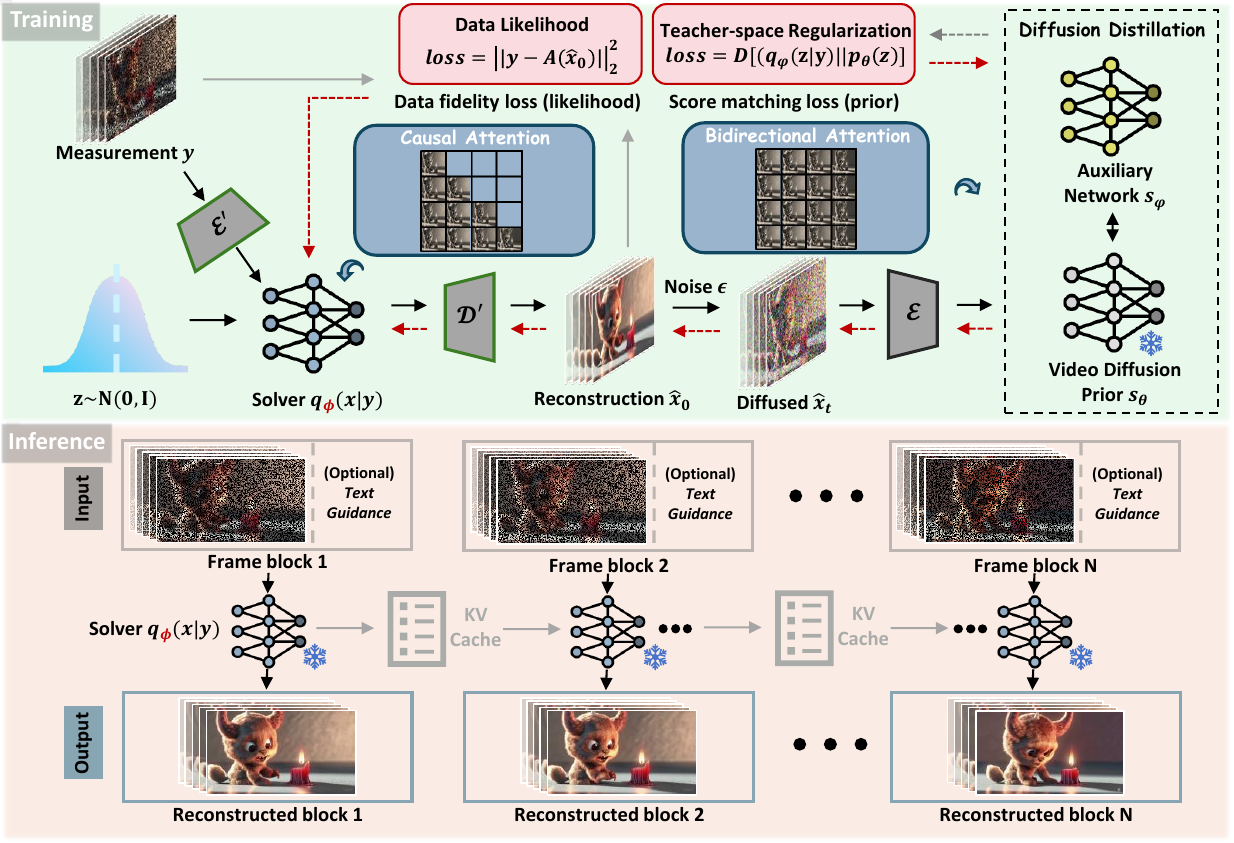}
    \caption{\textbf{Overview of the InstantViR framework.}
\textbf{(Top) Training:} We train a single-step solver $q_\phi$ using only degraded measurements $\boldsymbol{y}$. The solver is optimized with two objectives: a data fidelity loss (ensuring the reconstruction matches the measurement $\boldsymbol{y}$) and a prior distillation loss (using a frozen video diffusion prior~\cite{wan2.1}, $s_\theta$, to enforce temporal consistency and realism).
\textbf{(Bottom) Inference:} The trained solver $q_\phi$ operates as a fast, feed-forward network, processing the video in a causal, block-wise, and autoregressive manner. This enables real-time, single-step streaming reconstruction with optional text guidance.}
    \label{fig:overview}
    \vspace{-1em}
\end{figure*}

\section{Method}
\label{sec:method}

We now describe {InstantViR}, which transforms a slow, iterative video diffusion prior into a fast, single-step solver for video inverse problems.
In Sec.~\ref{sec:method:framework}, we introduce our variational amortized inference formulation and derive an objective that enables training without paired data.
In Sec.~\ref{sec:method:inference}, we present our causal video inverse solver architecture, which enables efficient block-wise streaming inference.
In Sec.~\ref{sec:method:acceleration}, we show how to replace the original video VAE with an ultra-efficient LeanVAE while carefully handling the induced latent distribution shift, which is crucial for achieving true real-time performance.

\subsection{Amortized Inference Framework}
\label{sec:method:framework}
Modern video diffusion models typically operate in a latent space.
A video $\boldsymbol{x}$ is first encoded into latent codes $\boldsymbol{z} = \mathcal{E}(\boldsymbol{x})$ by a video VAE, and then a Diffusion Transformer (DiT) backbone~\cite{peebles2023scalable} learns a generative prior $p(\boldsymbol{z})$ over these latents.
At sampling time, frames are decoded via $\boldsymbol{x} = \mathcal{D}(\boldsymbol{z})$, yielding high-fidelity, temporally coherent videos. 
Our goal is to leverage such a pre-trained video diffusion model as a prior to solve video inverse problems. Given degraded measurements $\boldsymbol{y}$ related to the clean video $\boldsymbol{x}$ via a known forward operator $\boldsymbol{y} = \mathcal{A}(\boldsymbol{x})$, we wish to sample from the posterior $p(\boldsymbol{x} | \boldsymbol{y})$ defined in Eq.~\ref{equ:baye}.

Traditional sampling-based approaches~\cite{daras2024warped, kwon2025solving, kwon2024visionxlhighdefinitionvideo} inject the likelihood $p(\boldsymbol{y} | \boldsymbol{x})$ (defined by $\mathcal{A}$) into the diffusion process and iteratively perturb the generation trajectory to enforce measurement consistency.
This yields high quality but is extremely slow, and repeatedly backpropagating through the decoder $\mathcal{D}$ adds further overhead.

Our amortized inference approach, instead, directly learns a single-step feed-forward solver, $q_\phi(\boldsymbol{x} | \boldsymbol{y})$, as a variational approximation to the posterior $p(\boldsymbol{x} | \boldsymbol{y})$.
We minimize the expected KL divergence
\begin{equation}
\label{eq:kl_objective}
\mathcal{L} = \mathbb{E}_{\boldsymbol{y} \sim p(\boldsymbol{y})} \Big[
D_{\mathrm{KL}}\!\left( q_\phi(\boldsymbol{x} | \boldsymbol{y}) \,\|\, p(\boldsymbol{x} | \boldsymbol{y}) \right)
\Big],
\end{equation}
which, up to an additive constant, can be decomposed as
\begin{equation}
\begin{split}
\mathbb{E}_{\boldsymbol{y}} \Bigl\{ \mathbb{E}_{\boldsymbol{x} \sim q_\phi(\boldsymbol{x}|\boldsymbol{y})}
\Big[
- \log p(\boldsymbol{y} | \boldsymbol{x})
\Big]
+ 
D_{\mathrm{KL}}\!\left(
q_\phi(\boldsymbol{x} | \boldsymbol{y})
\,\|\, p(\boldsymbol{x})
\right)\Bigl\},
\end{split}
\end{equation}
where the first term enforces data fidelity through the known degradation model, and the second term regularizes the solver to remain on the natural video manifold defined by the diffusion prior $p(\boldsymbol{x})$ (e.g., Wan2.1~\cite{wan2.1}).

Because the diffusion prior is defined in the latent space, we work with $\boldsymbol{z}$ instead of $\boldsymbol{x}$, where $\boldsymbol{x} = \mathcal{D}(\boldsymbol{z})$ and $\boldsymbol{z} \sim p(\boldsymbol{z})$.
We parameterize the solver as $q_\phi(\boldsymbol{z} | \boldsymbol{y})$ and obtain the latent-space objective
\begin{equation}
\label{eq:kl_objective_latent}
\mathbb{E}_{\boldsymbol{y}} \bigl\{
\underbrace{\mathbb{E}_{\boldsymbol{z} \sim q_\phi(\boldsymbol{z}|\boldsymbol{y})} \left[ - \log p\big(\boldsymbol{y} | \mathcal{D}(\boldsymbol{z})\big) \right]}_{\mathcal{L}_{\text{likelihood}}}
+
\underbrace{D_{\mathrm{KL}}\!\left(
q_\phi(\boldsymbol{z} | \boldsymbol{y}) \,\|\, p(\boldsymbol{z})
\right)}_{\mathcal{L}_{\text{prior}}} \bigl\}.
\end{equation}
The prior term $\mathcal{L}_{\text{prior}}$ involves a KL divergence to the diffusion prior $p(\boldsymbol{z})$, which is only implicitly defined via its score function $s_\theta(\boldsymbol{z}_t, t)$.
Following score-distillation methods~\cite{luo2023diff, yin2024one, lee2024diffusion}, we approximate this KL by a score-matching loss:
\begin{equation}
\label{eq:prior_score_loss}
\mathcal{L}_{\text{prior}}
\approx
\mathbb{E}_{t, \boldsymbol{\epsilon}, \boldsymbol{z} \sim q_\phi(\cdot|\boldsymbol{y})}
\big[
w(t)\,
\| s_\theta(\boldsymbol{z}_t, t) - s_{q_\phi}(\boldsymbol{z}_t, t) \|^2
\big],
\end{equation}
where $\boldsymbol{z}_t = \alpha_t \boldsymbol{z} + \sigma_t \boldsymbol{\epsilon}$ is the noised latent at time $t$, $w(t)$ is a weighting schedule, and $s_{q_\phi}$ denotes the score implied by the solver (implemented via a small auxiliary head $s_\varphi$).

Optimizing the combined objective in Eq.~\ref{eq:kl_objective_latent} trains $q_\phi(\boldsymbol{z} | \boldsymbol{y})$ to provide an amortized approximation of the true posterior $p(\boldsymbol{z}|\boldsymbol{y})$, and thus $p(\boldsymbol{x}|\boldsymbol{y})$ via $\mathcal{D}$.
This training requires only (i) the degraded measurements $\boldsymbol{y}$ and (ii) the frozen pre-trained diffusion model $s_\theta$. No paired $(\boldsymbol{x}, \boldsymbol{y})$ ground-truth dataset is needed, making the framework naturally scalable and flexibly conditioned on arbitrary measurement operators.

\subsection{Causal Autoregressive Solver}
\label{sec:method:inference}
A key distinction between offline and streaming video processing is causality. 
In offline generation, the model can attend to the entire video sequence, whereas in streaming reconstruction it only has access to past and current frames at time step $n$, not future ones. As a result, a standard DiT with full spatiotemporal attention cannot be directly used as a streaming video inverse solver.

Following this constraint, we design $q_\phi$ as a causal autoregressive solver that operates on temporal blocks of $T$ frames (Fig.~\ref{fig:overview}), inspired by~\cite{yin2024slow}.
Each block of degraded frames $\boldsymbol{y}_n$ is encoded to latents, and the model processes these blocks sequentially using a block-causal attention mechanism with two modes:

\begin{itemize}
    \item \textbf{Intra-block bidirectional attention.}
    Within the current block $n$, all tokens attend to each other to model rich local spatiotemporal structure.
    For a query $\mathbf{Q}_i$ from token $i$ in block $n$, with keys $\mathbf{K}_n$ and values $\mathbf{V}_n$ from all tokens in that block,
    \begin{equation}
    \mathrm{Att}_{\mathrm{intra}}(\mathbf{Q}_i, \mathbf{K}_n, \mathbf{V}_n)
    =
    \mathrm{softmax}\!\left(\frac{\mathbf{Q}_i \mathbf{K}_n^\top}{\sqrt{d_k}}\right)\mathbf{V}_n.
    \end{equation}

    \item \textbf{Inter-block causal attention.}
    Across blocks, tokens in block $n$ are allowed to attend only to previous reconstructed blocks $\hat{\boldsymbol{z}}_{<n}$.
    For the same query $\mathbf{Q}_i$, with keys $\mathbf{K}_{<n}$ and values $\mathbf{V}_{<n}$ from all past blocks,
    \begin{equation}
    \mathrm{Att}_{\mathrm{inter}}(\mathbf{Q}_i, \mathbf{K}_{<n}, \mathbf{V}_{<n})
    =
    \mathrm{softmax}\!\left(\frac{\mathbf{Q}_i \mathbf{K}_{<n}^\top}{\sqrt{d_k}}\right)\mathbf{V}_{<n},
    \end{equation}
\end{itemize}

In practice, we implement the inter-block attention with standard autoregressive KV caching: after reconstructing each block, we store its attention keys and values and reuse them as $\mathbf{K}_{<n}$ and $\mathbf{V}_{<n}$ for subsequent blocks. This avoids redundant computation over past frames while preserving exact causal behavior and keeping the per-frame cost low.

\subsection{Latent-Space Adaptation with Compact VAE}
\label{sec:method:acceleration}
The combination of variational diffusion distillation and our causal architecture already yields a strong solver, achieving about 15 FPS at $832 \times 480$ resolution with the original video VAE.
At this point, however, the DiT backbone is no longer the main bottleneck—the heavy video decoder becomes the limiting factor for real-time performance, consistent with prior observations~\cite{wang2025seedvr2, zhuang2025flashvsr} in latent video diffusion models.
A natural attempt to accelerate the system would replace the original VAE with a more efficient architecture.
However, this breaks the semantics of the latent space: the diffusion prior $p(\boldsymbol{z})$ and score $s_\theta$ are trained in the original latent space $\boldsymbol{z}$ induced by $(\mathcal{E}, \mathcal{D})$, while a new VAE $(\mathcal{E}', \mathcal{D}')$ defines a different latent space $\boldsymbol{z}'$.
Directly distilling a solver in $\boldsymbol{z}'$ without accounting for this distribution shift causes severe mismatch with the teacher prior.

To address this, we propose a teacher-space regularized distillation scheme that explicitly bridges the two latent spaces. We train a video inverse solver $q'_\phi(\boldsymbol{z}' | \boldsymbol{y})$ in the $\boldsymbol{z}'$-space using the objective
\begin{equation}
\label{eq:kl_objective_lean}
\begin{split}
\mathcal{L}(q'_\phi)
=
&~\mathbb{E}_{\boldsymbol{y}}
  \mathbb{E}_{\boldsymbol{z}' \sim q'_\phi(\boldsymbol{z}'|\boldsymbol{y})}
  \big[
  - \log p\big(\boldsymbol{y}| \mathcal{D}'(\boldsymbol{z}')\big)
  \big] \\
&+ \mathbb{E}_{t, \boldsymbol{\epsilon},\, \boldsymbol{z}' \sim q'_\phi(\cdot|\boldsymbol{y})}
  \Big[
  w(t)\,
  \big\|
  s_\theta(\boldsymbol{z}_t, t) - s_{q'}(\boldsymbol{z}_t, t)
  \big\|^2
  \Big],
\end{split}
\end{equation}
where we map the new latent $\boldsymbol{z}'$ back into the \emph{teacher} latent space via
\[
\boldsymbol{x} = \mathcal{D}'(\boldsymbol{z}'), \quad
\boldsymbol{z} = \mathcal{E}(\boldsymbol{x}),
\quad
\boldsymbol{z}_t = \alpha_t \boldsymbol{z} + \sigma_t \boldsymbol{\epsilon}.
\]
Thus, the likelihood term is evaluated in the new latent space, while the prior term uses the teacher diffusion score $s_\theta$ in the original latent space $\boldsymbol{z}$ after passing through $\mathcal{D}'$ and $\mathcal{E}$.
This constrains the new latent space $\boldsymbol{z}'$ so that, when decoded and re-encoded, it remains aligned with the teacher prior, enabling effective one-step distillation under a new VAE.

After exploring several efficient VAEs, we adopt {LeanVAE}~\cite{cheng2025leanvae} as our new video tokenizer.
LeanVAE is an ultra-efficient spatiotemporal VAE built on a lightweight NAF backbone with wavelet-based channel compression. Plugging LeanVAE into our teacher-space regularized distillation scheme yields an additional $> 2\times$ speedup over the original VAE, pushing InstantViR beyond 35 FPS while preserving diffusion-level fidelity and temporal coherence for streaming video inverse problems.

\begin{figure*}[ht]
    \centering
    \includegraphics[width=0.97\textwidth]{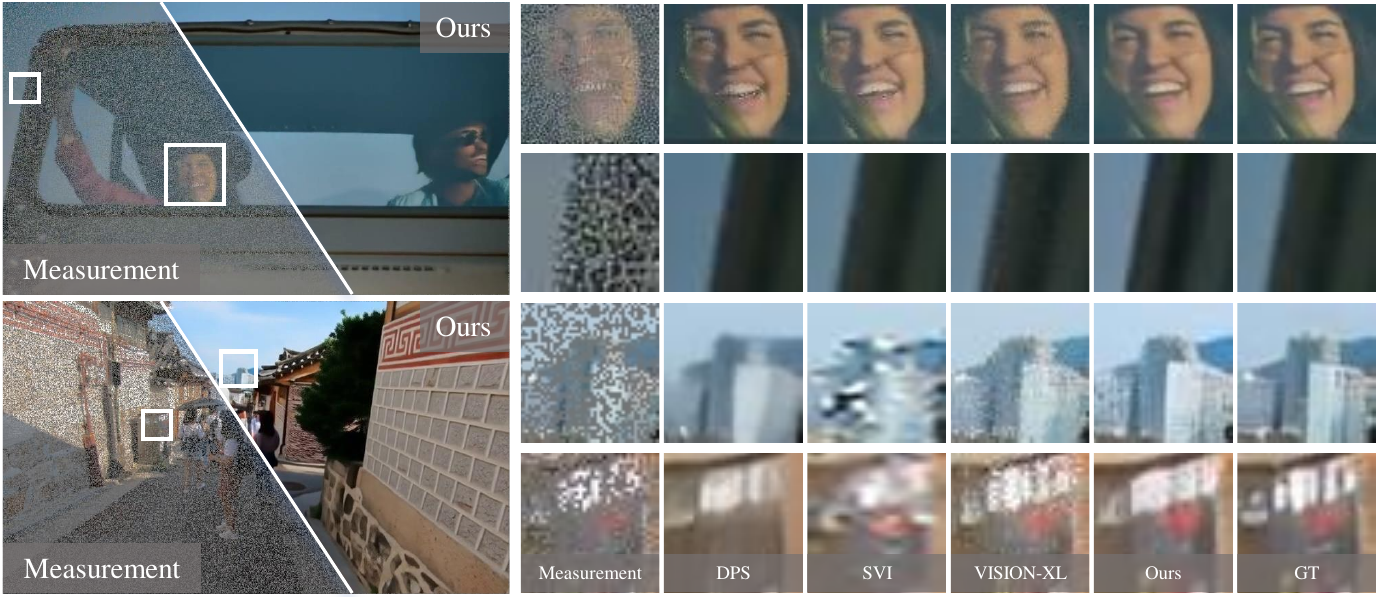}
    \caption{\textbf{Qualitative comparison for video random inpainting.}
    \textbf{(Top)} On the Open-Sora dataset~\cite{lin2024open}, our model reconstructs a high-fidelity and temporally consistent video from a 50\% masked measurement. The zoomed-in face sequence demonstrates the stability and fine detail of our single-step result. \textbf{(Bottom)} We demonstrate strong zero-shot generalization on the REDS dataset~\cite{zhang2018unreasonable}, our method generates a sharp, coherent video that is perceptually close to the ground truth.}
    \label{fig:inpainting}
    \vspace{-1em}
\end{figure*}

\section{Experiment}
\label{sec:experiment}
We evaluate InstantViR on a range of video inverse problems, compare it against strong diffusion-based baselines, analyze the contribution of its key components (amortized solver, causal architecture, LeanVAE), and demonstrate extended capabilities such as text-guided video reconstruction and editing. Further implementation details and additional results are provided in the supplementary material.


\begin{figure*}[t]
    \centering
    \includegraphics[width=0.999\textwidth]{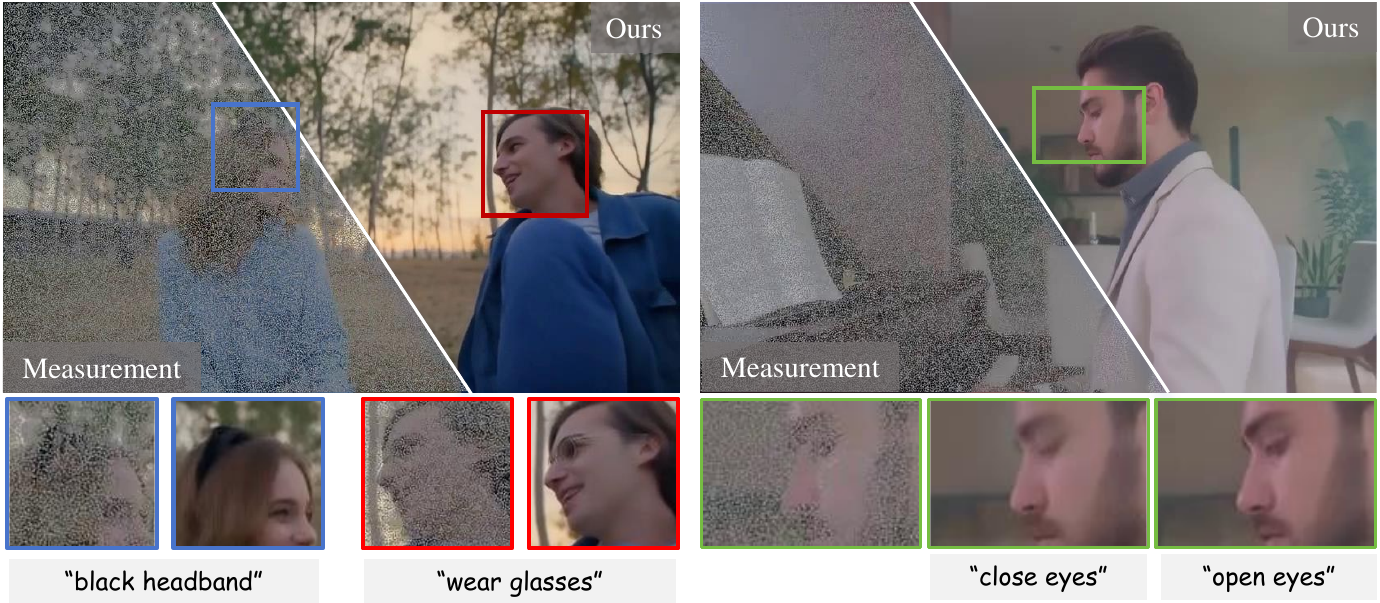}
    \vspace{-1.8em}
    \caption{\textbf{Results for text-guided video reconstruction.} InstantViR not only restores high-fidelity details from severely masked measurements (top row) but also uses text prompts to control the reconstruction. \textbf{(Left)} Given a masked input, our method can add specific details like a \textit{"black headband"} or \textit{"wear glasses"} as guided by the text. \textbf{(Right)} The framework can generate multi-modal, semantically distinct, and plausible outputs from the same masked input, such as reconstructing the subject with either \textit{"close eyes"} or \textit{"open eyes"}, demonstrating fine-grained controllability.}
    \label{fig:text_guidance}
    \vspace{-0.8em}
\end{figure*}

\subsection{Experimental Setup}
\label{sec:exp-setup}

\paragraph{Datasets and Base Model.}
Our primary training dataset is Open-Sora-v1.1~\cite{lin2024open}.
We use 6{,}000 video clips for training, without using any associated text labels.
Following our methodology in Sec.~\ref{sec:method:framework}, we do not require paired $(\mathbf{x}, \mathbf{y})$ data: degraded measurements $\mathbf{y}$ are generated on-the-fly from clean videos $\mathbf{x}$ via the known forward operators, and only $\mathbf{y}$ is only used to query the teacher diffusion model.
For evaluation, we use a held-out set of 500 Open-Sora videos and the standard REDS30 dataset~\cite{nah2019ntire} to test zero-shot generalization.
All videos are processed at $832\times480$ resolution.
Our video diffusion prior is the Wan2.1-1.3B model~\cite{wan2.1}. Training takes two weeks on 8 NVIDIA A100.


\begin{table}[t]
\centering
\caption{\textbf{Quantitative results of temporal quality and inference speed.} The former is evaluated with FVD $\downarrow$ and the latter is with FPS $\uparrow$. Best results are in \textbf{bold}, suboptimal are \underline{underlined}.}
\label{tab:temporal_fps}
\resizebox{\columnwidth}{!}{%
\begin{tabular}{@{}l|ccc|c@{}}
\toprule
\multirow{2}{*}{\textbf{Method}} & \multicolumn{3}{c|}{\textbf{FVD $\downarrow$}} & \multirow{2}{*}{\textbf{Avg. FPS $\uparrow$}} \\
\cmidrule(r){2-4}
& \textbf{Inpainting} & \textbf{Super-Res.} & \textbf{Deblur} & \\ 
\midrule
DPS & 375.81 & 711.61 & 783.10 & \textless 0.02 \\
DiffIR2VR & - & 311.61 & - & 0.12 \\
SVI & 219.90 & 176.60 & 154.38 & 0.29 \\
VISION-XL & 224.74 & 172.79 & 138.79 & \textless 0.17 \\
\textbf{InstantViR (Ours)} & \underline{136.06} & \textbf{153.13} & \underline{110.51} & \underline{13.91} \\
\textbf{InstantViR$^\dag$ (Ours)} & \textbf{132.59} & \underline{156.43} & \textbf{103.45} & \textbf{35.56} \\ 
\bottomrule
\end{tabular}%
}
\end{table}

\vspace{-0.9em}
\paragraph{Baselines.}
We compare InstantViR against a suite of strong diffusion-based video inverse problem solvers:
DPS~\cite{chung2022diffusion} as a foundational diffusion posterior sampler;
SVI~\cite{kwon2025solving} and VISION-XL~\cite{kwon2024visionxlhighdefinitionvideo}, which adapt image diffusion priors with batch-consistent sampling; and DiffIR2VR~\cite{yeh2024diffir2vr}, which uses hierarchical latent warping for video restoration.
We use official implementations for all baselines to ensure high-quality generation and fair comparisons.

\begin{table*}[t]
\centering
\caption{\textbf{Quantitative results of spatial quality.} Metrics include PSNR, SSIM, LPIPS. Best results are in \textbf{bold}, suboptimal are \underline{underlined}.}
\vspace{-0.5em}
\label{tab:spatial_quality}
\resizebox{\textwidth}{!}{%
\begin{tabular}{@{}l|ccc|ccc|ccc@{}}
\toprule
\multirow{2}{*}{\textbf{Method}} & \multicolumn{3}{c|}{\textbf{50\% Random Inpainting}} & \multicolumn{3}{c|}{\textbf{4$\times$ Super-Resolution}} & \multicolumn{3}{c}{\textbf{Gaussian Deblur}} \\
\cmidrule(l){2-10} 
& PSNR $\uparrow$ & SSIM $\uparrow$ & LPIPS $\downarrow$ & PSNR $\uparrow$ & SSIM $\uparrow$ & LPIPS $\downarrow$ & PSNR $\uparrow$ & SSIM $\uparrow$ & LPIPS $\downarrow$ \\ 
\midrule
DPS & 27.68 & 0.92 & 0.32 & 22.78 & 0.91 & 0.46 & 23.54 & 0.88 & 0.46 \\
DiffIR2VR & - & - & - & 33.44 & 0.92 & 0.33 & - & - & - \\
SVI & 29.42 & 0.90 & 0.17 & 33.85 & 0.96 & \textbf{0.17} & 26.93 & 0.89 & 0.31 \\
VISION-XL & \underline{30.83} & 0.95 & 0.25 & \textbf{35.69} & \textbf{0.98} & 0.24 & 30.03 & 0.93 & 0.28 \\
\textbf{InstantViR (Ours)} & 30.54 & \textbf{0.97} & \underline{0.12} & \underline{34.91} & \underline{0.96} & 0.23 & \textbf{31.85} & \textbf{0.97} & \underline{0.17} \\
\textbf{InstantViR$^\dag$ (Ours)} & \textbf{31.78} & \underline{0.96} & \textbf{0.13} & 27.04 & 0.95 & \underline{0.22} & \underline{31.16} & \underline{0.97} & \textbf{0.15} \\ 
\bottomrule
\end{tabular}%
}
\vspace{-1em}
\end{table*}

\paragraph{Metrics.}
We evaluate performance using four complementary criteria: (1) Per-frame reconstruction quality: PSNR and SSIM; (2) Perceptual quality: LPIPS~\cite{zhang2018unreasonable}; (3) Temporal consistency: Fréchet Video Distance (FVD)~\cite{unterthiner2019fvd}, which measures the distributional similarity and temporal coherence across frames; (4) Inference speed: frames per second (FPS) on a single NVIDIA A800 80GB GPU.

\subsection{Standard Video Inverse Problems} \label{sec:results}
We first evaluate InstantViR on three canonical video inverse problems: random inpainting, Gaussian deblurring, and $4\times$ super-resolution.
Table~\ref{tab:spatial_quality} summarizes per-frame reconstruction and perceptual quality, while Table~\ref{tab:temporal_fps} reports temporal consistency and runtime.

As shown in Table~\ref{tab:spatial_quality}, our one-step solver \textbf{InstantViR} (using the original VAE) and its accelerated variant \textbf{InstantViR$^\dag$} (with LeanVAE) achieve state-of-the-art or highly competitive performance across all tasks, surpassing iterative methods such as SVI and VISION-XL in PSNR, SSIM, and LPIPS.
This confirms that the amortized student solver successfully inherits the high-fidelity prior of the teacher video diffusion model.

Table~\ref{tab:temporal_fps} and Fig.~\ref{fig:teaser} (Bottom-Right) highlight the substantial inference speedup achieved by InstantViR and InstantViR$^\dag$.
Sampling-based baselines~\cite{chung2022diffusion, kwon2025solving, kwon2024visionxlhighdefinitionvideo} operate at less than 1 FPS and thus are impractical for streaming.
In contrast, InstantViR achieves superior temporal consistency (lowest FVD) on all tasks while running at $\sim 14$ FPS.
By replacing the original VAE with LeanVAE via our teacher-space regularized distillation scheme (Sec.~\ref{sec:method:acceleration}), InstantViR$^\dag$ reaches over \textbf{35 FPS}, achieving up to over \textbf{100$\times$} speedup over SVI while further improving temporal coherence.

\paragraph{Random Inpainting.}
Fig.~\ref{fig:inpainting} shows qualitative results for $50\%$ random inpainting.
On the Open-Sora test set, InstantViR recovers fine, high-frequency details (e.g., individual teeth) that are blurred or missing in the baselines. Moreover, the reconstructions remain temporally stable across frames.
On the unseen REDS~\cite{nah2019ntire} dataset, our method still produces sharp, coherent distant city views, whereas baseline methods often exhibit blur and temporal jitter, demonstrating strong zero-shot generalization.

\begin{figure}[t]
    \centering
    \setlength{\tabcolsep}{1pt}
    \resizebox{\columnwidth}{!}{%
    \begin{tabular}{cccc} 
    & \multicolumn{3}{c}{
        \begin{tikzpicture}[baseline]
            \draw[->, >=latex, line width=0.35mm] (0,0.1) -- (7.2cm,0.1) 
            node[right, xshift=2mm] at (7.2cm, 0.1) {Time};
        \end{tikzpicture}
      } \\
    \begin{turn}{90} \,\small Mea. (Blur) \end{turn} &
    \includegraphics[width=0.33\columnwidth]{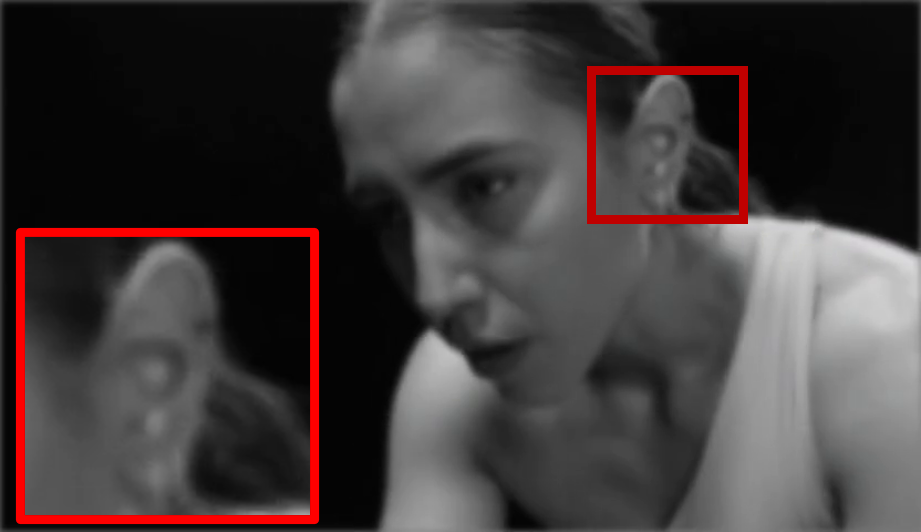} &
    \includegraphics[width=0.33\columnwidth]{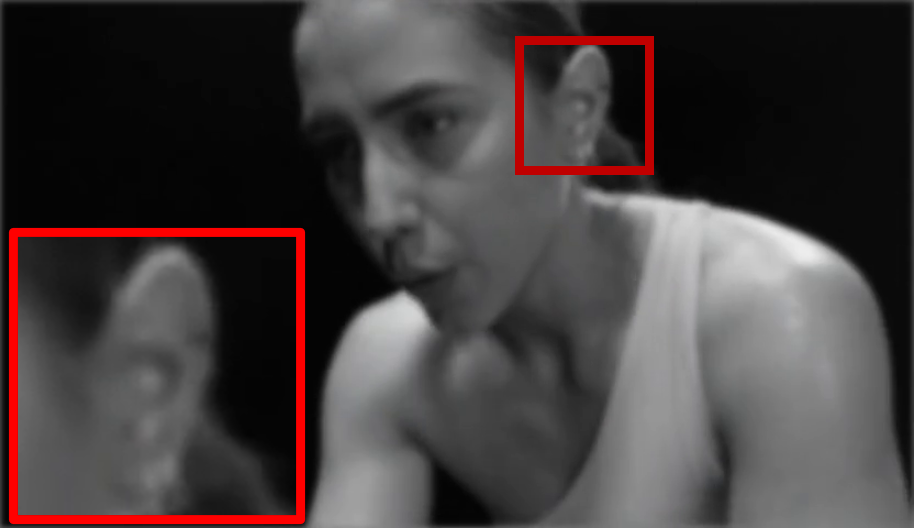} &
    \includegraphics[width=0.33\columnwidth]{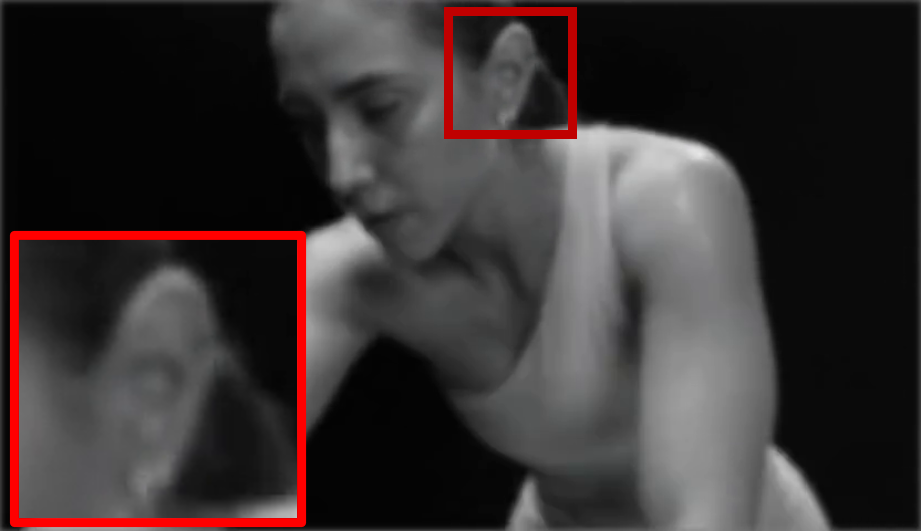} \\
    \begin{turn}{90} \,\,\,\,\small DPS~\cite{chung2022diffusion} \end{turn} &
    \includegraphics[width=0.33\columnwidth]{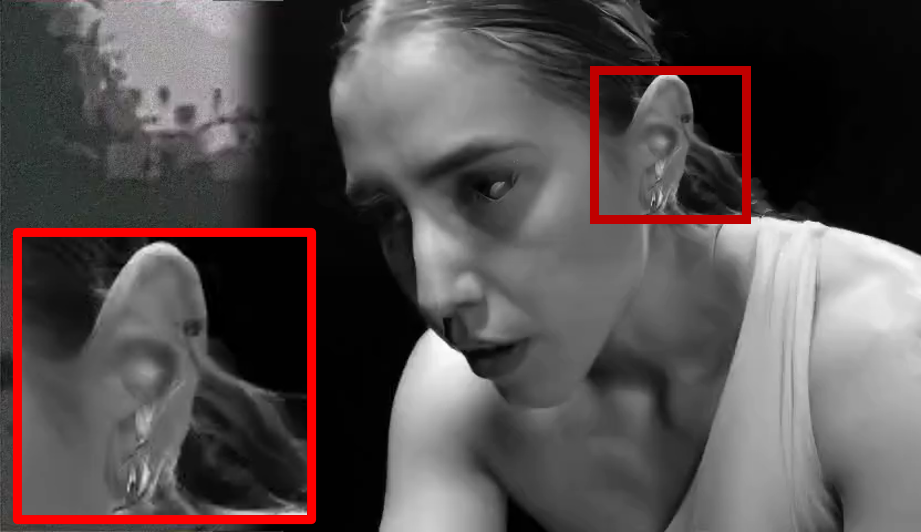} &
    \includegraphics[width=0.33\columnwidth]{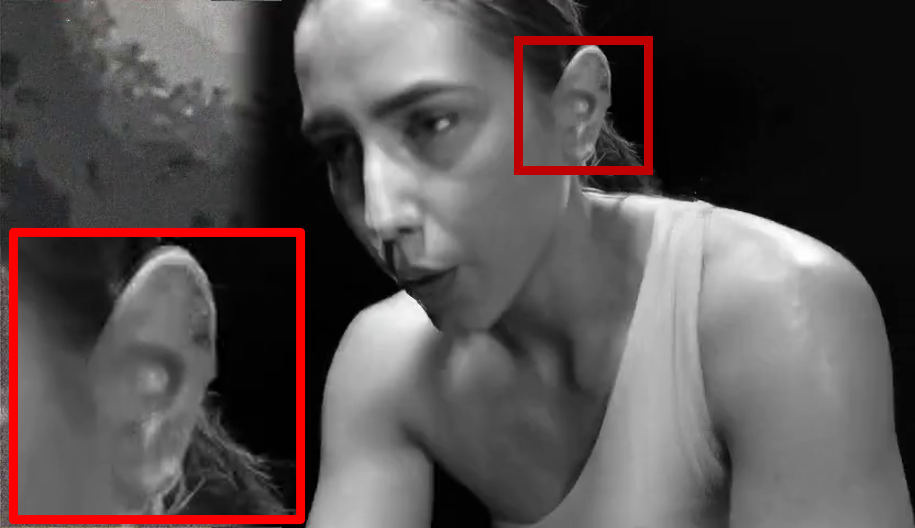} &
    \includegraphics[width=0.33\columnwidth]{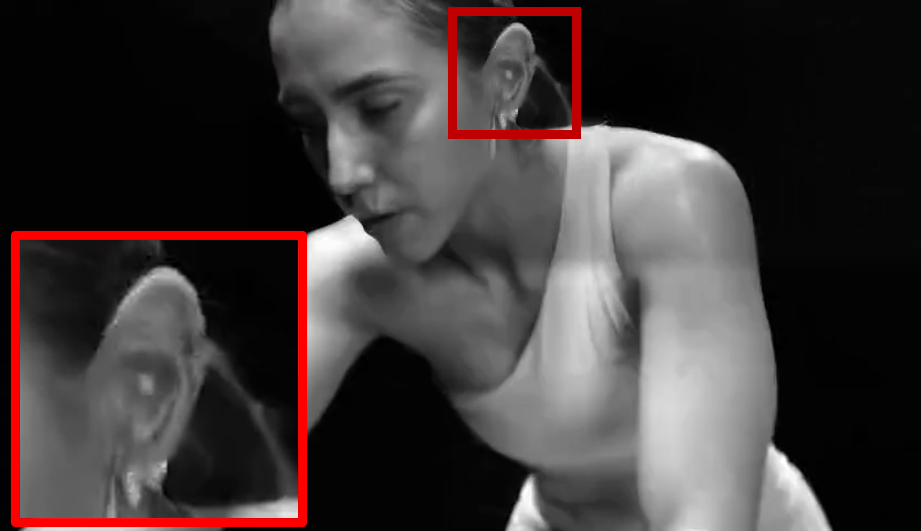} \\
    \begin{turn}{90} \,\,\,\,\small \begin{tabular}{c} SVI~\cite{kwon2025solving} \end{tabular} \end{turn} &
    \includegraphics[width=0.33\columnwidth]{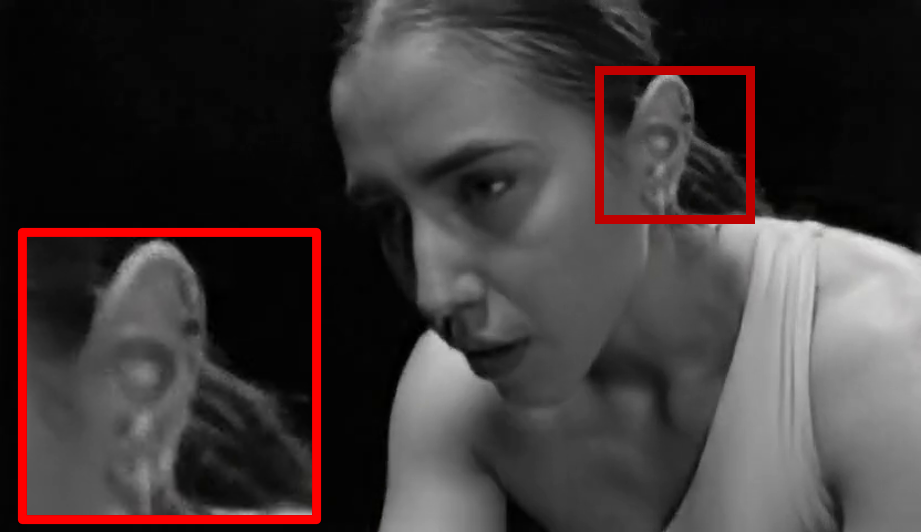} &
    \includegraphics[width=0.33\columnwidth]{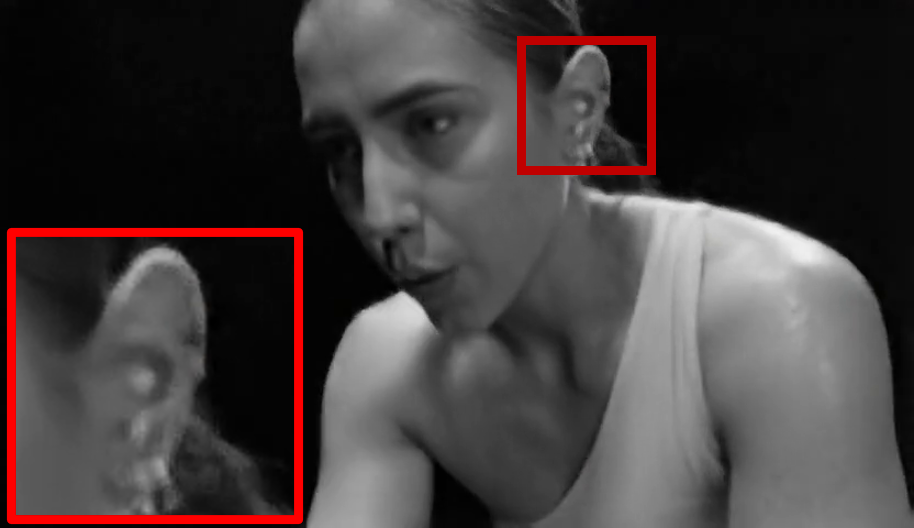} &
    \includegraphics[width=0.33\columnwidth]{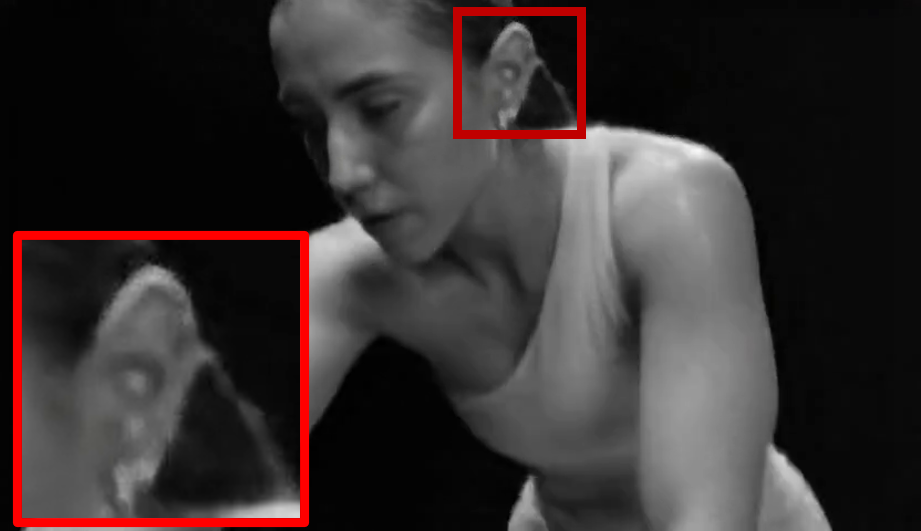} \\
    \begin{turn}{90} \small {Vis-XL~\cite{kwon2024visionxlhighdefinitionvideo}} \end{turn} &
    \includegraphics[width=0.33\columnwidth]{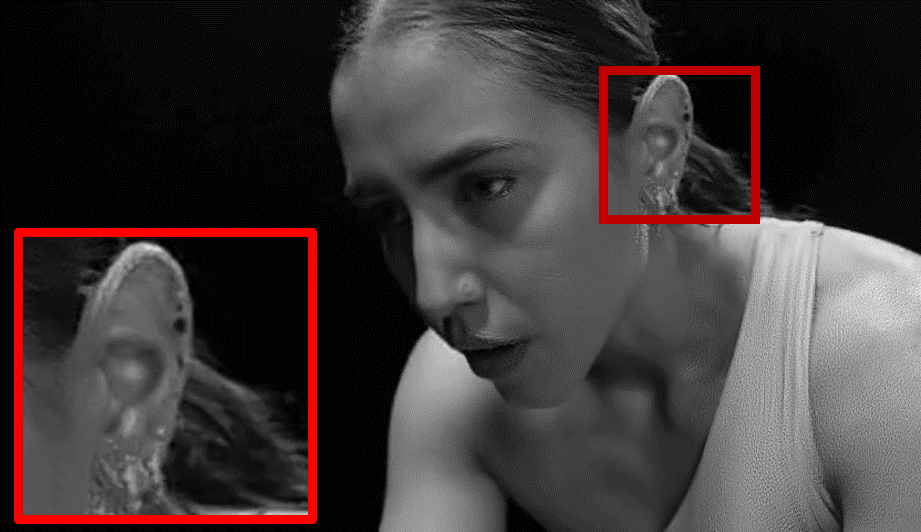} &
    \includegraphics[width=0.33\columnwidth]{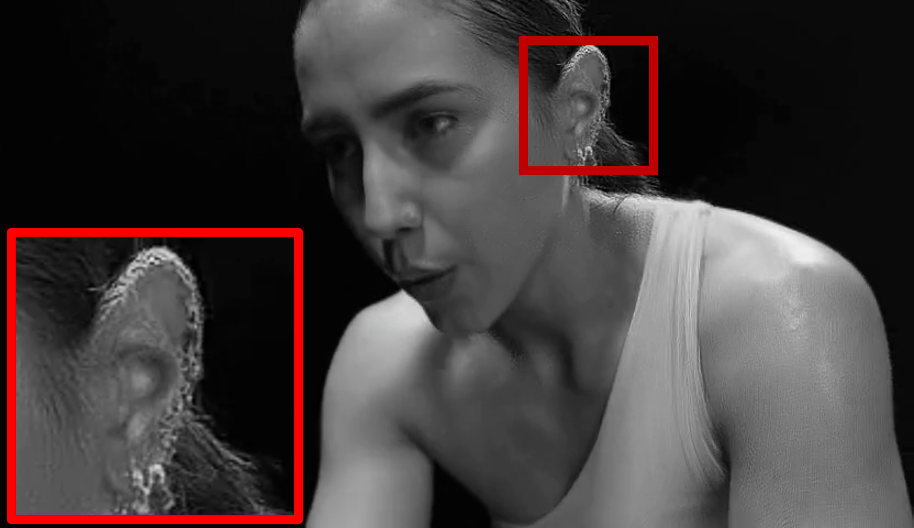} &
    \includegraphics[width=0.33\columnwidth]{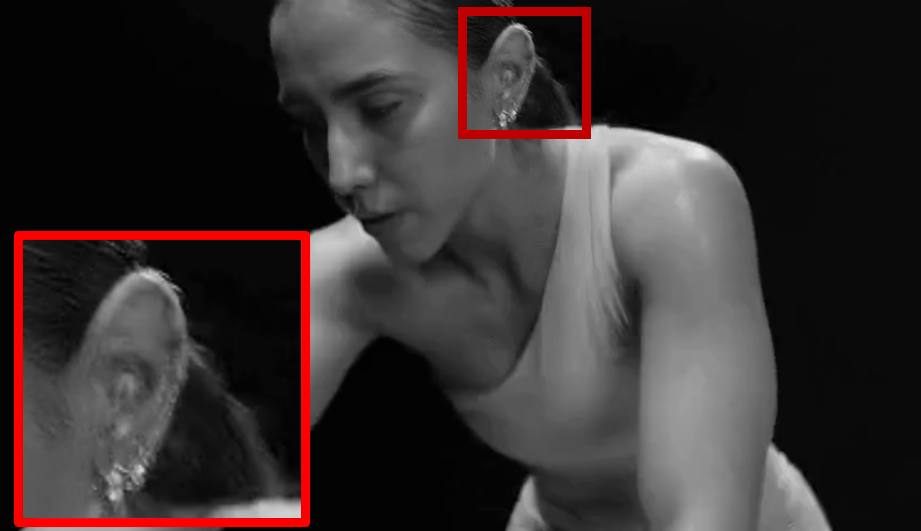} \\
    \begin{turn}{90} \,\,\,\,\,\,\,\,\,\,\small Ours \end{turn} &
    \includegraphics[width=0.33\columnwidth]{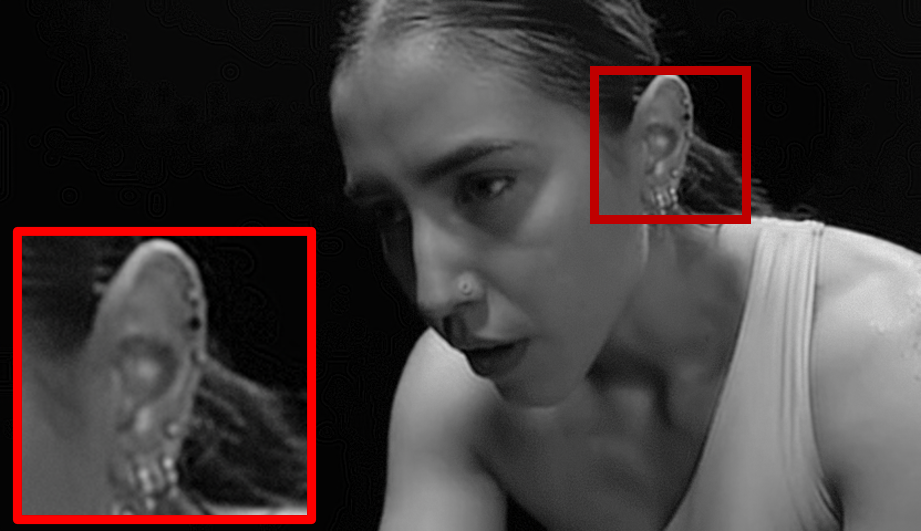} &
    \includegraphics[width=0.33\columnwidth]{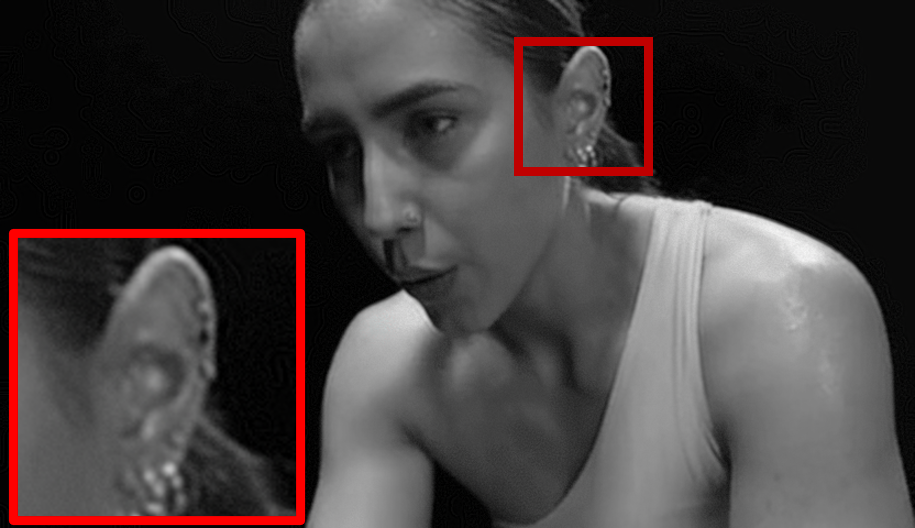} &
    \includegraphics[width=0.33\columnwidth]{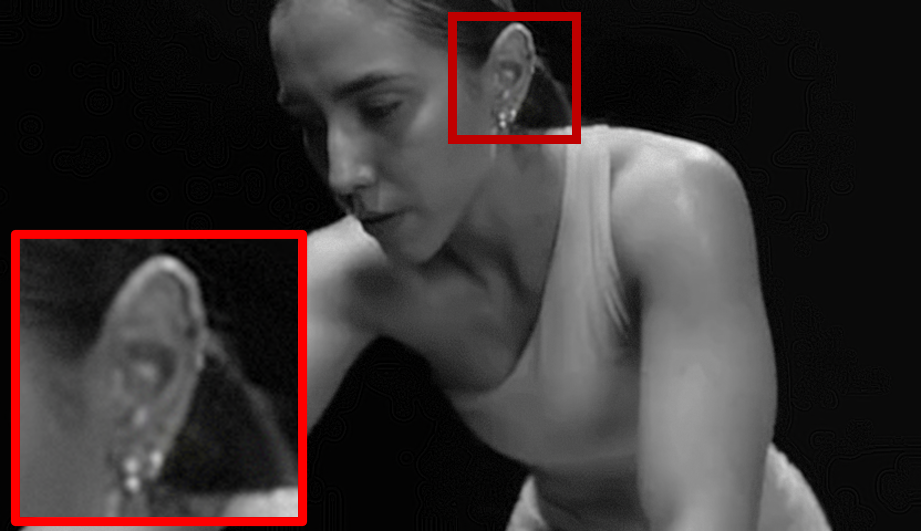} \\
    \begin{turn}{90} \,\,\,\,\,\,\,\,\small GT \end{turn} &
    \includegraphics[width=0.33\columnwidth]{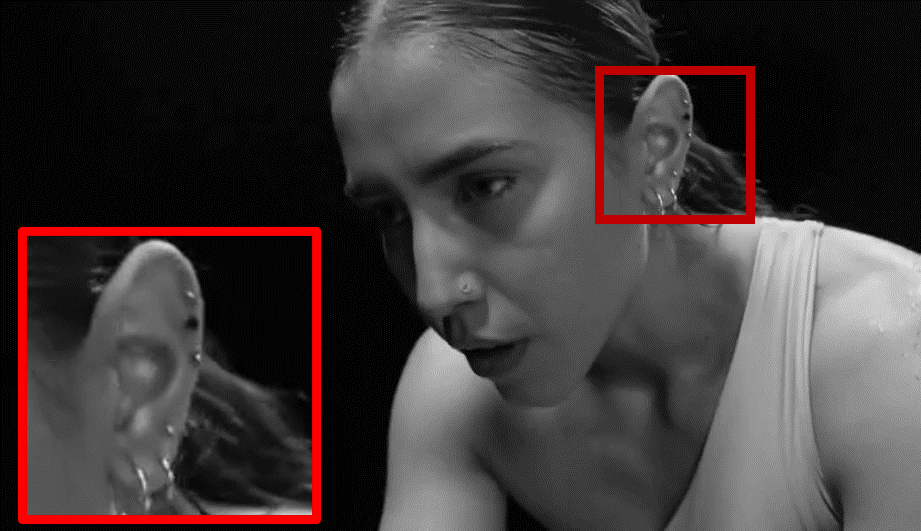} &
    \includegraphics[width=0.33\columnwidth]{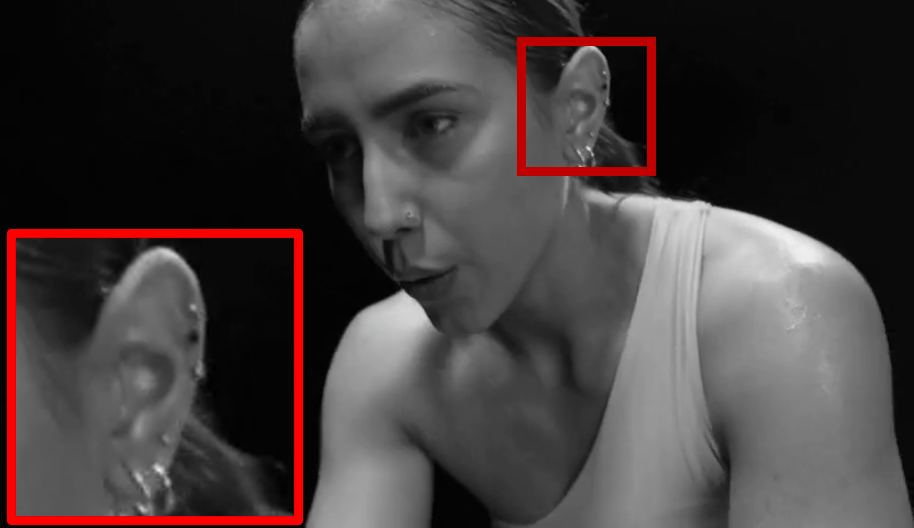} &
    \includegraphics[width=0.33\columnwidth]{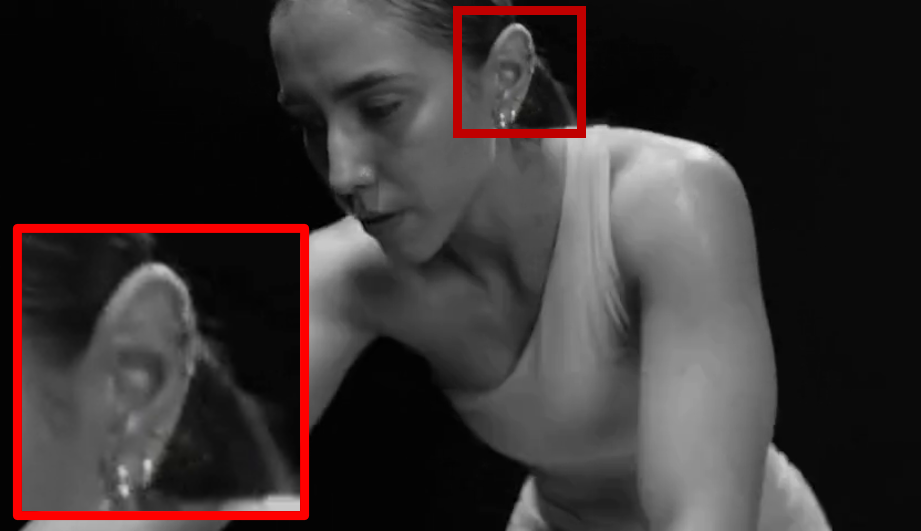} \\
    \end{tabular}%
    } 
    \vspace{-0.7em}
    \caption{{Qualitative comparison for video Gaussian deblurring.}}
    \vspace{-1.1em}
    \label{fig:deblur}
\end{figure}
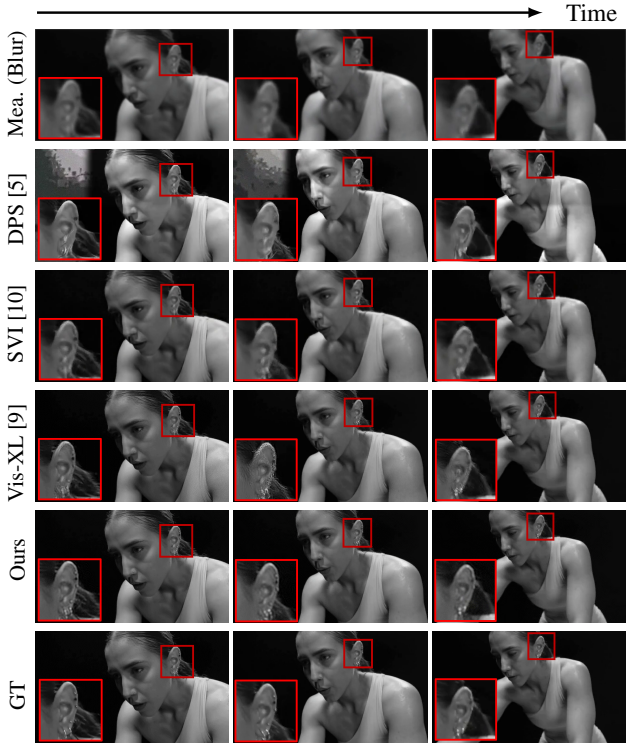

\vspace{-0.5em}
\paragraph{Gaussian Deblurring.}
For Gaussian deblurring (Fig.~\ref{fig:deblur}), sampling-based methods such as SVI~\cite{kwon2025solving} tend to suffer from noticeable temporal flicker, leading to unstable videos.
InstantViR, by contrast, produces sharp and consistent reconstruction across frames, restoring fine details, such as the athlete’s earrings, without sacrificing temporal coherence.

\begin{figure}[t]
    \centering
    \setlength{\tabcolsep}{1pt}
    \resizebox{\columnwidth}{!}{%
    \begin{tabular}{cccc} 
    & \multicolumn{3}{c}{
        \begin{tikzpicture}[baseline]
            \draw[->, >=latex, line width=0.35mm] (0,0.1) -- (7.2cm,0.1) 
            node[right, xshift=2mm] at (7.2cm, 0.1) {Time};
        \end{tikzpicture}
      } \\
    \begin{turn}{90} \,\,\,\small Mea. (LR) \end{turn} &
    \includegraphics[width=0.33\columnwidth]{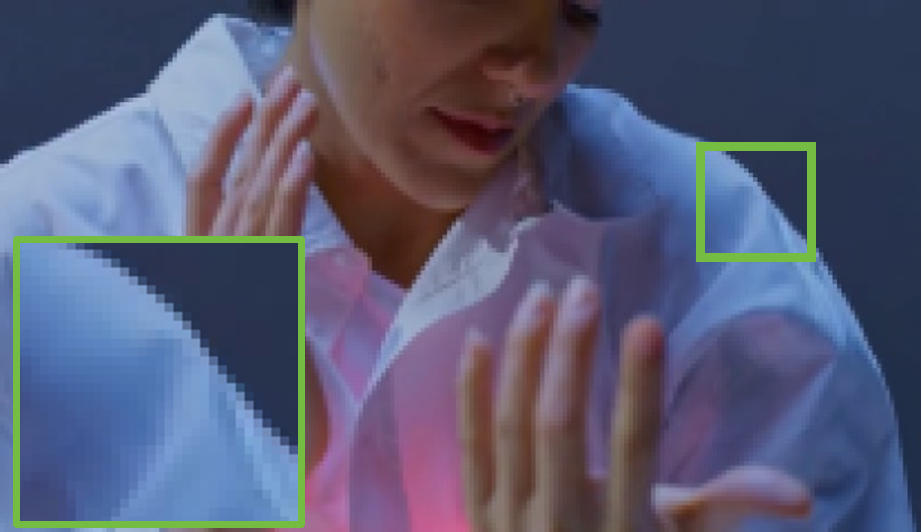} &
    \includegraphics[width=0.33\columnwidth]{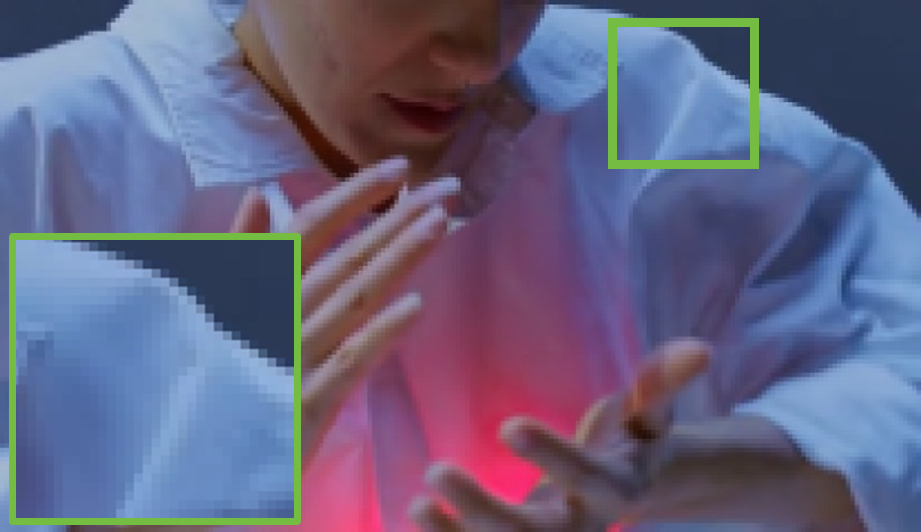} &
    \includegraphics[width=0.33\columnwidth]{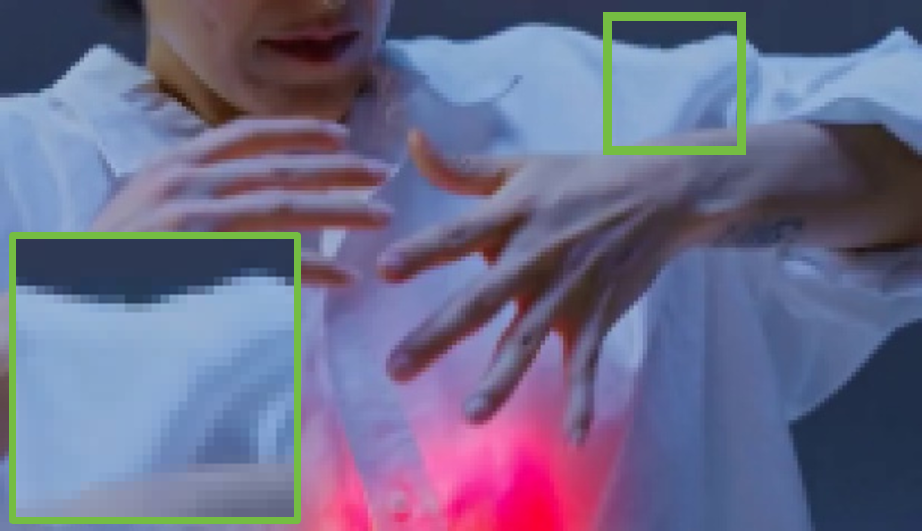} \\
    \begin{turn}{90} \,\,\small DiffIR2VR \end{turn} &
    \includegraphics[width=0.33\columnwidth]{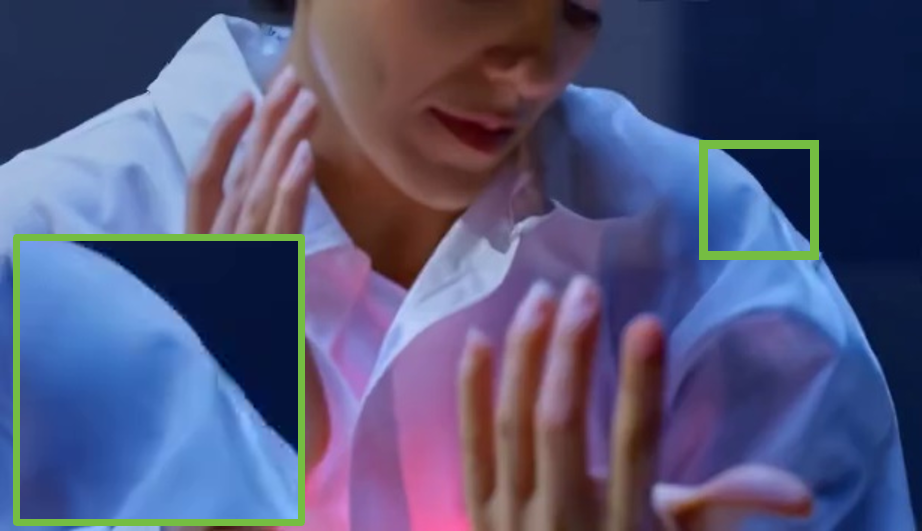} &
    \includegraphics[width=0.33\columnwidth]{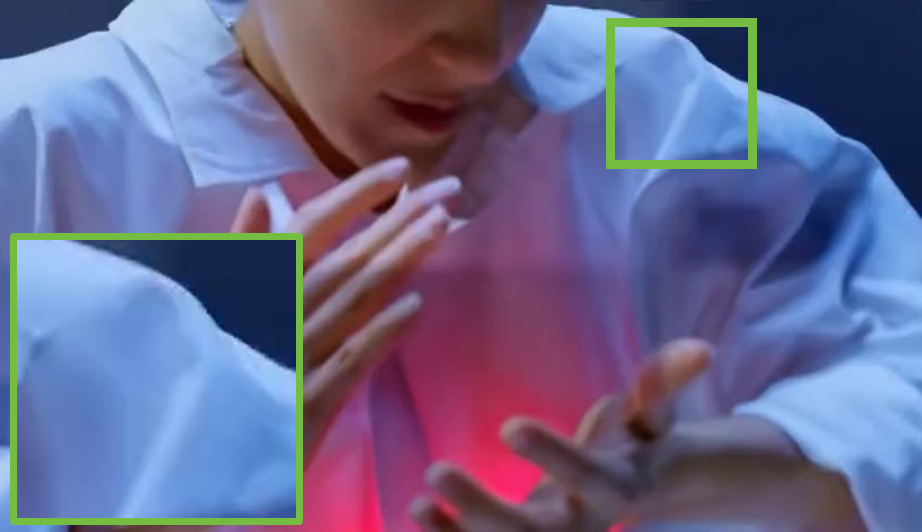} &
    \includegraphics[width=0.33\columnwidth]{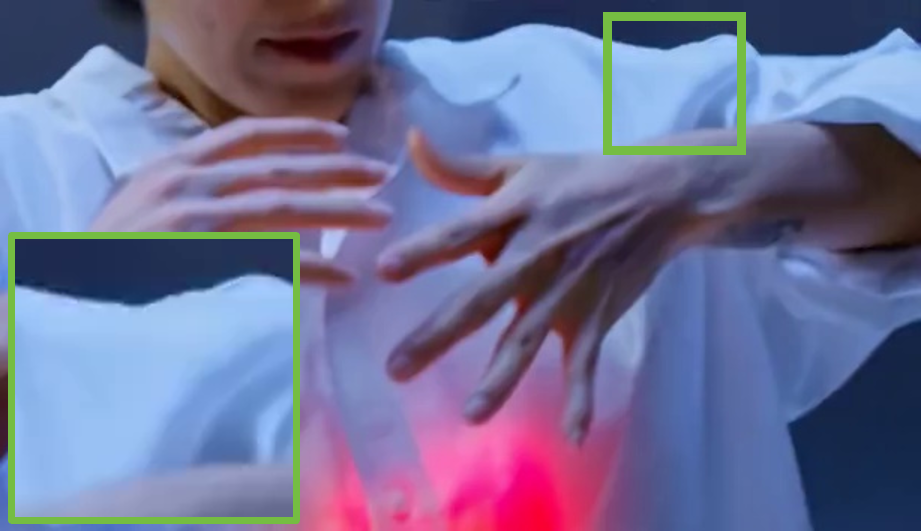} \\
    \begin{turn}{90} \,\,\,\,\small \begin{tabular}{c} SVI~\cite{kwon2025solving} \end{tabular} \end{turn} &
    \includegraphics[width=0.33\columnwidth]{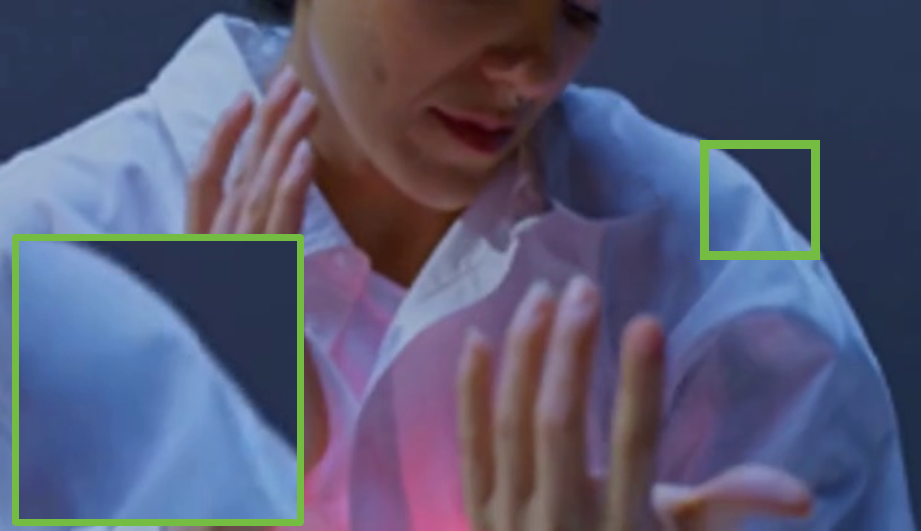} &
    \includegraphics[width=0.33\columnwidth]{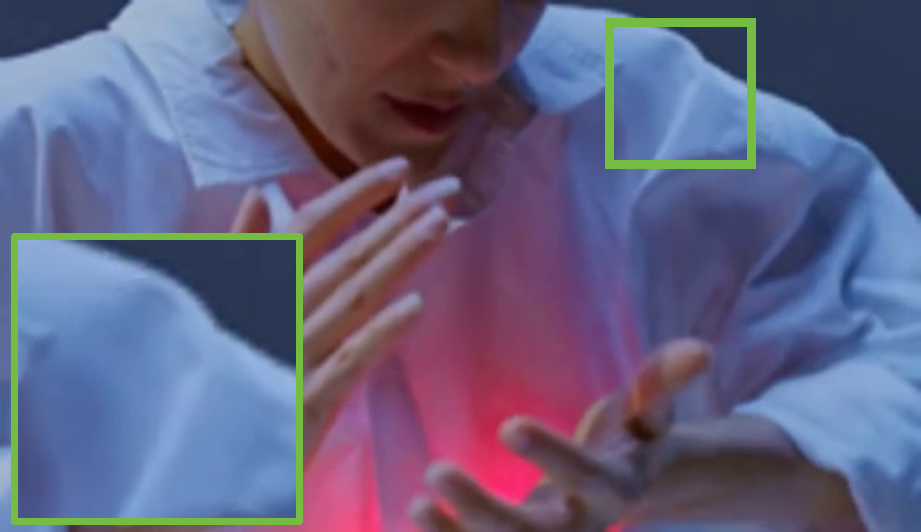} &
    \includegraphics[width=0.33\columnwidth]{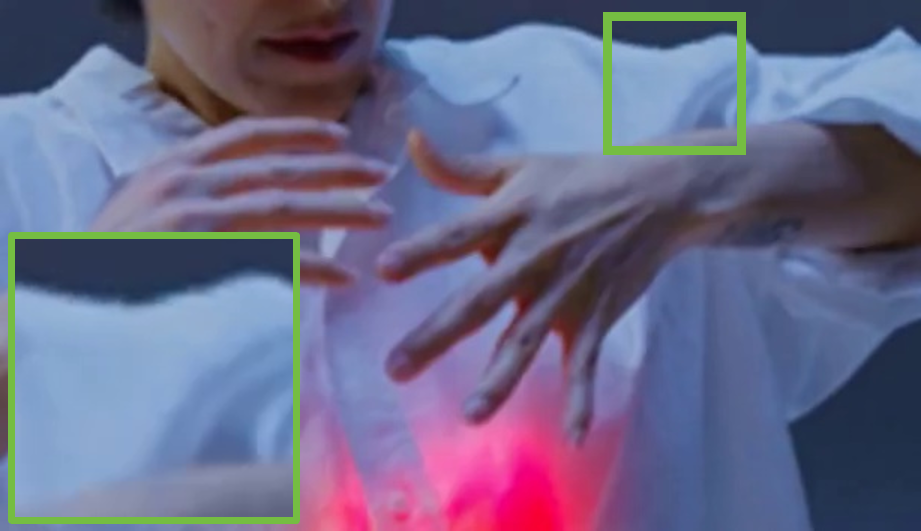} \\
    \begin{turn}{90} \small {Vis-XL~\cite{kwon2024visionxlhighdefinitionvideo}} \end{turn} &
    \includegraphics[width=0.33\columnwidth]{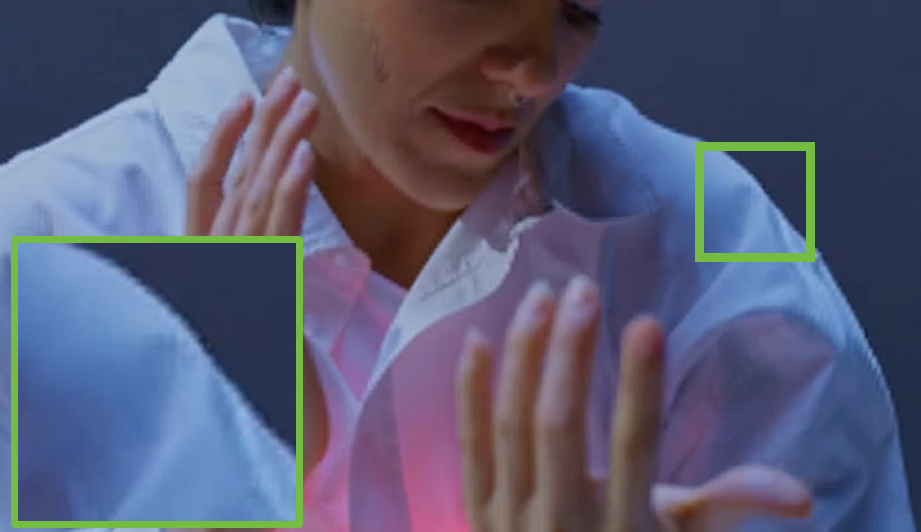} &
    \includegraphics[width=0.33\columnwidth]{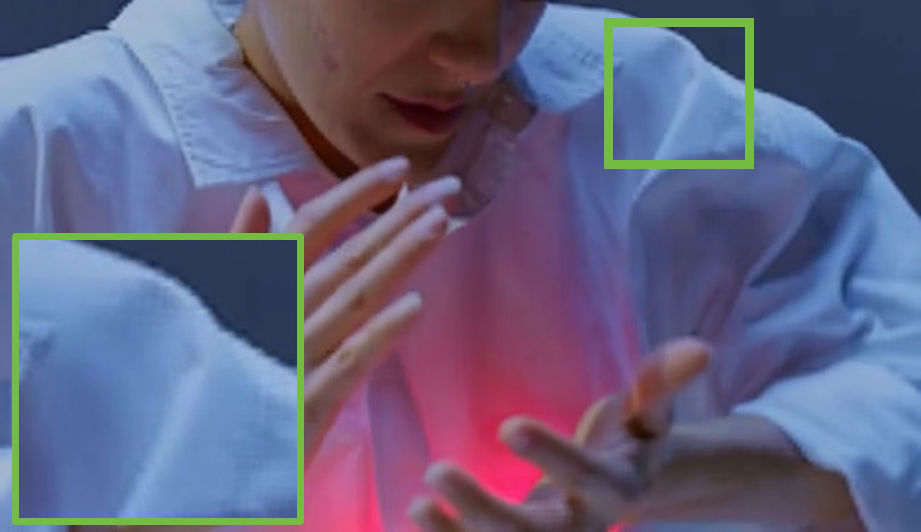} &
    \includegraphics[width=0.33\columnwidth]{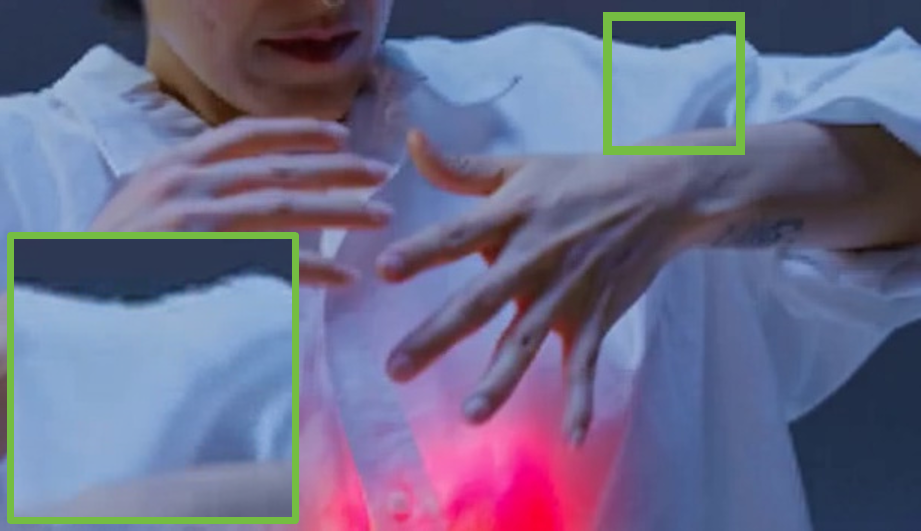} \\
    \begin{turn}{90} \,\,\,\,\,\,\,\,\small Ours \end{turn} &
    \includegraphics[width=0.33\columnwidth]{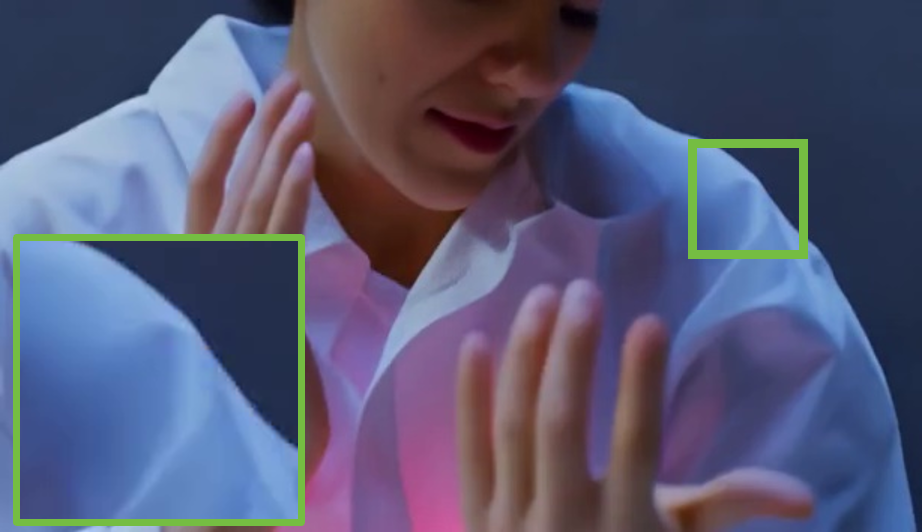} &
    \includegraphics[width=0.33\columnwidth]{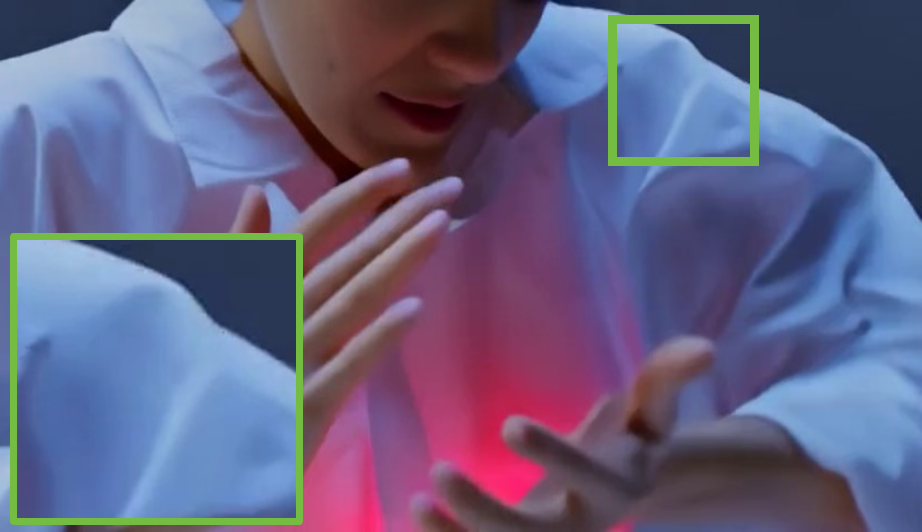} &
    \includegraphics[width=0.33\columnwidth]{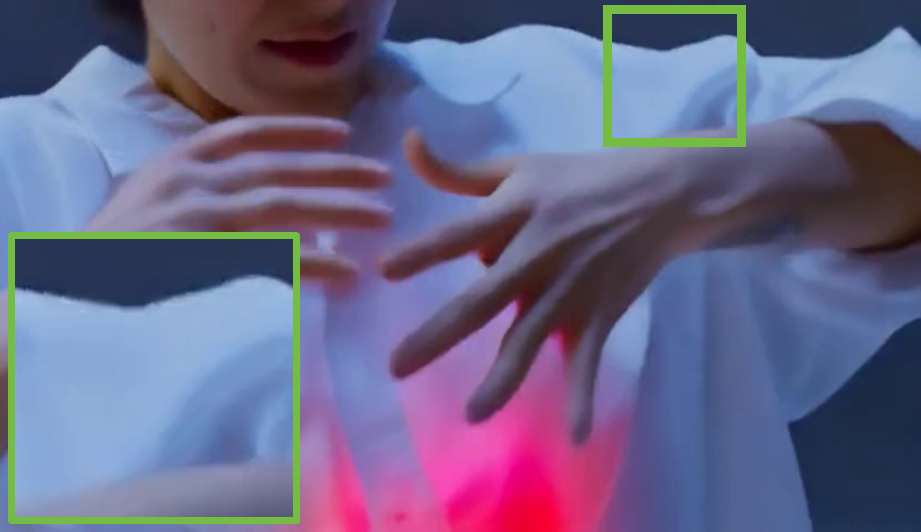} \\
    \begin{turn}{90} \,\,\,\,\,\,\,\,\,\,\small GT \end{turn} &
    \includegraphics[width=0.33\columnwidth]{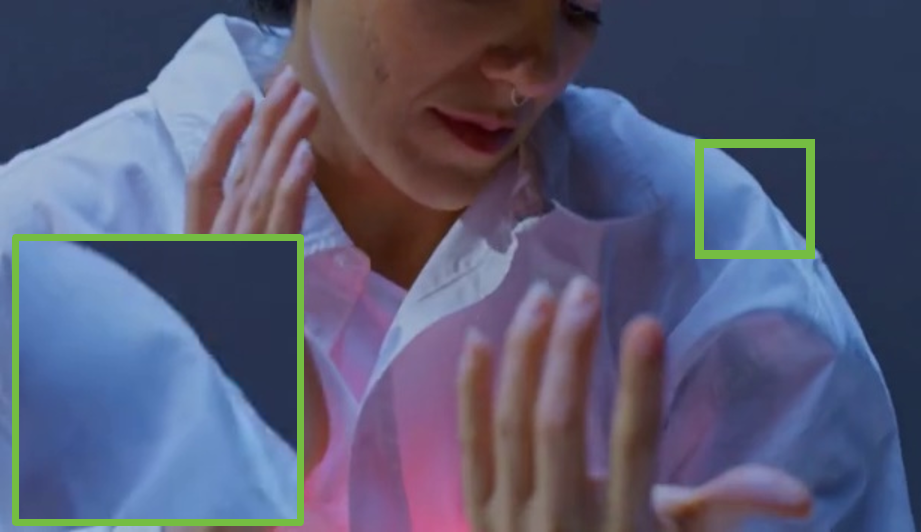} &
    \includegraphics[width=0.33\columnwidth]{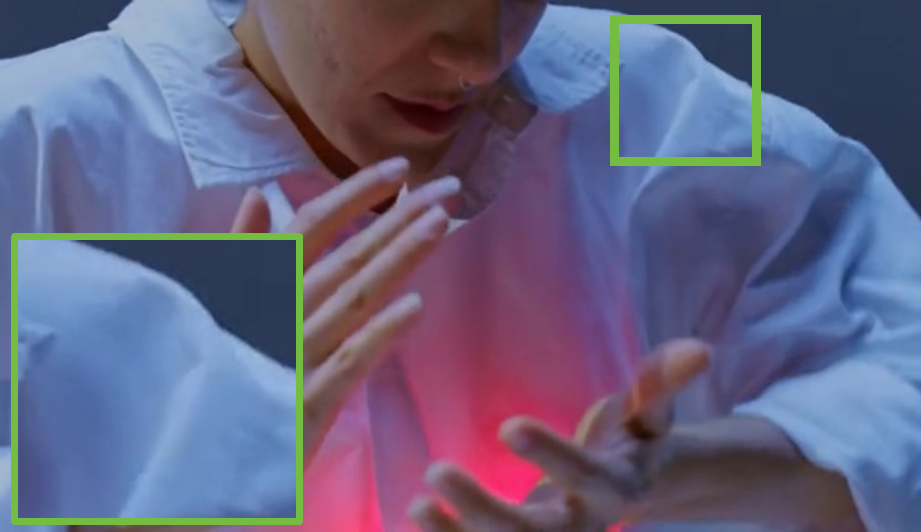} &
    \includegraphics[width=0.33\columnwidth]{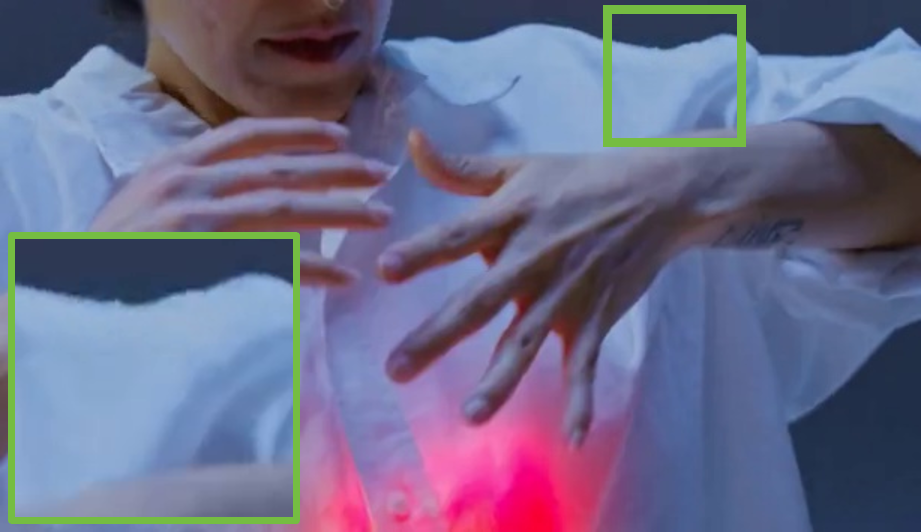} \\
    \end{tabular}%
    } 
    \vspace{-0.7em}
    \caption{{Qualitative comparison for video super-resolution (4x).}}
    \vspace{-1.1em}
    \label{fig:sr}
\end{figure}
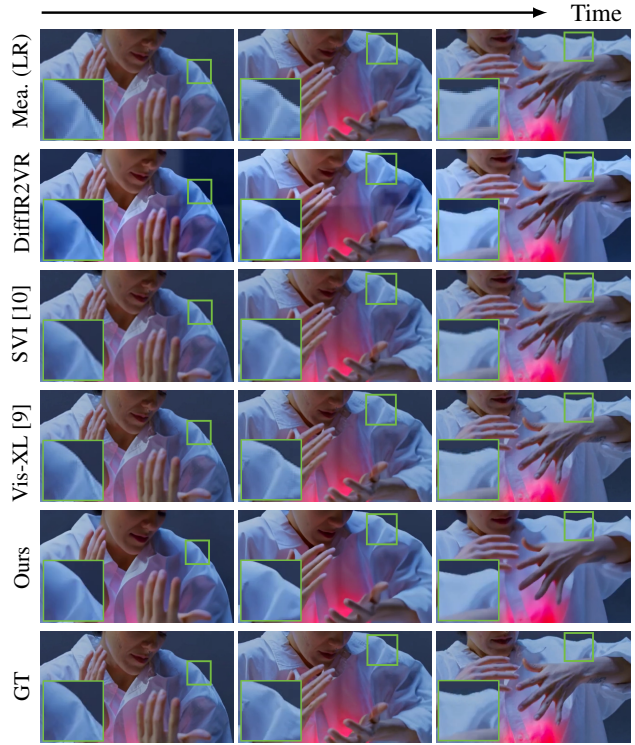

\vspace{-0.5em}
\paragraph{$4\times$ Super-Resolution.}
In the $4\times$ super-resolution setting (Fig.~\ref{fig:sr}), InstantViR remains highly competitive with or superior to SVI and VISION-XL. While these baselines are spatially stable, they tend to produce unrealistic jagged edges or overly smoothed details. Our method recovers sharp textures that are both perceptually detailed and temporally aligned with the ground truth.

\subsection{Text-Guided Video Reconstruction}
\label{sec:application_text}
Because our framework builds on a text-conditional video diffusion prior (Wan2.1), InstantViR and InstantViR$^\dag$ can naturally generalize to text-guided video reconstruction and editing by conditioning the amortized inference objective (Eq.~\ref{eq:kl_objective_latent}) on prompts.
As illustrated in Fig.~\ref{fig:teaser} and \ref{fig:text_guidance}, this enables fine-grained semantic control over the reconstructed content.

Given a partially masked input video and a text prompt, the model generates plausible, semantically aligned completions in the missing regions.
For the same input sequence, different prompts (e.g., \textit{"wearing glasses"} vs. \textit{"wearing a headband"}) produce distinct but temporally consistent edits of the subject.
In another example, we can control whether a musician \textit{"closes eyes"} or \textit{"keeps eyes open"} while playing the piano.
These results demonstrate that InstantViR not only reconstructs measurement-consistent content (likelihood term), but can also hallucinate coherent new details guided by text (prior term).

This unique combination of high speed, causal streaming, and prompt-level controllability opens up new possibilities for real-time interactive applications, such as live broadcast enhancement, streaming video restoration, and on-the-fly creative editing.

\section{Conclusion}
\label{sec:conclusion}
In this paper, we presented \textbf{InstantViR}, a novel amortized inference framework that resolves the core trade-off between reconstruction quality and speed in video inverse problems. By distilling a powerful bidirectional video diffusion prior into a lightweight, one-step causal autoregressive solver, InstantViR achieves both robust temporal consistency and real-time performance. Our framework eliminates the need for paired ground truth, replaces the heavy video VAE with a lightweight alternative via teacher-space regularization, and bypasses the slow, iterative posterior sampler. We demonstrate these gains on various streaming video tasks, attaining state-of-the-art temporal quality while surpassing 35 FPS.

\vspace{-0.9em}
\paragraph{Limitations and Future Directions.}
Despite these advances, our accelerated model with latent-space adaptation still slightly lags behind the distilled model using the original VAE in reconstruction quality. This suggests that latent distribution shift remains a limiting factor; jointly fine-tuning the lightweight VAE to better align its latent space with the teacher’s original space may further close this gap. 

Beyond natural video restoration, the label-free, amortized nature of our framework makes it well suited for scaling to larger unlabeled video corpora and for deployment in other real-time domains such as medical video enhancement. We view InstantViR as a step toward practical, diffusion-level video reconstruction that is both fast and controllable.



{
    \small
    \bibliographystyle{ieeenat_fullname}
    \bibliography{main}

@String(ICCV  = {Int. Conf. Comput. Vis.})

@String(NeurIPS = {Adv. Neural Inform. Process. Syst.})

@String(ICCV  = {ICCV})

@String(NeurIPS = {NeurIPS})

@String(ICCV= {Int. Conf. Comput. Vis.})

@inproceedings{sohl2015deep,
  title={Deep unsupervised learning using nonequilibrium thermodynamics},
  author={Sohl-Dickstein, Jascha and Weiss, Eric and Maheswaranathan, Niru and Ganguli, Surya},
  booktitle={International conference on machine learning},
  pages={2256--2265},
  year={2015},
  organization={pmlr}
}

@article{bai2024blind,
  title={Blind inversion using latent diffusion priors},
  author={Bai, Weimin and Chen, Siyi and Chen, Wenzheng and Sun, He},
  journal={arXiv preprint arXiv:2407.01027},
  year={2024}
}

@article{luo2023diff,
  title={Diff-instruct: A universal approach for transferring knowledge from pre-trained diffusion models},
  author={Luo, Weijian and Hu, Tianyang and Zhang, Shifeng and Sun, Jiacheng and Li, Zhenguo and Zhang, Zhihua},
  journal={Advances in Neural Information Processing Systems},
  volume={36},
  pages={76525--76546},
  year={2023}
}

@article{unterthiner2019fvd,
  title={FVD: A new metric for video generation},
  author={Unterthiner, Thomas and Van Steenkiste, Sjoerd and Kurach, Karol and Marinier, Rapha{\"e}l and Michalski, Marcin and Gelly, Sylvain},
  year={2019}
}

@inproceedings{zhang2018unreasonable,
  title={The unreasonable effectiveness of deep features as a perceptual metric},
  author={Zhang, Richard and Isola, Phillip and Efros, Alexei A and Shechtman, Eli and Wang, Oliver},
  booktitle={Proceedings of the IEEE conference on computer vision and pattern recognition},
  pages={586--595},
  year={2018}
}

@article{zhuang2025flashvsr,
  title={FlashVSR: Towards Real-Time Diffusion-Based Streaming Video Super-Resolution},
  author={Zhuang, Junhao and Guo, Shi and Cai, Xin and Li, Xiaohui and Liu, Yihao and Yuan, Chun and Xue, Tianfan},
  journal={arXiv preprint arXiv:2510.12747},
  year={2025}
}

@article{wang2025seedvr2,
  title={Seedvr2: One-step video restoration via diffusion adversarial post-training},
  author={Wang, Jianyi and Lin, Shanchuan and Lin, Zhijie and Ren, Yuxi and Wei, Meng and Yue, Zongsheng and Zhou, Shangchen and Chen, Hao and Zhao, Yang and Yang, Ceyuan and others},
  journal={arXiv preprint arXiv:2506.05301},
  year={2025}
}

@article{brooks2024video,
  title={Video generation models as world simulators},
  author={Brooks, Tim and Peebles, Bill and Holmes, Connor and DePue, Will and Guo, Yufei and Jing, Li and Schnurr, David and Taylor, Joe and Luhman, Troy and Luhman, Eric and others},
  journal={OpenAI Blog},
  volume={1},
  number={8},
  pages={1},
  year={2024}
}

@article{polyak2024movie,
  title={Movie gen: A cast of media foundation models},
  author={Polyak, Adam and Zohar, Amit and Brown, Andrew and Tjandra, Andros and Sinha, Animesh and Lee, Ann and Vyas, Apoorv and Shi, Bowen and Ma, Chih-Yao and Chuang, Ching-Yao and others},
  journal={arXiv preprint arXiv:2410.13720},
  year={2024}
}

@article{zheng2024open,
  title={Open-sora: Democratizing efficient video production for all},
  author={Zheng, Zangwei and Peng, Xiangyu and Yang, Tianji and Shen, Chenhui and Li, Shenggui and Liu, Hongxin and Zhou, Yukun and Li, Tianyi and You, Yang},
  journal={arXiv preprint arXiv:2412.20404},
  year={2024}
}

@article{yeh2024diffir2vr,
  title={Diffir2vr-zero: Zero-shot video restoration with diffusion-based image restoration models},
  author={Yeh, Chang-Han and Lin, Chin-Yang and Wang, Zhixiang and Hsiao, Chi-Wei and Chen, Ting-Hsuan and Shiu, Hau-Shiang and Liu, Yu-Lun},
  journal={arXiv preprint arXiv:2407.01519},
  year={2024}
}

@inproceedings{nah2019ntire,
  title={Ntire 2019 challenge on video deblurring and super-resolution: Dataset and study},
  author={Nah, Seungjun and Baik, Sungyong and Hong, Seokil and Moon, Gyeongsik and Son, Sanghyun and Timofte, Radu and Mu Lee, Kyoung},
  booktitle={Proceedings of the IEEE/CVF conference on computer vision and pattern recognition workshops},
  pages={0--0},
  year={2019}
}

@article{huang2024flow,
  title={Flow generator matching},
  author={Huang, Zemin and Geng, Zhengyang and Luo, Weijian and Qi, Guo-jun},
  journal={arXiv preprint arXiv:2410.19310},
  year={2024}
}

@article{cheng2025leanvae,
  title={LeanVAE: An Ultra-Efficient Reconstruction VAE for Video Diffusion Models},
  author={Cheng, Yu and Yuan, Fajie},
  journal={arXiv preprint arXiv:2503.14325},
  year={2025}
}

@article{lin2024open,
  title={Open-Sora Plan: Open-Source Large Video Generation Model},
  author={Lin, Bin and Ge, Yunyang and Cheng, Xinhua and Li, Zongjian and Zhu, Bin and Wang, Shaodong and He, Xianyi and Ye, Yang and Yuan, Shenghai and Chen, Liuhan and others},
  journal={arXiv preprint arXiv:2412.00131},
  year={2024}
}

@article{wang2024integrating,
  title={Integrating amortized inference with diffusion models for learning clean distribution from corrupted images},
  author={Wang, Yifei and Bai, Weimin and Luo, Weijian and Chen, Wenzheng and Sun, He},
  journal={arXiv preprint arXiv:2407.11162},
  year={2024}
}

@inproceedings{yin2024one,
  title={One-step diffusion with distribution matching distillation},
  author={Yin, Tianwei and Gharbi, Micha{\"e}l and Zhang, Richard and Shechtman, Eli and Durand, Fredo and Freeman, William T and Park, Taesung},
  booktitle={Proceedings of the IEEE/CVF conference on computer vision and pattern recognition},
  pages={6613--6623},
  year={2024}
}

@article{yin2024slow,
  title={From slow bidirectional to fast autoregressive video diffusion models},
  author={Yin, Tianwei and Zhang, Qiang and Zhang, Richard and Freeman, William T and Durand, Fredo and Shechtman, Eli and Huang, Xun},
  journal={arXiv preprint arXiv:2412.07772},
  volume={2},
  year={2024}
}

@inproceedings{zhou2024score,
  title={Score identity distillation: Exponentially fast distillation of pretrained diffusion models for one-step generation},
  author={Zhou, Mingyuan and Zheng, Huangjie and Wang, Zhendong and Yin, Mingzhang and Huang, Hai},
  booktitle={Forty-first International Conference on Machine Learning},
  year={2024}
}

@inproceedings{kwon2025solving,
    title={Solving Video Inverse Problems Using Image Diffusion Models},
    author={Taesung Kwon and Jong Chul Ye},
    booktitle={The Thirteenth International Conference on Learning Representations},
    year={2025},
    url={https://openreview.net/forum?id=TRWxFUzK9K}
}

@article{wan2.1,
    title   = {Wan: Open and Advanced Large-Scale Video Generative Models},
    author  = {Wan Team},
    journal = {},
    year    = {2025}
}

@inproceedings{lee2024diffusion,
  title={Diffusion prior-based amortized variational inference for noisy inverse problems},
  author={Lee, Sojin and Park, Dogyun and Kong, Inho and Kim, Hyunwoo J},
  booktitle={European Conference on Computer Vision},
  pages={288--304},
  year={2024},
  organization={Springer}
}

@article{wan2025,
      title={Wan: Open and Advanced Large-Scale Video Generative Models}, 
      author={Team Wan and Ang Wang and Baole Ai and Bin Wen and Chaojie Mao and Chen-Wei Xie and Di Chen and Feiwu Yu and Haiming Zhao and Jianxiao Yang and Jianyuan Zeng and Jiayu Wang and Jingfeng Zhang and Jingren Zhou and Jinkai Wang and Jixuan Chen and Kai Zhu and Kang Zhao and Keyu Yan and Lianghua Huang and Mengyang Feng and Ningyi Zhang and Pandeng Li and Pingyu Wu and Ruihang Chu and Ruili Feng and Shiwei Zhang and Siyang Sun and Tao Fang and Tianxing Wang and Tianyi Gui and Tingyu Weng and Tong Shen and Wei Lin and Wei Wang and Wei Wang and Wenmeng Zhou and Wente Wang and Wenting Shen and Wenyuan Yu and Xianzhong Shi and Xiaoming Huang and Xin Xu and Yan Kou and Yangyu Lv and Yifei Li and Yijing Liu and Yiming Wang and Yingya Zhang and Yitong Huang and Yong Li and You Wu and Yu Liu and Yulin Pan and Yun Zheng and Yuntao Hong and Yupeng Shi and Yutong Feng and Zeyinzi Jiang and Zhen Han and Zhi-Fan Wu and Ziyu Liu},
      journal = {arXiv preprint arXiv:2503.20314},
      year={2025}
}

@article{chang2025warped,
  title={How i warped your noise: a temporally-correlated noise prior for diffusion models},
  author={Chang, Pascal and Tang, Jingwei and Gross, Markus and Azevedo, Vinicius C},
  journal={arXiv preprint arXiv:2504.03072},
  year={2025}
}

@inproceedings{zou2025flair,
  title={Flair: A conditional diffusion framework with applications to face video restoration},
  author={Zou, Zihao and Liu, Jiaming and Shoushtari, Shirin and Wang, Yubo and Kamilov, Ulugbek S},
  booktitle={2025 IEEE/CVF Winter Conference on Applications of Computer Vision (WACV)},
  pages={5228--5238},
  year={2025},
  organization={IEEE}
}

@misc{kwon2024visionxlhighdefinitionvideo,
      title={VISION-XL: High Definition Video Inverse Problem Solver using Latent Image Diffusion Models}, 
      author={Taesung Kwon and Jong Chul Ye},
      year={2024},
      eprint={2412.00156},
      archivePrefix={arXiv},
      primaryClass={cs.CV},
      url={https://arxiv.org/abs/2412.00156}, 
}

@article{mardani2023variational,
  title={A variational perspective on solving inverse problems with diffusion models},
  author={Mardani, Morteza and Song, Jiaming and Kautz, Jan and Vahdat, Arash},
  journal={arXiv preprint arXiv:2305.04391},
  year={2023}
}

@article{rout2023solving,
  title={Solving linear inverse problems provably via posterior sampling with latent diffusion models},
  author={Rout, Litu and Raoof, Negin and Daras, Giannis and Caramanis, Constantine and Dimakis, Alex and Shakkottai, Sanjay},
  journal={Advances in Neural Information Processing Systems},
  volume={36},
  pages={49960--49990},
  year={2023}
}

@article{song2023solving,
  title={Solving inverse problems with latent diffusion models via hard data consistency},
  author={Song, Bowen and Kwon, Soo Min and Zhang, Zecheng and Hu, Xinyu and Qu, Qing and Shen, Liyue},
  journal={arXiv preprint arXiv:2307.08123},
  year={2023}
}

@article{zilberstein2024repulsive,
  title={Repulsive Latent Score Distillation for Solving Inverse Problems},
  author={Zilberstein, Nicolas and Mardani, Morteza and Segarra, Santiago},
  journal={arXiv preprint arXiv:2406.16683},
  year={2024}
}

@article{zhang2025step,
  title={Step: A general and scalable framework for solving video inverse problems with spatiotemporal diffusion priors},
  author={Zhang, Bingliang and Wu, Zihui and Feng, Berthy T and Song, Yang and Yue, Yisong and Bouman, Katherine L},
  journal={arXiv preprint arXiv:2504.07549},
  year={2025}
}

@inproceedings{daras2024warped,
title={Warped Diffusion: Solving Video Inverse Problems with Image Diffusion Models},
author={Giannis Daras and Weili Nie and Karsten Kreis and Alex Dimakis and Morteza Mardani and Nikola Borislavov Kovachki and Arash Vahdat},
booktitle={The Thirty-eighth Annual Conference on Neural Information Processing Systems},
year={2024},
url={https://openreview.net/forum?id=LH94zPv8cu}
}

@inproceedings{rombach2022high,
  title={High-resolution image synthesis with latent diffusion models},
  author={Rombach, Robin and Blattmann, Andreas and Lorenz, Dominik and Esser, Patrick and Ommer, Bj{\"o}rn},
  booktitle={Proceedings of the IEEE/CVF conference on computer vision and pattern recognition},
  pages={10684--10695},
  year={2022}
}

@article{chung2022diffusion,
  title={Diffusion posterior sampling for general noisy inverse problems},
  author={Chung, Hyungjin and Kim, Jeongsol and Mccann, Michael T and Klasky, Marc L and Ye, Jong Chul},
  journal={arXiv preprint arXiv:2209.14687},
  year={2022}
}

@article{song2020score,
  title={Score-based generative modeling through stochastic differential equations},
  author={Song, Yang and Sohl-Dickstein, Jascha and Kingma, Diederik P and Kumar, Abhishek and Ermon, Stefano and Poole, Ben},
  journal={arXiv preprint arXiv:2011.13456},
  year={2020}
}

@article{song2019generative,
  title={Generative modeling by estimating gradients of the data distribution},
  author={Song, Yang and Ermon, Stefano},
  journal={Advances in neural information processing systems},
  volume={32},
  year={2019}
}

@inproceedings{ho2020denoising,
  title={Denoising diffusion probabilistic models},
  author={Ho, Jonathan and Jain, Ajay and Abbeel, Pieter},
  booktitle={NeurIPS},
  volume={33},
  pages={6840--6851},
  year={2020}
}

@inproceedings{peebles2023scalable,
  title={Scalable diffusion models with transformers},
  author={Peebles, William and Xie, Saining},
  booktitle={ICCV},
  pages={4195--4205},
  year={2023}
}
}

\clearpage
\setcounter{page}{1}
\setcounter{section}{0}
\setcounter{figure}{0}
\setcounter{table}{0}
\maketitlesupplementary

\section{Implementation Details}
\label{sec:suppl:impl}

We implement {InstantViR} based on the pre-trained {Wan2.1-1.3B} text-to-video model~\cite{wan2.1}. 
The student solver is initialized with the teacher's weights and fine-tuned using our proposed amortized distillation objective. 
Our framework supports both the original WanVAE~\cite{wan2.1} and the accelerated LeanVAE~\cite{cheng2025leanvae}. 
The training process utilizes the measurement consistency loss (likelihood) to enforce data fidelity and the distribution matching distillation (prior) loss~\cite{yin2024one} to inherit the generative prior. 
We utilize the Open-Sora dataset~\cite{lin2024open} solely as a source of raw videos to synthesize degraded measurements $\boldsymbol{y}$. 
The ground-truth clean videos $\boldsymbol{x}$ are never used in the loss computation; the student learns to reconstruct $\boldsymbol{x}$ solely through the guidance of the likelihood term and the teacher prior.
Below, we detail the specific configurations for operators, baselines, and hyperparameters.

\paragraph{Forward Operators.}
We evaluate our method on three standard video inverse problems, where the degradation operators $\mathcal{A}(\cdot)$ are implemented as follows:
\begin{itemize}
    \item \textbf{Random Inpainting:} We apply random binary masks with a masking ratio of {50\%}. The masks are applied in the latent space for training efficiency.
    \item \textbf{Gaussian Deblurring:} We apply a spatial Gaussian blur kernel with a size of $61 \times 61$ and a standard deviation $\sigma=3.0$ to each video frame.
    \item \textbf{4$\times$ Super-Resolution:} We perform anti-aliased downsampling with a factor of 4. Specifically, the high-resolution video is downsampled in pixel space using a Resizer, and the resulting low-resolution video is then encoded back into the latent space to serve as the model input.
\end{itemize}

\paragraph{Baselines.}
For image-based diffusion baselines such as DPS~\cite{chung2022diffusion} and SVI~\cite{kwon2025solving}, due to memory constraints and their inherent image-processing nature, the visual results are generated by processing patches ($256\times256$) and stitching them together. 
All other video-based methods~\cite{kwon2024visionxlhighdefinitionvideo} process frames or blocks directly.

\paragraph{Hyperparameters.}
To ensure stable training and efficient convergence, our configuration is based on the established settings of CausVid~\cite{yin2024slow}. 
We utilize the AdamW optimizer with a fixed learning rate of $2 \times 10^{-6}$ and apply gradient clipping with a threshold of 1.0 to mitigate potential instability during the distillation process.
The distribution matching distillation~\cite{yin2024one} (prior) loss incorporates a warmup phase of 1,000 steps. 
Consistent with the teacher model's distillation requirements, we set the timestep shift parameter to 8.0, the real guidance scale to 3.5, and the fake generation update ratio to 5, while processing temporal data with a block size of 3 frames. 

Regarding the objective function, the weight of the measurement consistency loss (likelihood) is tuned specifically for the difficulty of each degradation: it is set to 0.1 for random inpainting, 1.0 for Gaussian deblurring, and 0.3 for 4$\times$ super-resolution. 
Uniquely for the inpainting task, we compute the prior loss exclusively on the masked regions to force the student to focus its generative capacity on hallucinating the missing content rather than reconstructing visible pixels. 
Finally, computational settings are adjusted based on the VAE employed; we use a batch size of 2 for the standard WanVAE~\cite{wan2.1}, which we increase to 4 when utilizing the more memory-efficient LeanVAE~\cite{cheng2025leanvae}. 
For the latter, the latent shape is explicitly configured as $1 \times 21 \times 16 \times 60 \times 104$ to correspond with the high-resolution target outputs.

\begin{algorithm*}[t]
\caption{InstantViR: Training Pipeline}
\label{alg:training}
\begin{algorithmic}[1]
\Require Dataset of raw videos (unpaired), frozen teacher diffusion model $s_\theta$ (Wan2.1), teacher VAE $\mathcal{E}/\mathcal{D}$, student VAE $\mathcal{E}'/\mathcal{D}'$ (Wan or Lean), degradation operator $\mathcal{A}$, student solver $q_\phi$, auxiliary score network $s_{\varphi}$.
\State Initialize student parameters $\phi \leftarrow \theta$ (if architectures match) or custom init. Initialize $\varphi$.
\State Freeze teacher $s_\theta$ and VAEs.
\While{not converged}
    \State Sample batch $\boldsymbol{x}_{raw}$; Generate measurement $\boldsymbol{y} = \mathcal{A}(\boldsymbol{x}_{raw}) + \boldsymbol{n}$; Encode $\boldsymbol{z}_{\text{in}} = \mathcal{E}'(\boldsymbol{y})$.
    
    \State \textbf{// 1. Student Prediction}
    \State Predict clean latent: $\hat{\boldsymbol{z}}_0 = q_\phi(\boldsymbol{z}_{\text{in}})$. \Comment{Latent in Student space}
    
    \State \textbf{// 2. Likelihood Term (Measurement Consistency)}
    \State Decode to pixel: $\hat{\boldsymbol{x}}_0 = \mathcal{D}'(\hat{\boldsymbol{z}}_0)$.
    \State $\mathcal{L}_{\text{data}} = \|\boldsymbol{y} - \mathcal{A}(\hat{\boldsymbol{x}}_0)\|^2$.
    
    \State \textbf{// 3. Prior Term (DMD with Latent Alignment)}
    \State Sample $t, \boldsymbol{\epsilon}$.
    \If{Using LeanVAE ($\mathcal{E}' \neq \mathcal{E}$)}
        \State \textbf{Bridge to Teacher Space:} $\boldsymbol{z}_{target} = \mathcal{E}(\hat{\boldsymbol{x}}_0)$. \Comment{Diff. through LeanDec $\to$ WanEnc}
    \Else
        \State $\boldsymbol{z}_{target} = \hat{\boldsymbol{z}}_0$.
    \EndIf
    \State Diffuse target: $\boldsymbol{z}_t = \alpha_t \boldsymbol{z}_{target} + \sigma_t \boldsymbol{\epsilon}$. \Comment{Noise in Teacher space}
    \State Teacher score: $\boldsymbol{g}_\theta = s_\theta(\boldsymbol{z}_t, t, \text{text})$.
    \State Student score (aux): $\boldsymbol{g}_\phi = s_{\varphi}(\boldsymbol{z}_t, t)$. 
    \State $\mathcal{L}_{\text{prior}} = w(t) \| \boldsymbol{g}_\theta - \boldsymbol{g}_\phi \|^2$.
    
    \State \textbf{// 4. Updates}
    \State Update $\phi \leftarrow \phi - \eta_1 \nabla_\phi (\mathcal{L}_{\text{data}} + \lambda \mathcal{L}_{\text{prior}})$.
    \State Update $\varphi$ to match score of noised $\boldsymbol{z}_{target}$ (via Denoising Score Matching).
\EndWhile
\end{algorithmic}
\end{algorithm*}

\section{Algorithm}
\label{sec:suppl:algo}

We provide detailed pseudocodes for the training pipeline (Algorithm~\ref{alg:training}) and the efficient block-wise streaming inference (Algorithm~\ref{alg:inference}).

\paragraph{Training Protocol and VAE Alignment.}
Algorithm~\ref{alg:training} outlines the core training loop. A crucial implementation detail involves the handling of the VAE when training the accelerated version, {InstantViR$^\dag$}.
For the standard version, the student $q_\phi$ and teacher $s_\theta$ share the same {WanVAE}~\cite{wan2.1} latent space. 
However, when training {InstantViR$^\dag$} to utilize the ultra-efficient {LeanVAE}~\cite{cheng2025leanvae}, a latent space mismatch arises: the student $q_\phi$ predicts Lean-latents to maximize inference speed, while the frozen teacher $s_\theta$ (Wan2.1) requires Wan-latents to compute the score distillation loss ($\mathcal{L}_{\text{prior}}$).
Therefore, in the actual implementation of Algorithm~\ref{alg:training}, if the VAE is switched to LeanVAE, we perform an additional differentiable bridging step before teacher score computation. 
Specifically, the student's predicted latent $\hat{\boldsymbol{z}}_0$ is passed through the {LeanDecoder} to pixel space and immediately re-encoded via the frozen {WanEncoder}. This ensures the teacher provides valid guidance despite the student operating in a more efficient latent manifold.

\paragraph{Streaming Inference with KV Cache.}
Algorithm~\ref{alg:inference} details the deployment phase. 
Unlike standard diffusion sampling, which requires iterating over time $t$, our solver is a one-step feed-forward network ($t=0$). 
To achieve high throughput for long videos, we process the video in non-overlapping temporal blocks ($n=1 \dots N$).
Crucially, to satisfy the causal dependency without redundant computation, we maintain a running {Key-Value (KV) Cache} ($\mathcal{K}, \mathcal{V}$). 
As shown in lines 10-16, for each new block $\boldsymbol{y}^{(n)}$, we only compute the Query $\mathbf{Q}^{(n)}$ for the current frames, while retrieving the Keys and Values from the history ($\mathbf{K}_{\text{past}}, \mathbf{V}_{\text{past}}$) to perform attention. 
This reduces the complexity of the attention mechanism from quadratic with respect to total video length to linear, enabling theoretically infinite streaming generation.

\begin{algorithm*}[h]
\caption{InstantViR: Inference Pipeline}
\label{alg:inference}
\begin{algorithmic}[1]
\Require Degraded video stream $\boldsymbol{y}$, trained solver $q_\phi$, VAE decoder $\mathcal{D}$, block size $L$.
\State Initialize KV Cache: $\mathcal{K} \leftarrow \emptyset, \mathcal{V} \leftarrow \emptyset$.
\State Split $\boldsymbol{y}$ into temporal blocks $\{\boldsymbol{y}^{(1)}, \boldsymbol{y}^{(2)}, \dots, \boldsymbol{y}^{(N)}\}$.
\For{$n = 1$ to $N$}
    \State Encode measurement: $\boldsymbol{z}^{(n)}_{\text{in}} = \mathcal{E}(\boldsymbol{y}^{(n)})$ (or resize/embed).
    \State Set time step $t=0$ (deterministic inference).
    
    \State \textbf{// Causal Attention Block}
    \State $\mathbf{Q}^{(n)} = \text{Proj}_Q(\boldsymbol{z}^{(n)}_{\text{in}})$
    \State $\mathbf{K}^{(n)} = \text{Proj}_K(\boldsymbol{z}^{(n)}_{\text{in}}), \mathbf{V}^{(n)} = \text{Proj}_V(\boldsymbol{z}^{(n)}_{\text{in}})$
    
    \State \textit{Inter-block Attention (Read Cache):}
    \State Retrieve $\mathbf{K}_{\text{past}}, \mathbf{V}_{\text{past}}$ from $\mathcal{K}, \mathcal{V}$.
    \State $\mathbf{K}_{\text{full}} = [\mathbf{K}_{\text{past}}, \mathbf{K}^{(n)}], \mathbf{V}_{\text{full}} = [\mathbf{V}_{\text{past}}, \mathbf{V}^{(n)}]$.
    
    \State \textit{Compute Reconstruction:}
    \State $\hat{\boldsymbol{z}}^{(n)}_0 = \text{Attention}(\mathbf{Q}^{(n)}, \mathbf{K}_{\text{full}}, \mathbf{V}_{\text{full}})$.
    \State Pass through MLP and remaining layers of $q_\phi$.
    
    \State \textit{Update Cache:}
    \State Update $\mathcal{K} \leftarrow [\mathcal{K}, \mathbf{K}^{(n)}], \mathcal{V} \leftarrow [\mathcal{V}, \mathbf{V}^{(n)}]$.
    
    \State \textbf{// Stream Output}
    \State Decode block: $\hat{\boldsymbol{x}}^{(n)}_0 = \mathcal{D}(\hat{\boldsymbol{z}}^{(n)}_0)$.
    \State Yield $\hat{\boldsymbol{x}}^{(n)}_0$.
\EndFor
\end{algorithmic}
\end{algorithm*}

\section{Additional Results}
\label{sec:suppl:results}

We present extensive qualitative comparisons on three challenging video inverse problems: video inpainting, Gaussian deblurring, and 4$\times$ super-resolution. We compare InstantViR against strong diffusion-based baselines, including DPS~\cite{chung2022diffusion}, SVI~\cite{kwon2025solving}, and Vision-XL~\cite{kwon2024visionxlhighdefinitionvideo}. Additionally, we showcase the text-guided controllability of our model in restoration tasks.

\begin{figure*}[h]
    \centering
    \setlength{\tabcolsep}{1pt}
    \renewcommand{\arraystretch}{1.1}
    \resizebox{\linewidth}{!}{%
    \begin{tabular}{cccccc}
        & \multicolumn{5}{c}{
        \begin{tikzpicture}[baseline]
            \draw[->, >=latex, line width=0.35mm] (0,0.1) -- (15.2cm,0.1) 
            node[right, xshift=2mm] at (15.2cm, 0.1) {Time};
        \end{tikzpicture}
      } \\
        \rotatebox{90}{\small Masked Input} &
        \includegraphics[width=0.19\textwidth]{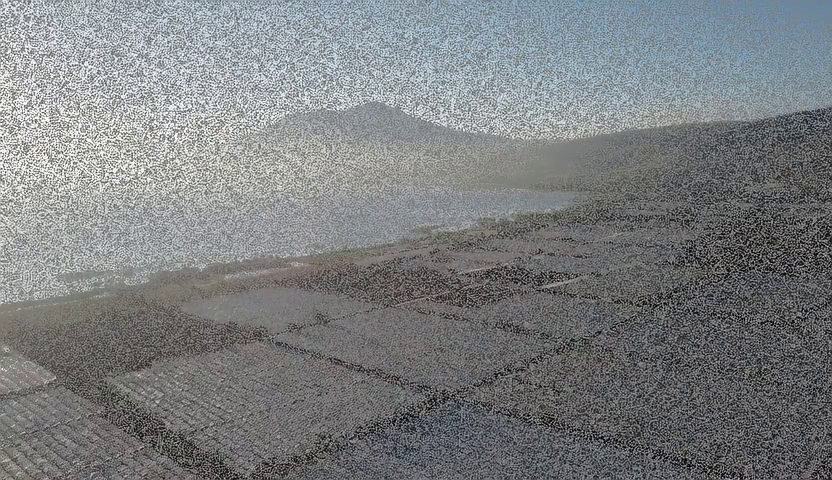} &
        \includegraphics[width=0.19\textwidth]{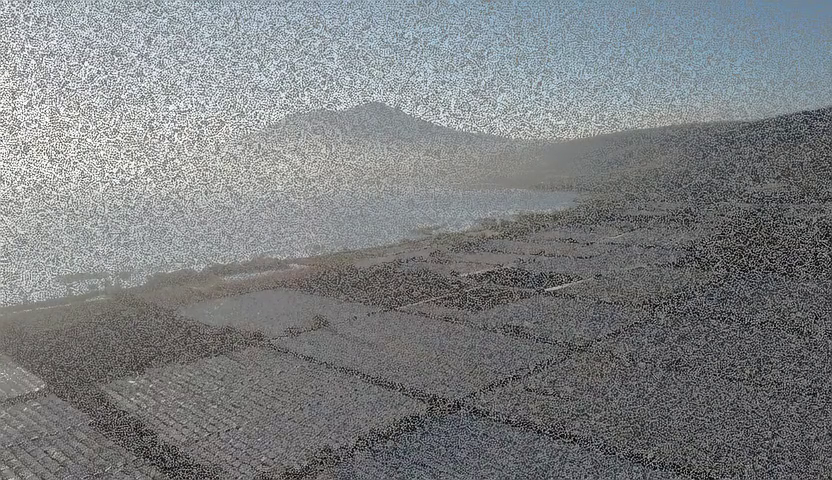} &
        \includegraphics[width=0.19\textwidth]{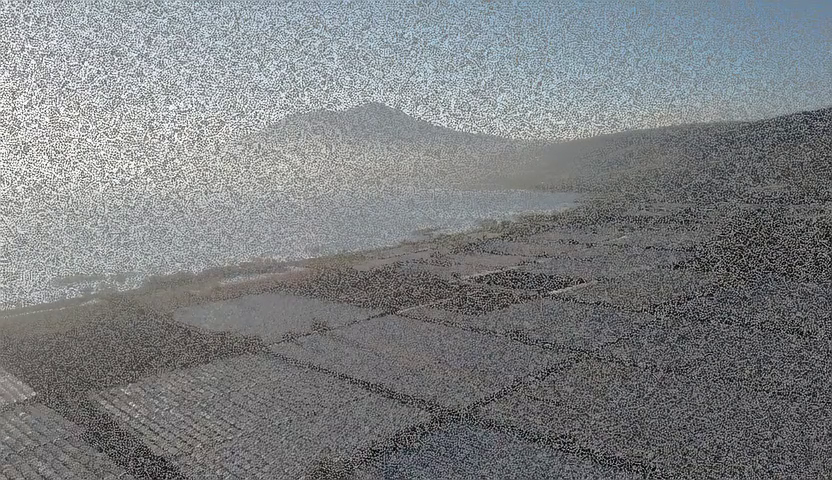} &
        \includegraphics[width=0.19\textwidth]{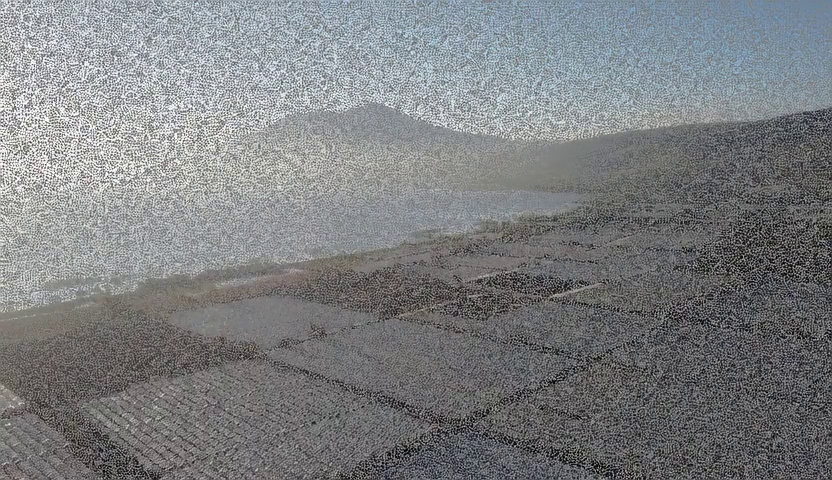} &
        \includegraphics[width=0.19\textwidth]{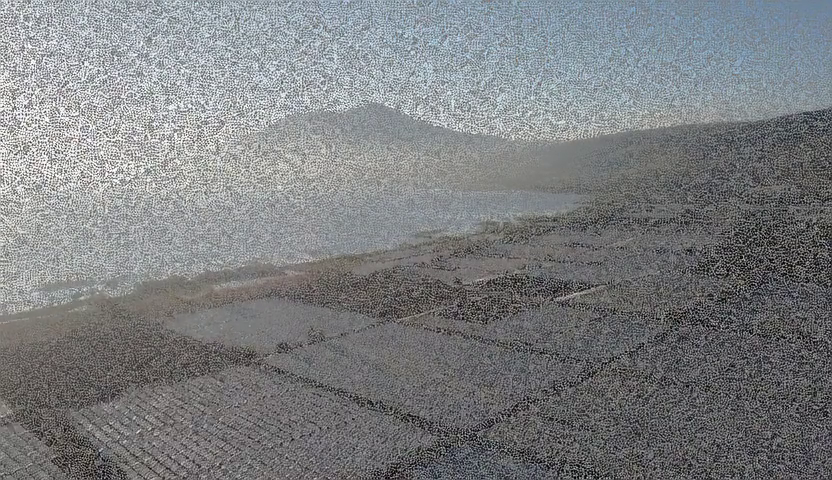} \\
        \rotatebox{90}{\,\,\,\,\,\,\,\,\small DPS~\cite{chung2022diffusion}} &
        \includegraphics[width=0.19\textwidth]{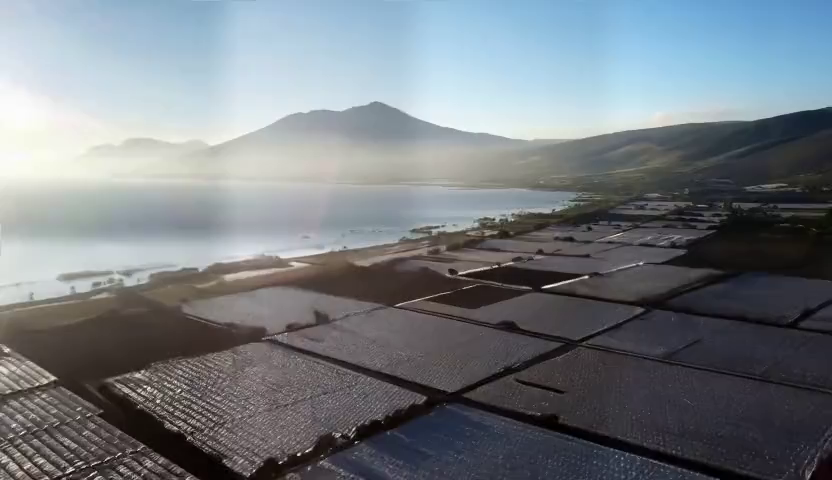} &
        \includegraphics[width=0.19\textwidth]{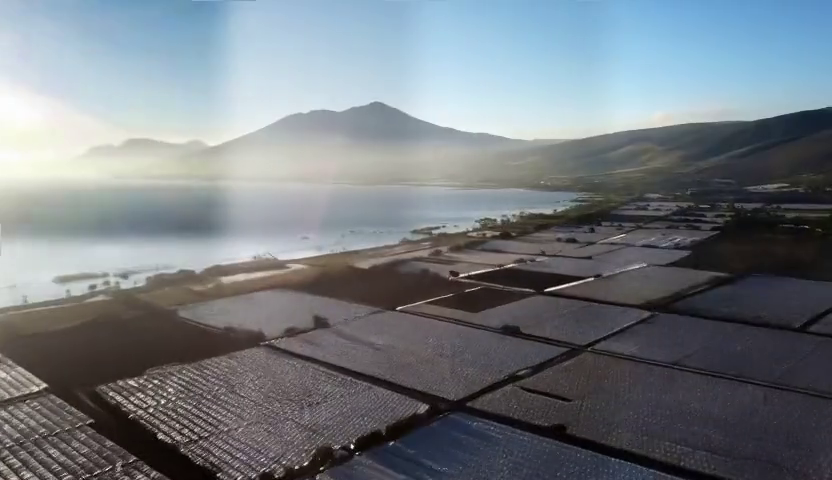} &
        \includegraphics[width=0.19\textwidth]{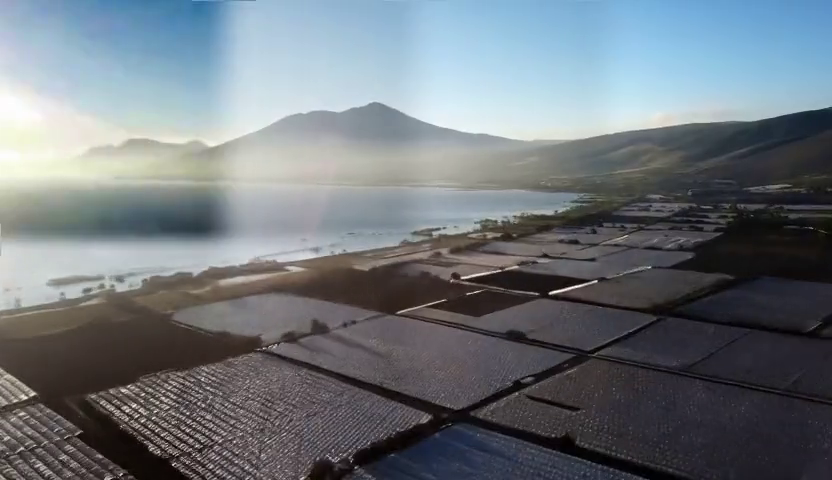} &
        \includegraphics[width=0.19\textwidth]{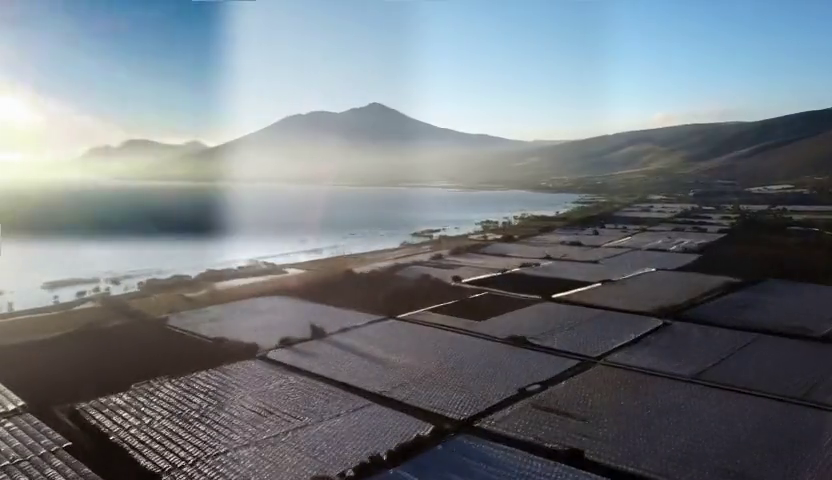} &
        \includegraphics[width=0.19\textwidth]{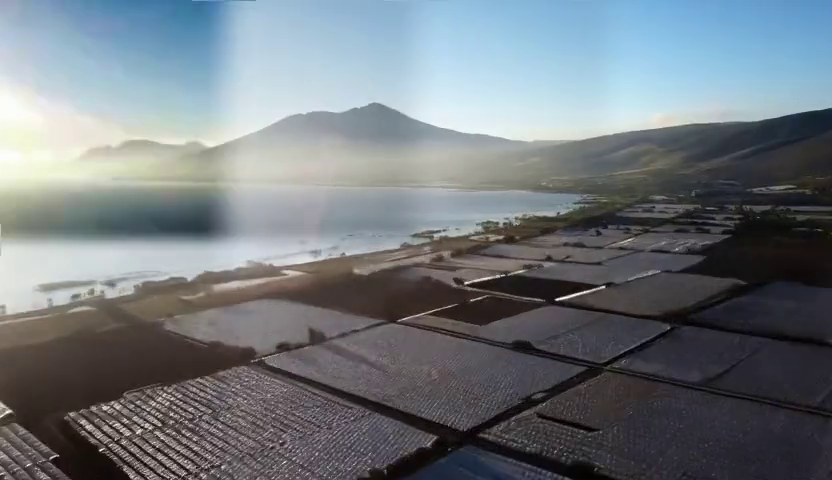} \\
        \rotatebox{90}{\,\,\,\,\,\,\,\,\small SVI~\cite{kwon2025solving}} &
        \includegraphics[width=0.19\textwidth]{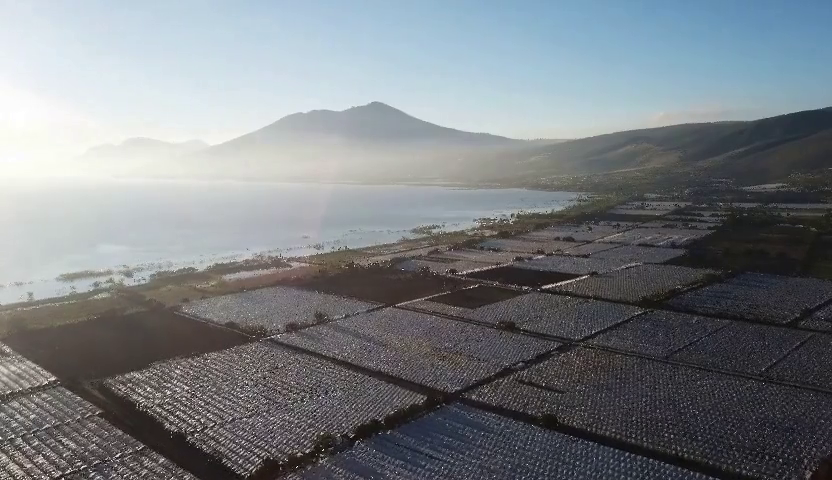} &
        \includegraphics[width=0.19\textwidth]{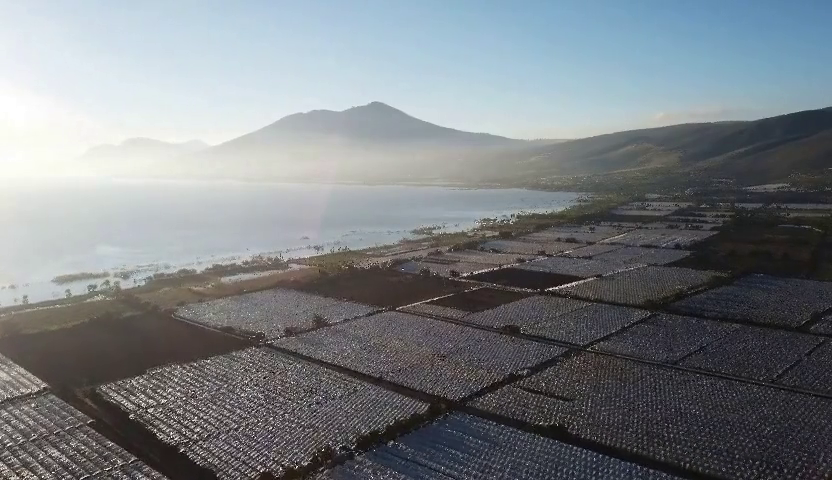} &
        \includegraphics[width=0.19\textwidth]{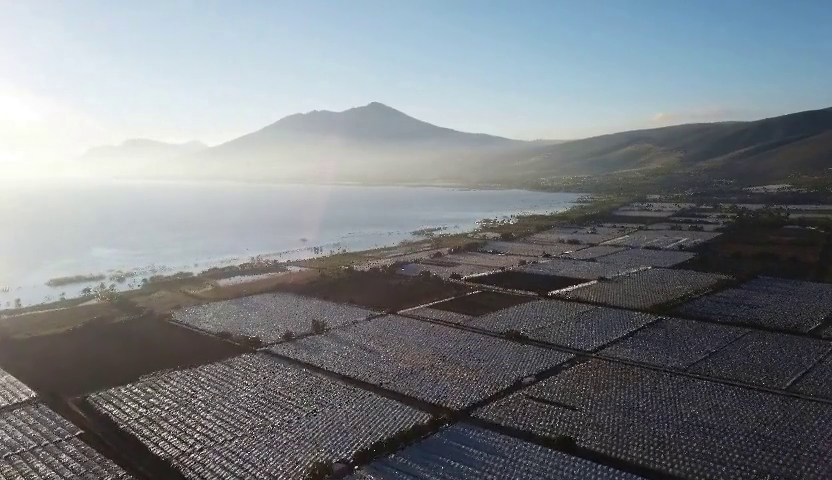} &
        \includegraphics[width=0.19\textwidth]{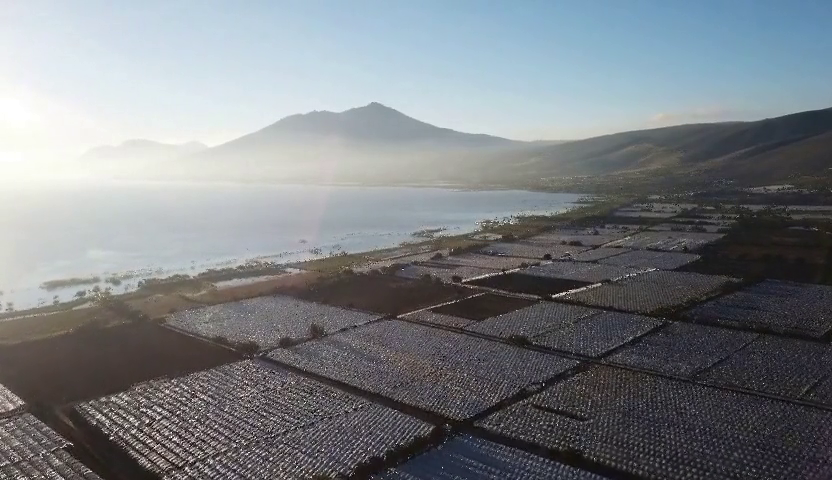} &
        \includegraphics[width=0.19\textwidth]{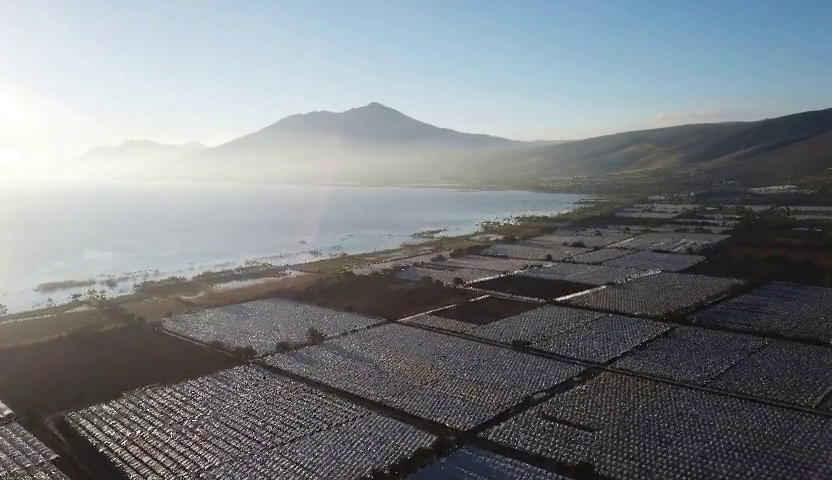} \\
        \rotatebox{90}{\small Vision-XL~\cite{kwon2024visionxlhighdefinitionvideo}} &
        \includegraphics[width=0.19\textwidth]{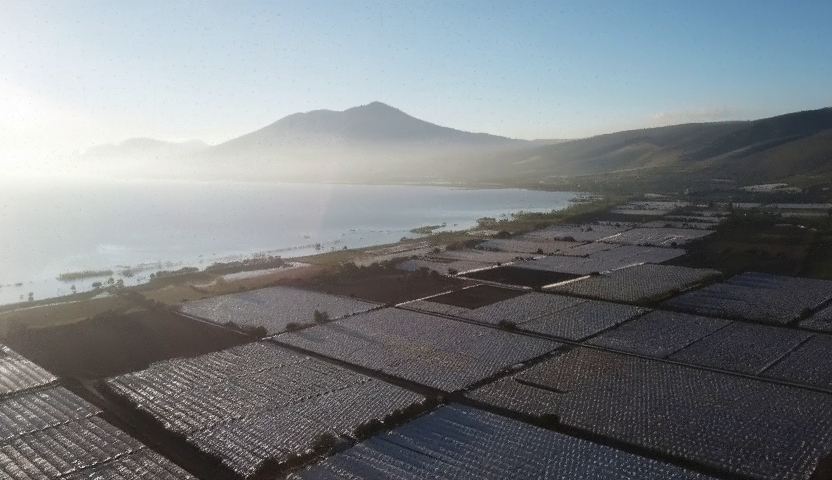} &
        \includegraphics[width=0.19\textwidth]{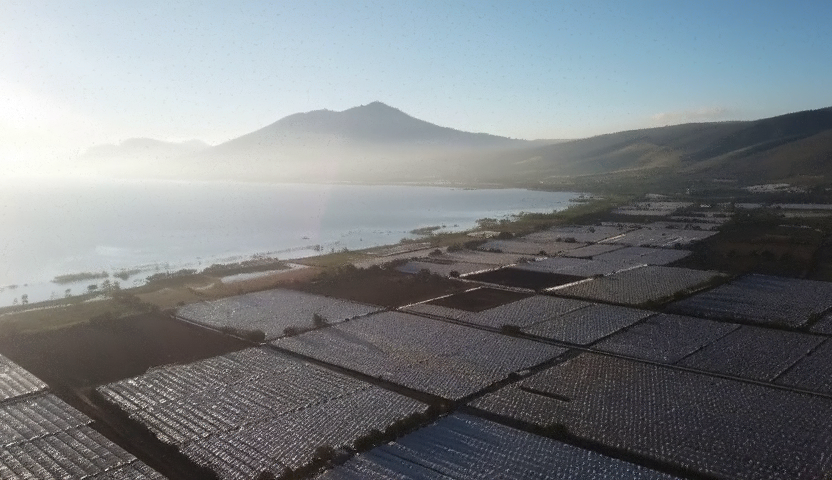} &
        \includegraphics[width=0.19\textwidth]{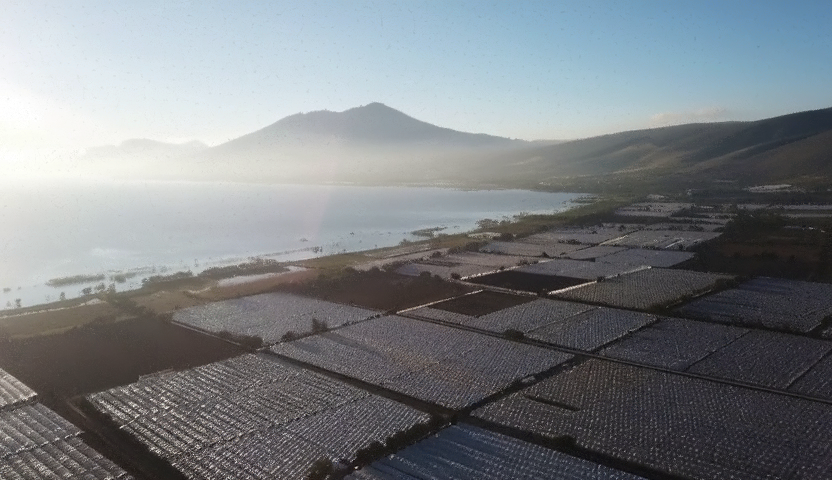} &
        \includegraphics[width=0.19\textwidth]{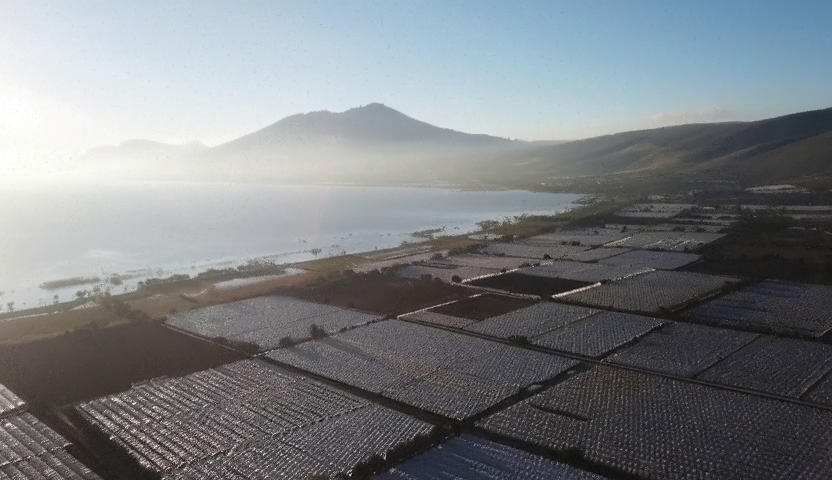} &
        \includegraphics[width=0.19\textwidth]{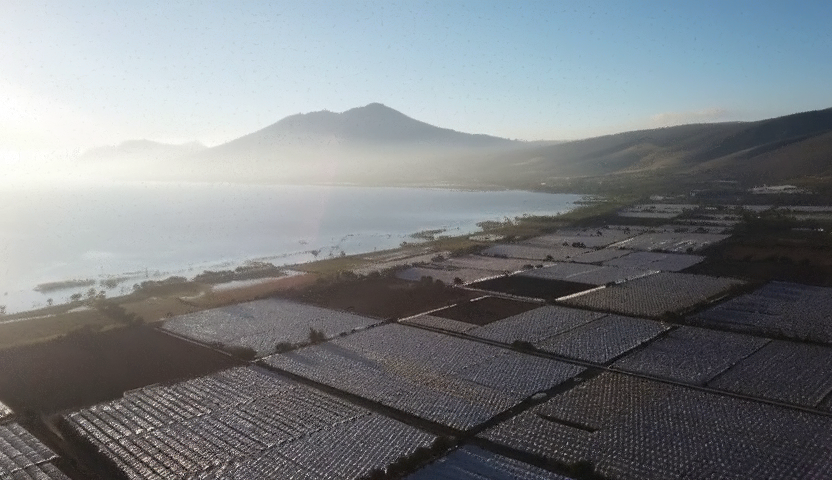} \\
        \rotatebox{90}{\,\,\,\,\,\,\,\,\,\,\small Ours} &
        \includegraphics[width=0.19\textwidth]{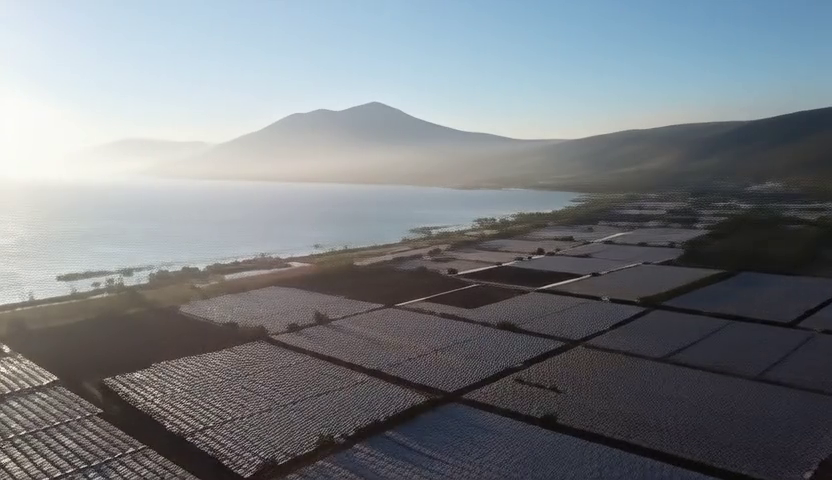} &
        \includegraphics[width=0.19\textwidth]{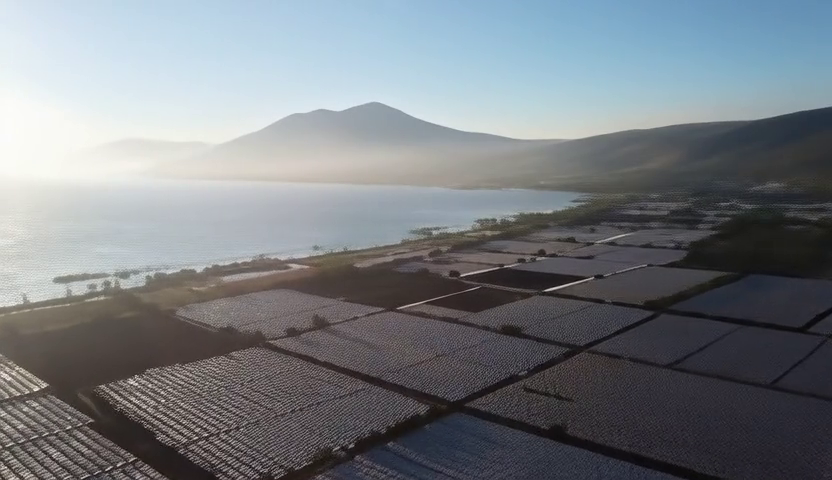} &
        \includegraphics[width=0.19\textwidth]{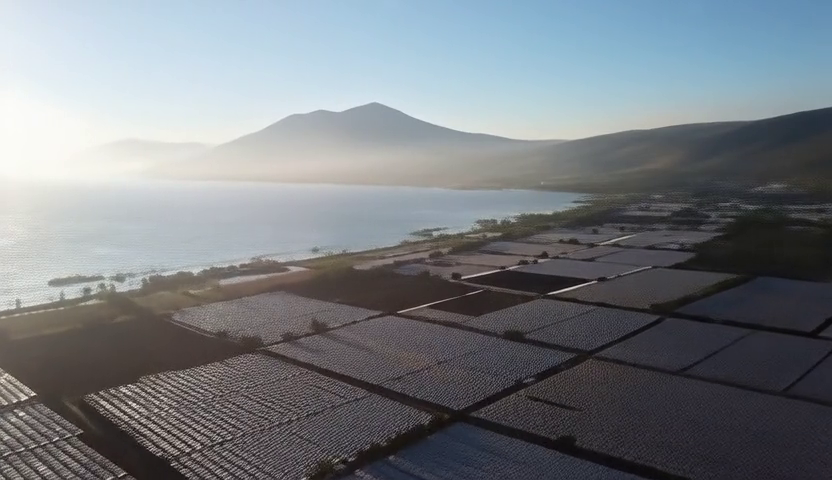} &
        \includegraphics[width=0.19\textwidth]{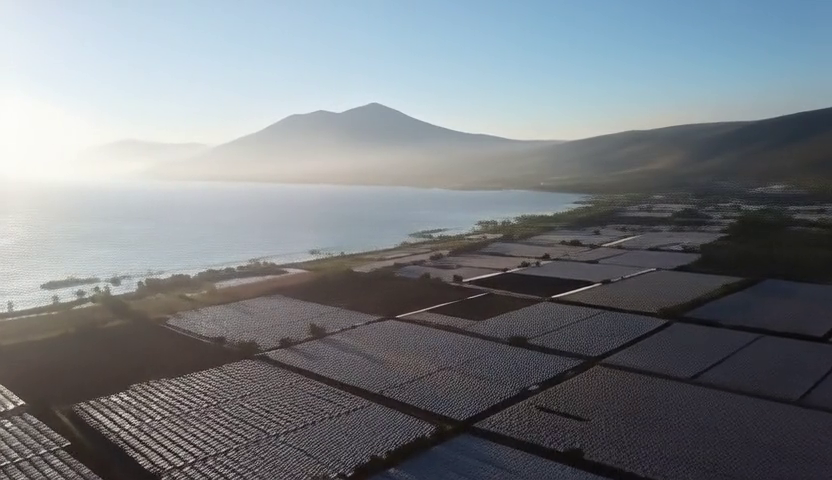} &
        \includegraphics[width=0.19\textwidth]{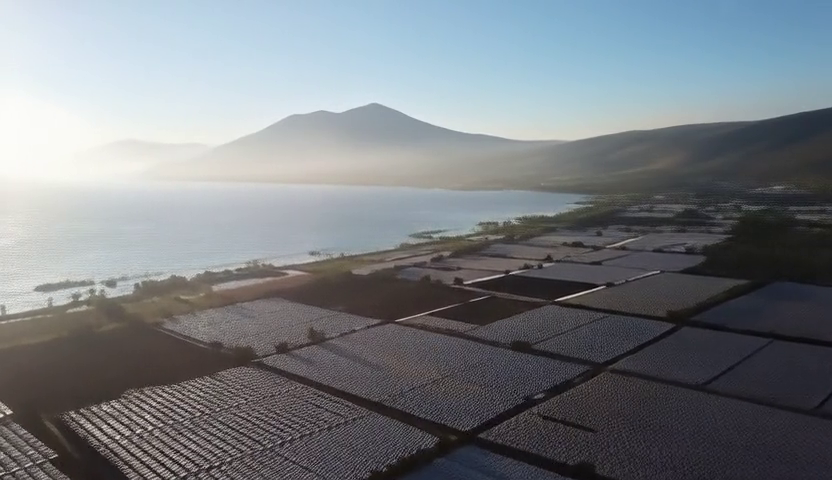} \\
        \rotatebox{90}{\,\,\,\,\,\,\,\,\,\,\small Ours$^\dag$} &
        \includegraphics[width=0.19\textwidth]{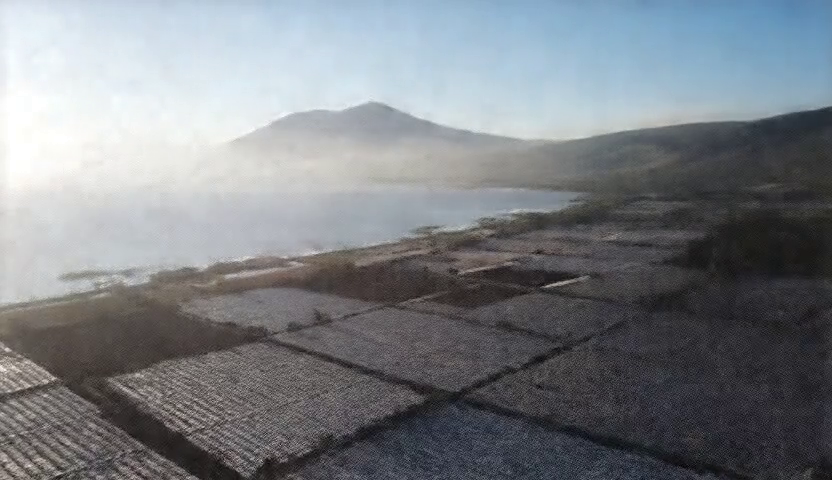} &
        \includegraphics[width=0.19\textwidth]{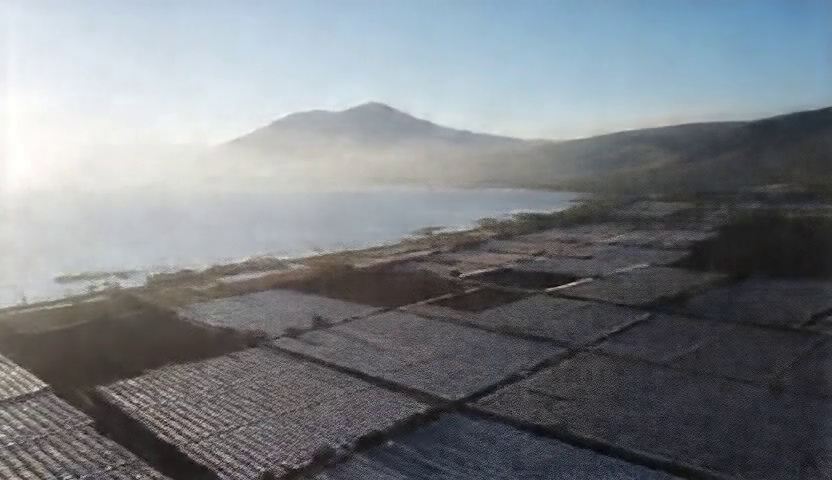} &
        \includegraphics[width=0.19\textwidth]{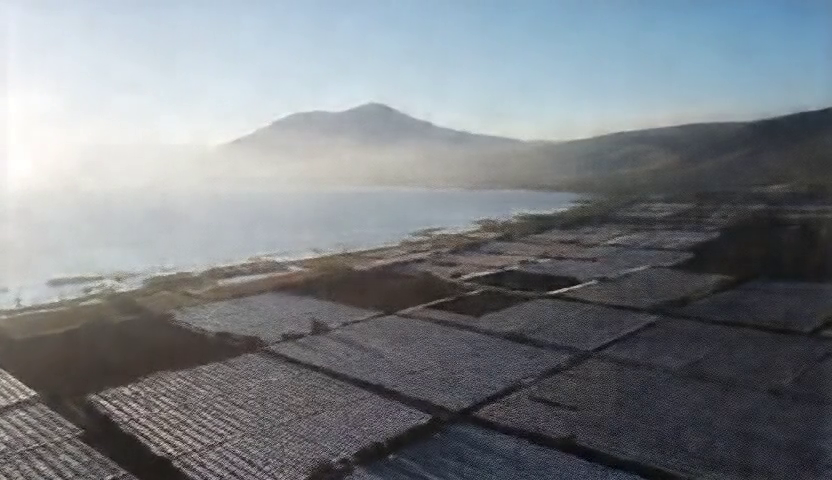} &
        \includegraphics[width=0.19\textwidth]{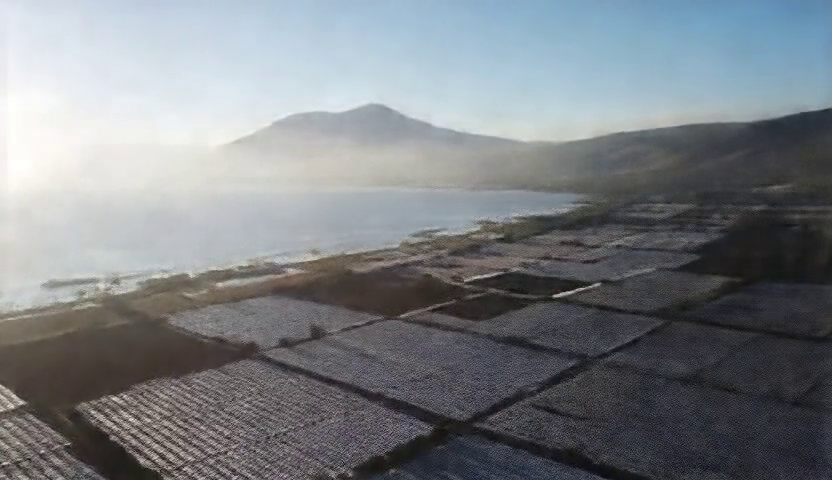} &
        \includegraphics[width=0.19\textwidth]{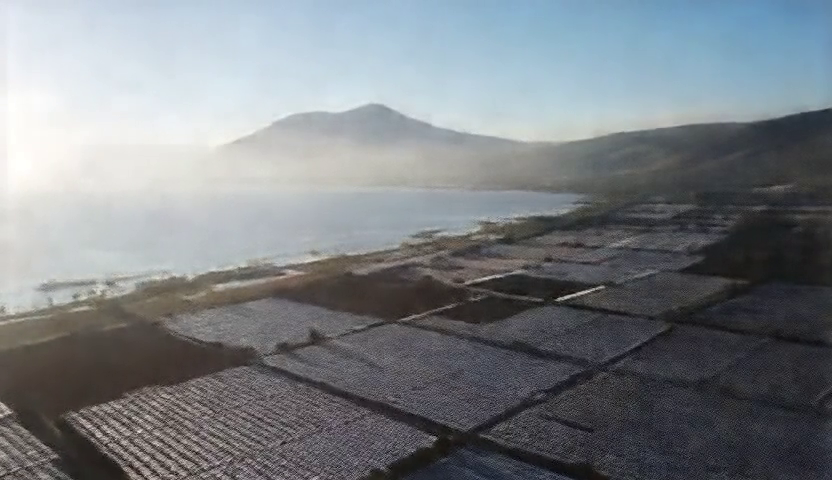} \\     
        \rotatebox{90}{\,\,\,\,\,\,\,\,\,\,\small GT} &
        \includegraphics[width=0.19\textwidth]{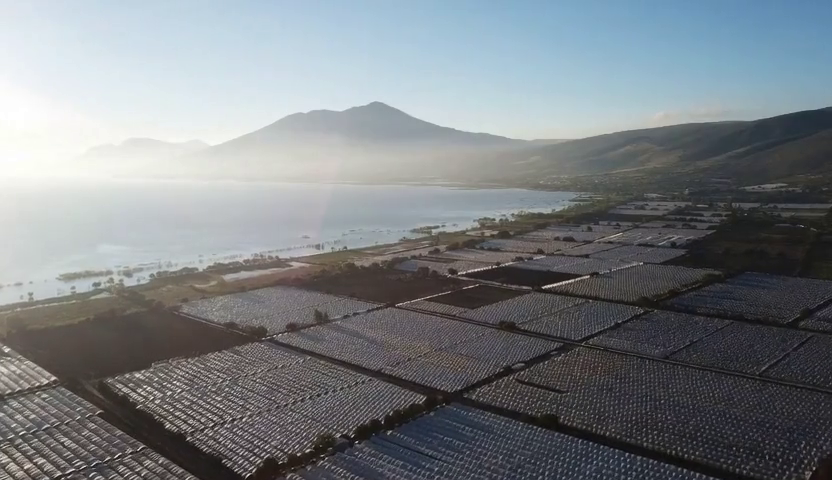} &
        \includegraphics[width=0.19\textwidth]{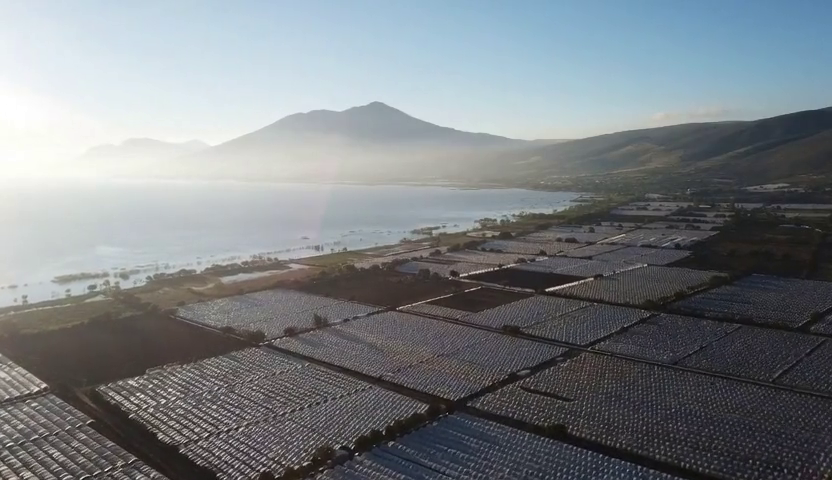} &
        \includegraphics[width=0.19\textwidth]{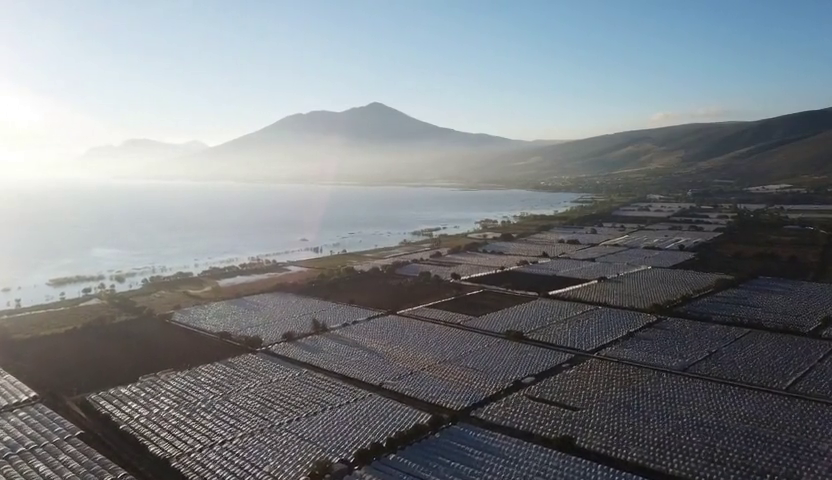} &
        \includegraphics[width=0.19\textwidth]{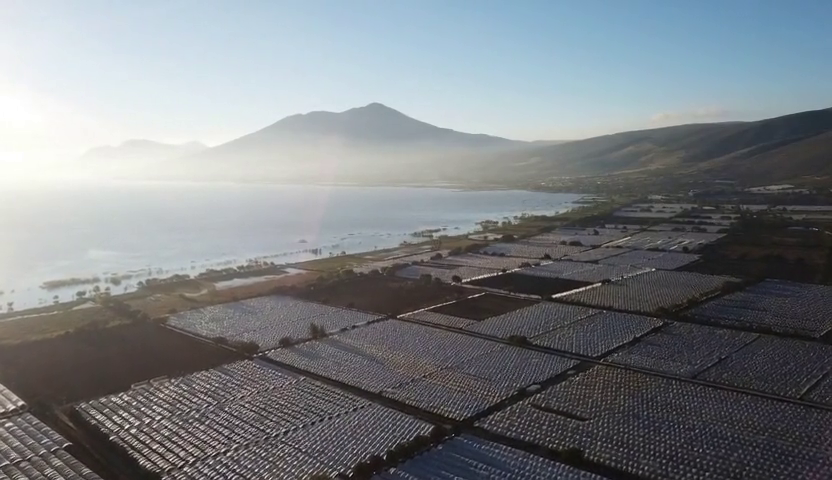} &
        \includegraphics[width=0.19\textwidth]{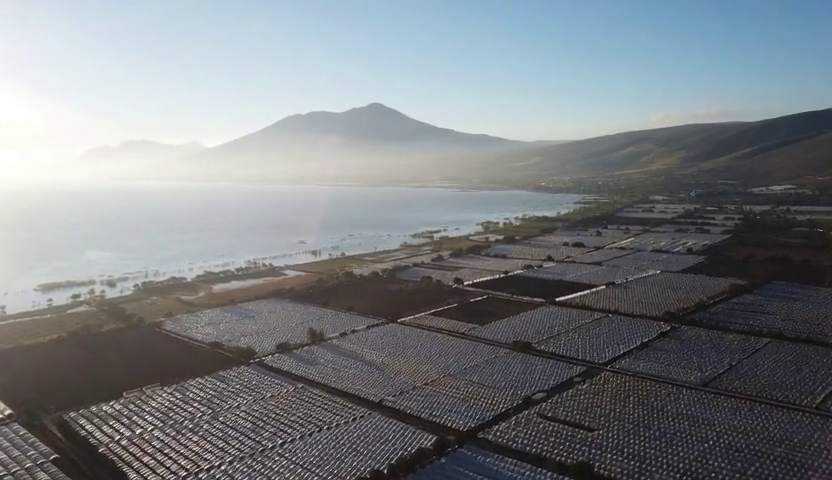} \\
    \end{tabular}
    }
    \caption{\textbf{Video Inpainting qualitative comparison.} Each row shows a complete sequence reconstructed by a specific method. InstantViR (both WanVAE \textbf{Ours} and LeanVAE \textbf{Ours$^\dag$} variants) produces coherent content for every frame while requiring only a single feed-forward pass.}
    \label{fig:suppl_inpainting}
\end{figure*}

\paragraph{Video Inpainting Comparison.}
Figure~\ref{fig:suppl_inpainting} compares our method against DPS, SVI, and Vision-XL on a random inpainting task (50\% mask). 
Sampling-based methods like DPS and SVI often struggle with consistency or produce blurry artifacts in large masked regions. Vision-XL produces higher quality but remains computationally expensive.
InstantViR (Ours) and its accelerated variant (Ours$^\dag$) achieve superior visual fidelity and temporal coherence in a single step, effectively hallucinating missing content that is consistent with the ground truth.

\begin{figure*}[h]
    \centering
    \setlength{\tabcolsep}{1pt}
    \renewcommand{\arraystretch}{1.1}
    \resizebox{\linewidth}{!}{%
    \begin{tabular}{cccccc}
        & \multicolumn{5}{c}{
        \begin{tikzpicture}[baseline]
            \draw[->, >=latex, line width=0.35mm] (0,0.1) -- (15.2cm,0.1) 
            node[right, xshift=2mm] at (15.2cm, 0.1) {Time};
        \end{tikzpicture}
      } \\
        \rotatebox{90}{\small Blurred Input} &
        \includegraphics[width=0.19\textwidth]{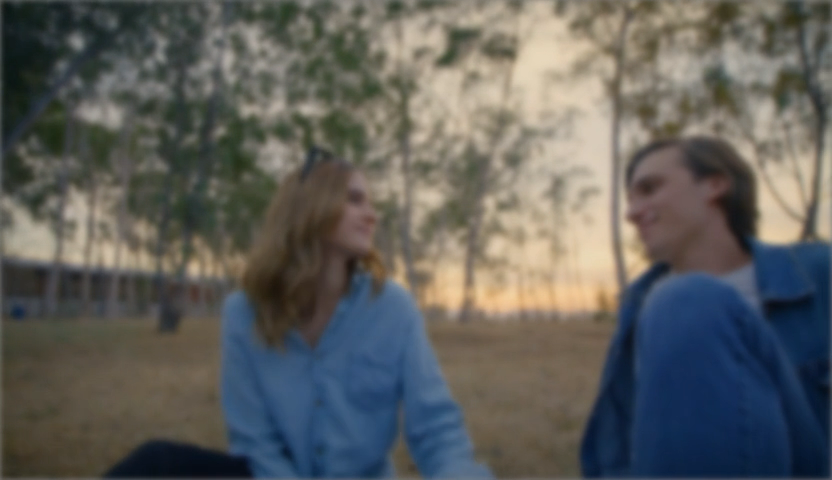} &
        \includegraphics[width=0.19\textwidth]{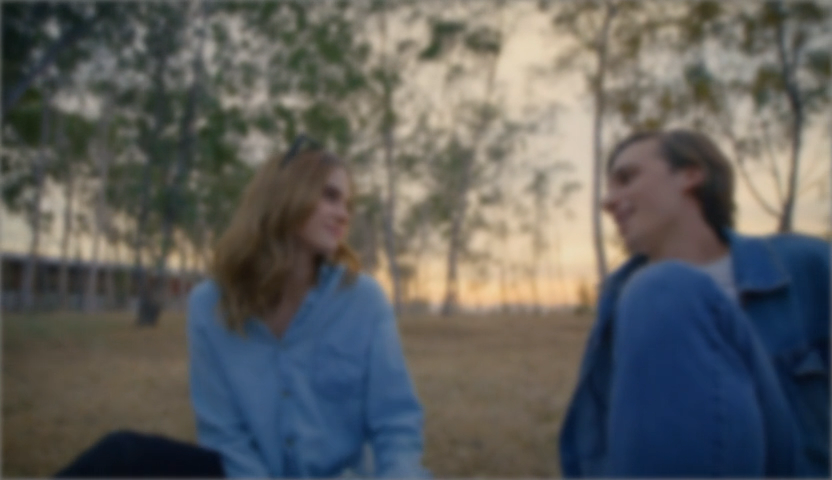} &
        \includegraphics[width=0.19\textwidth]{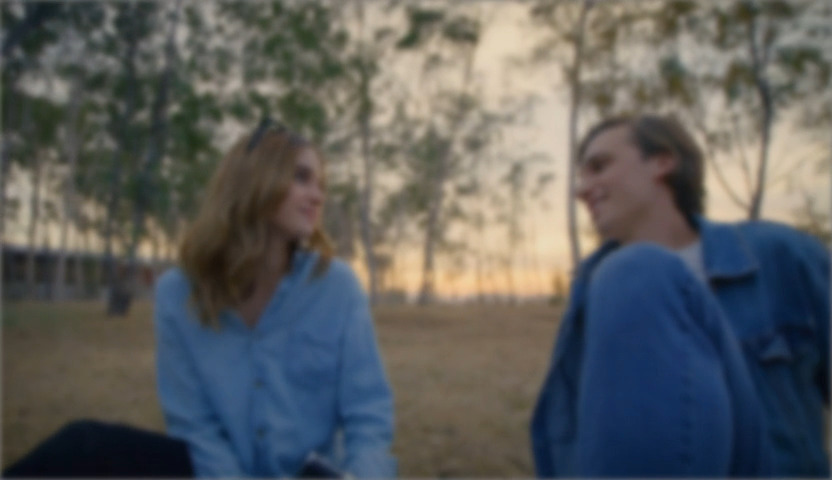} &
        \includegraphics[width=0.19\textwidth]{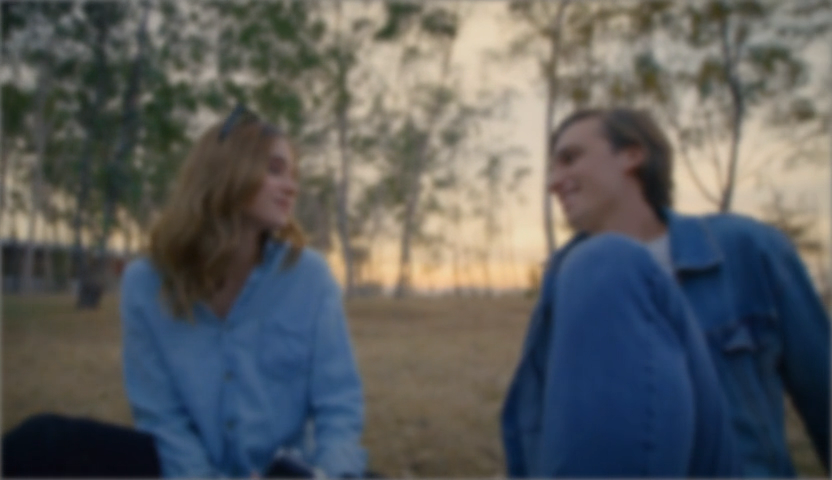} &
        \includegraphics[width=0.19\textwidth]{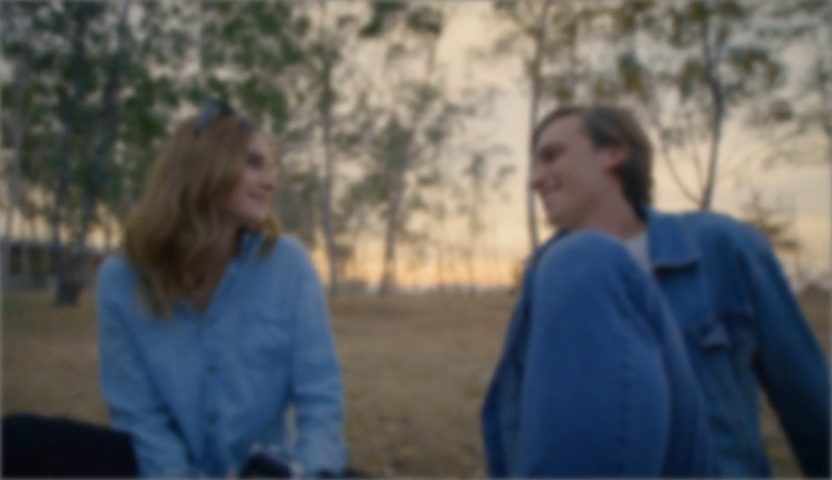} \\
        \rotatebox{90}{\,\,\,\,\,\,\,\,\small DPS~\cite{chung2022diffusion}} &
        \includegraphics[width=0.19\textwidth]{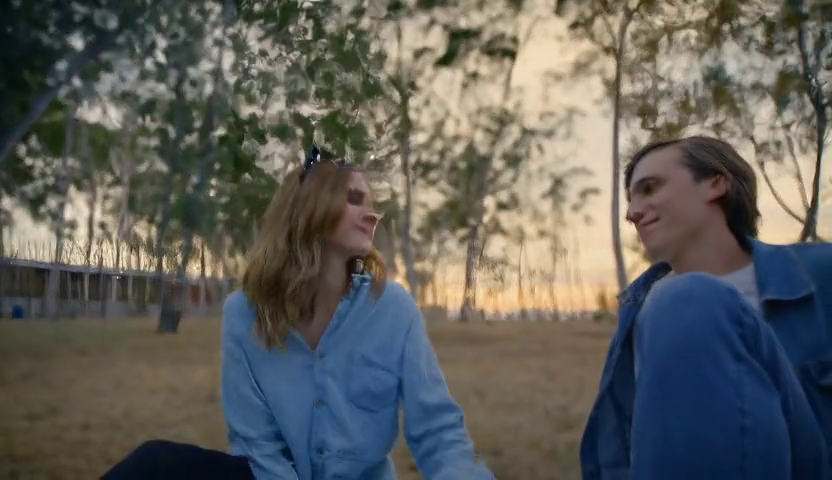} &
        \includegraphics[width=0.19\textwidth]{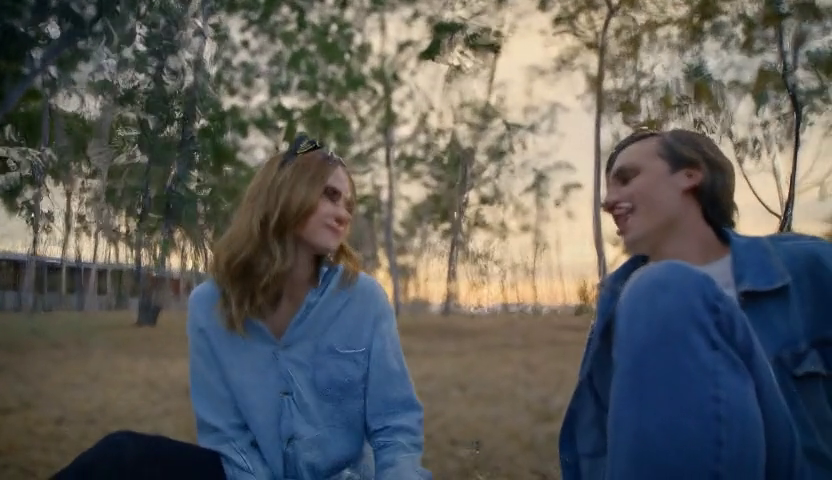} &
        \includegraphics[width=0.19\textwidth]{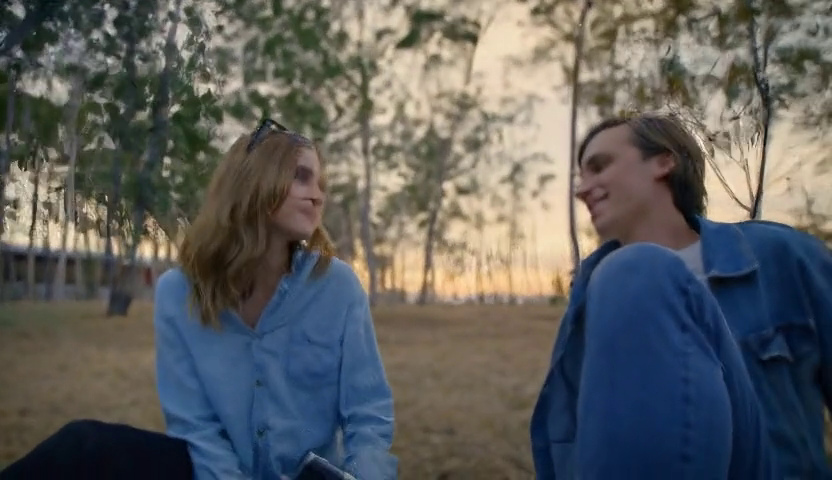} &
        \includegraphics[width=0.19\textwidth]{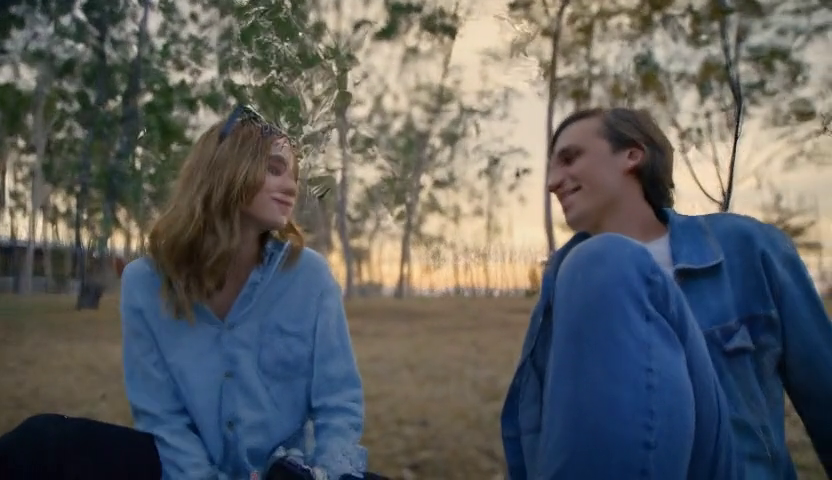} &
        \includegraphics[width=0.19\textwidth]{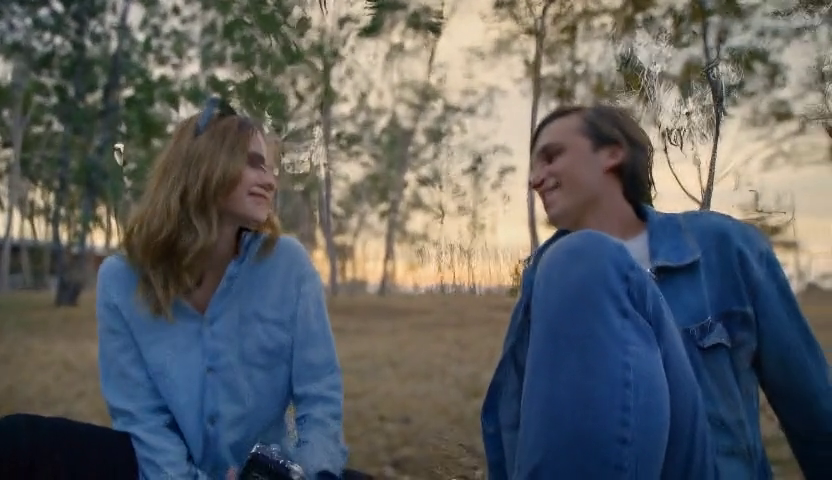} \\
        \rotatebox{90}{\,\,\,\,\,\,\,\,\small SVI~\cite{kwon2025solving}} &
        \includegraphics[width=0.19\textwidth]{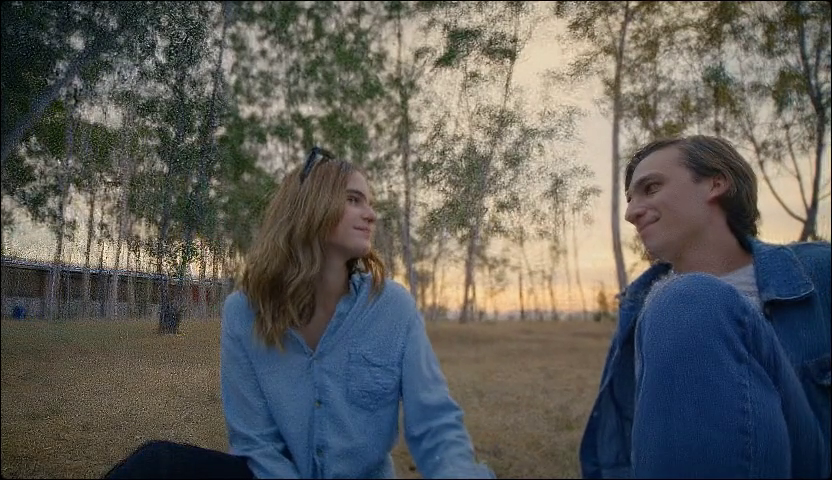} &
        \includegraphics[width=0.19\textwidth]{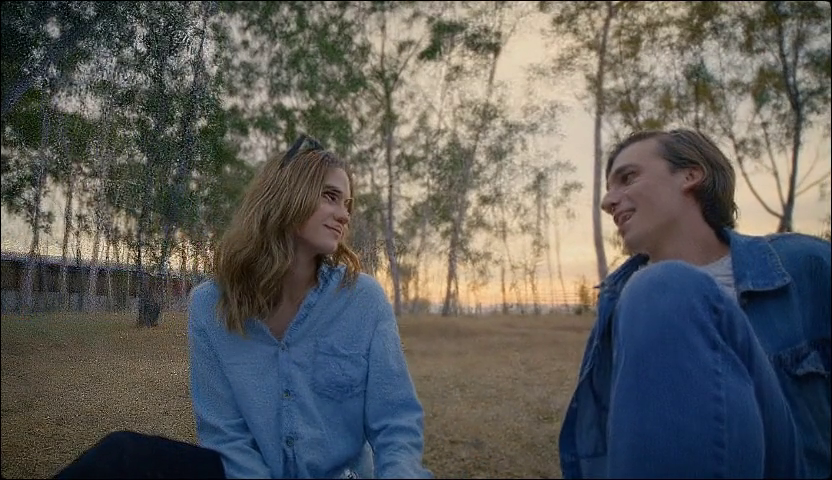} &
        \includegraphics[width=0.19\textwidth]{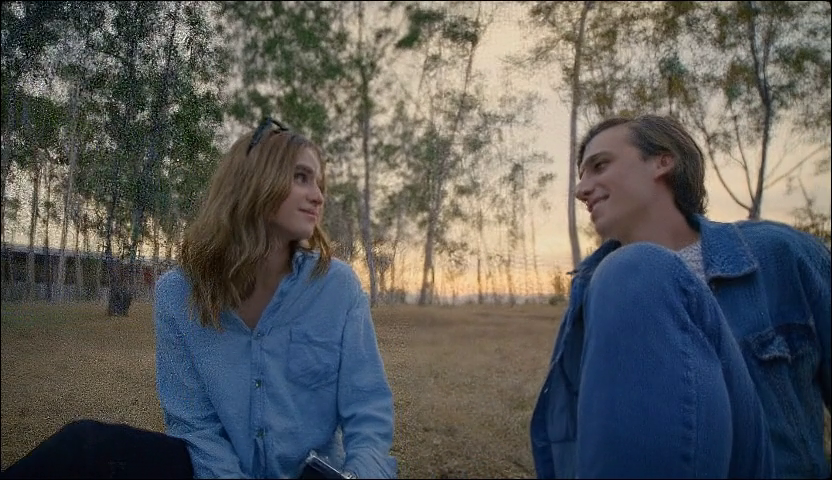} &
        \includegraphics[width=0.19\textwidth]{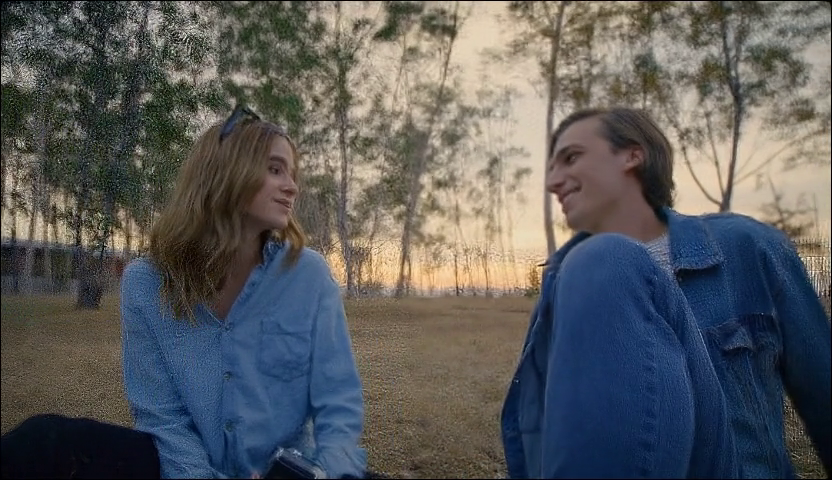} &
        \includegraphics[width=0.19\textwidth]{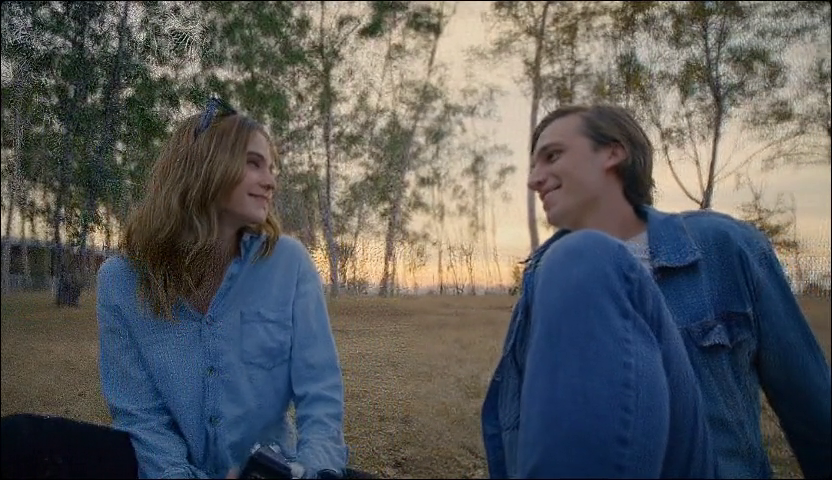} \\
        \rotatebox{90}{\small Vision-XL~\cite{kwon2024visionxlhighdefinitionvideo}} &
        \includegraphics[width=0.19\textwidth]{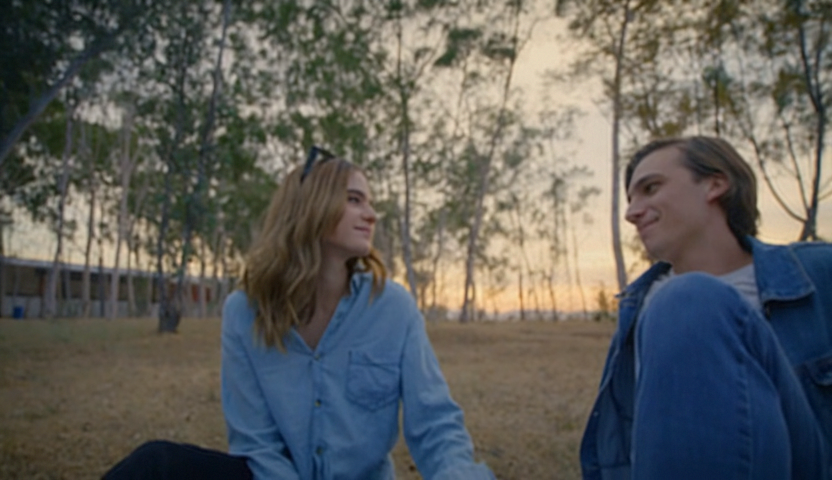} &
        \includegraphics[width=0.19\textwidth]{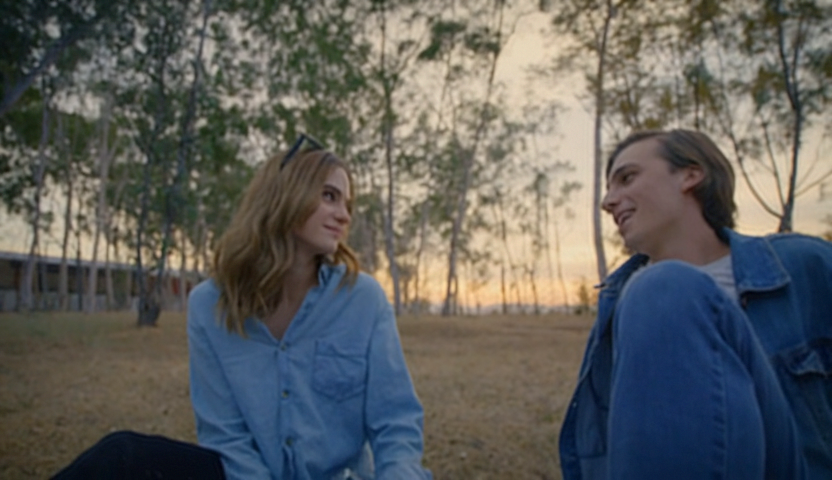} &
        \includegraphics[width=0.19\textwidth]{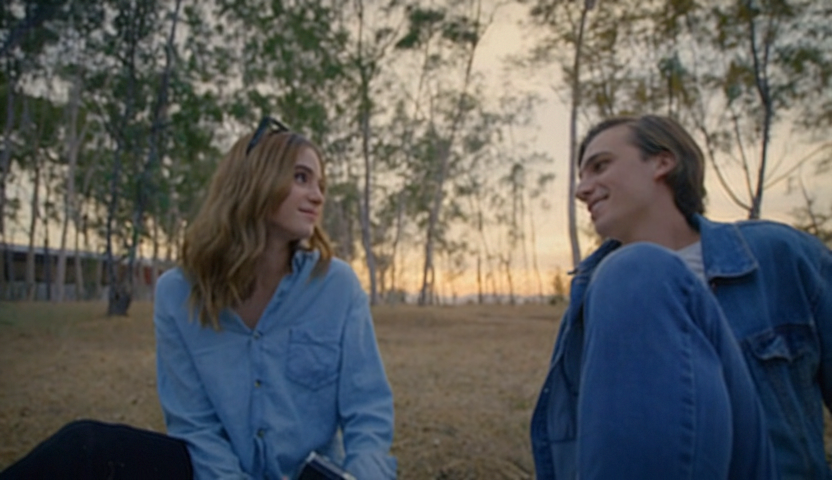} &
        \includegraphics[width=0.19\textwidth]{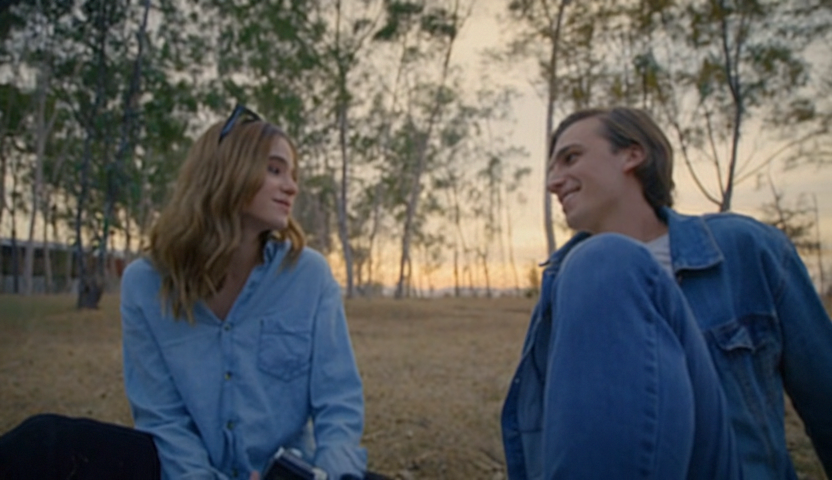} &
        \includegraphics[width=0.19\textwidth]{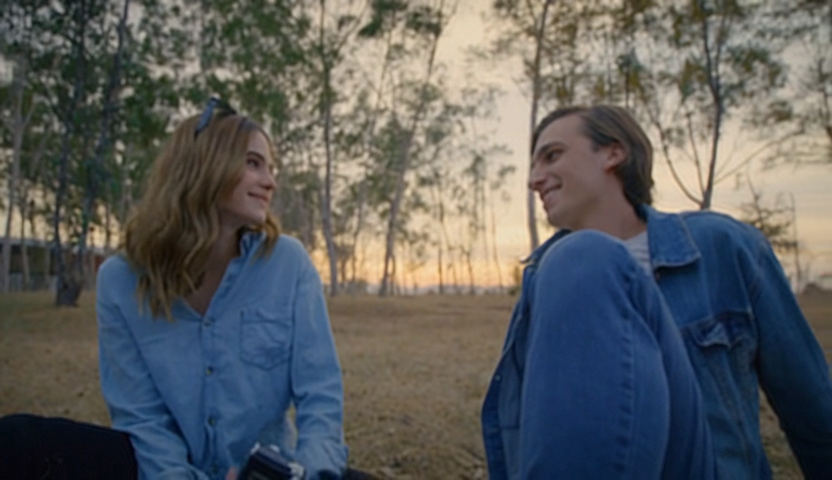} \\
        \rotatebox{90}{\,\,\,\,\,\,\,\,\,\,\small Ours} &
        \includegraphics[width=0.19\textwidth]{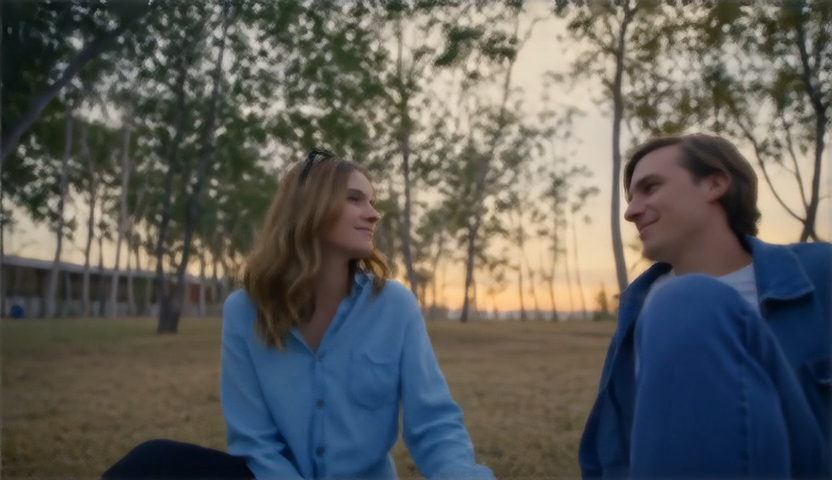} &
        \includegraphics[width=0.19\textwidth]{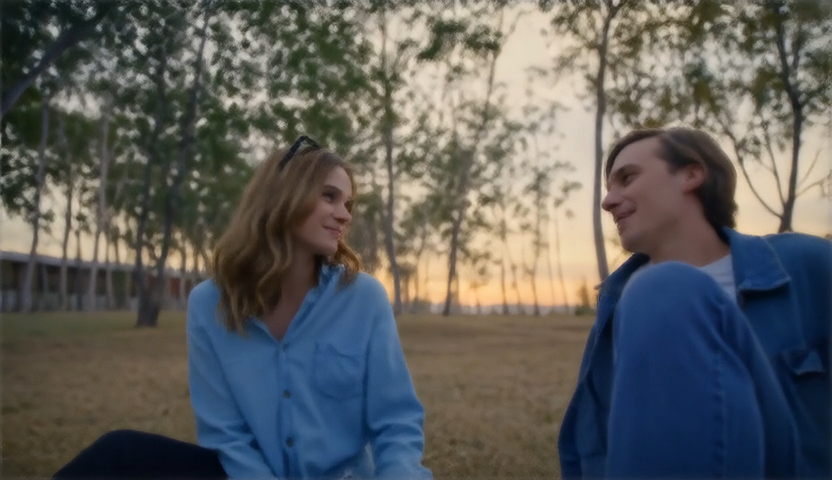} &
        \includegraphics[width=0.19\textwidth]{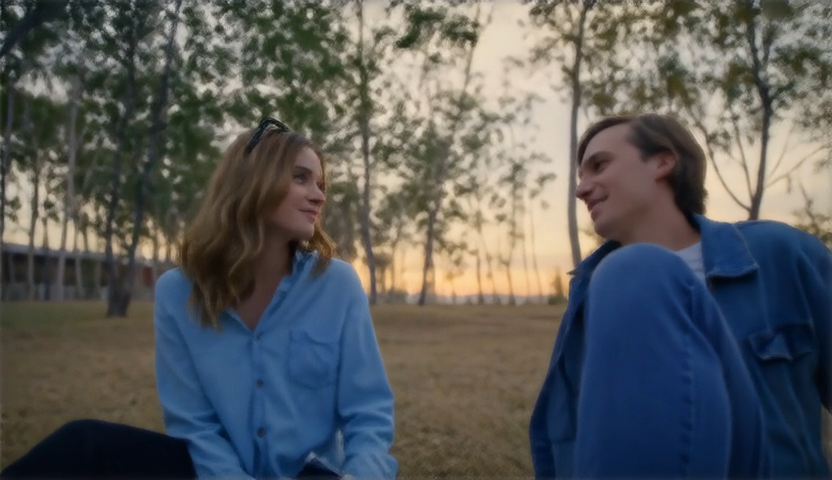} &
        \includegraphics[width=0.19\textwidth]{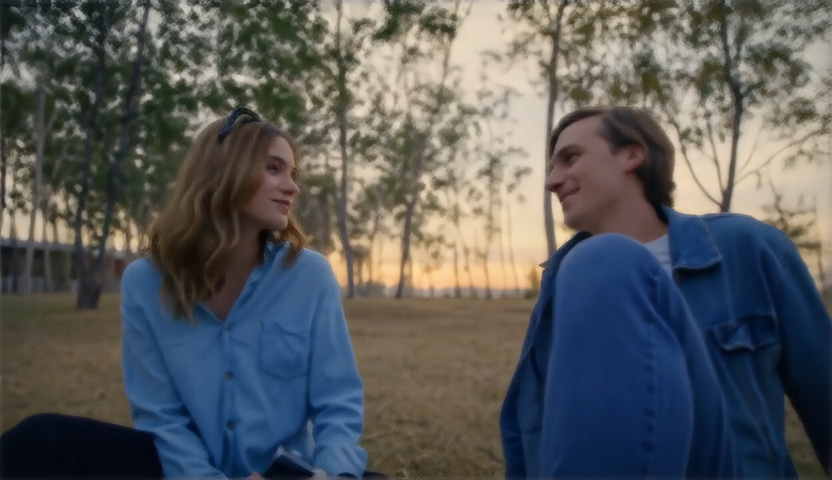} &
        \includegraphics[width=0.19\textwidth]{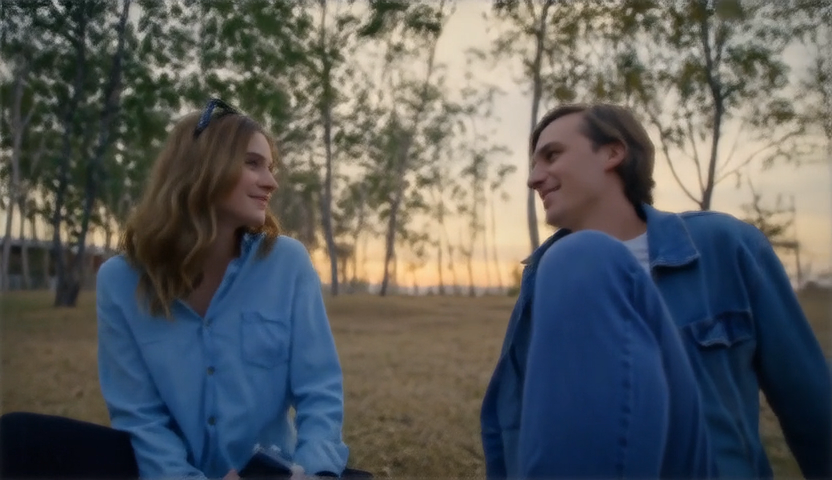} \\
        \rotatebox{90}{\,\,\,\,\,\,\,\,\,\,\small Ours$^\dag$} &
        \includegraphics[width=0.19\textwidth]{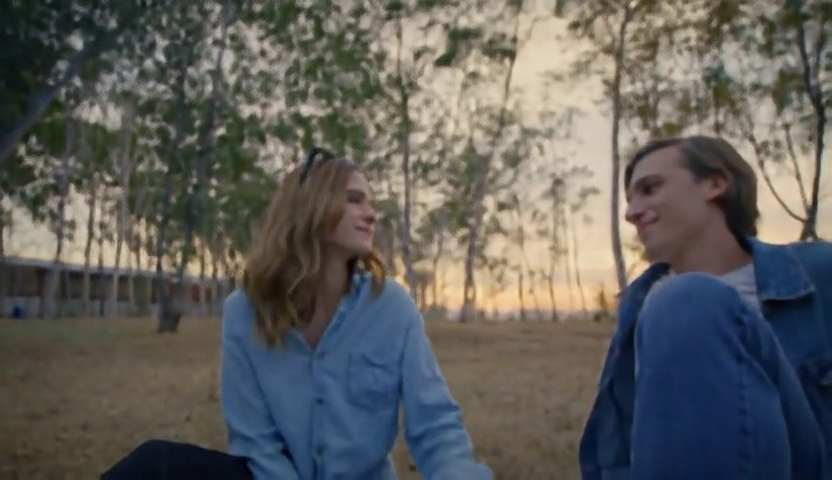} &
        \includegraphics[width=0.19\textwidth]{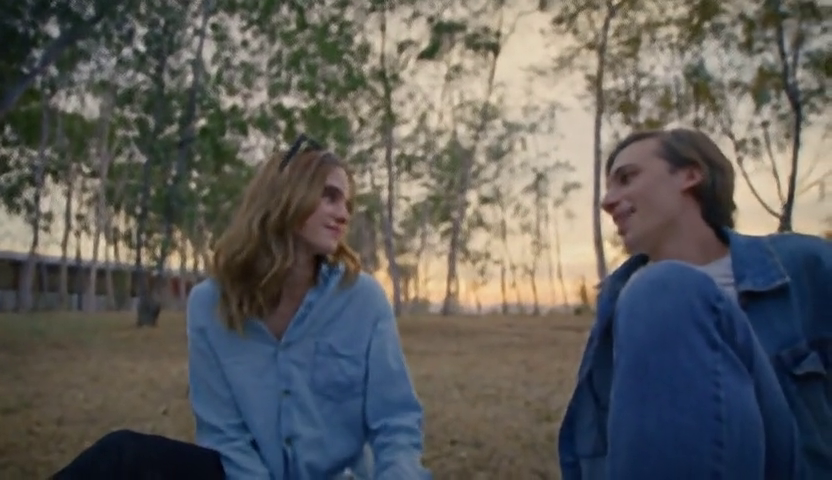} &
        \includegraphics[width=0.19\textwidth]{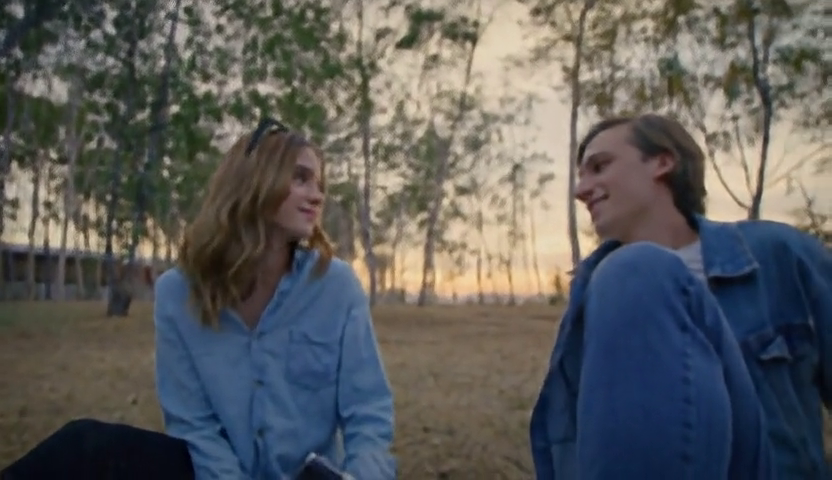} &
        \includegraphics[width=0.19\textwidth]{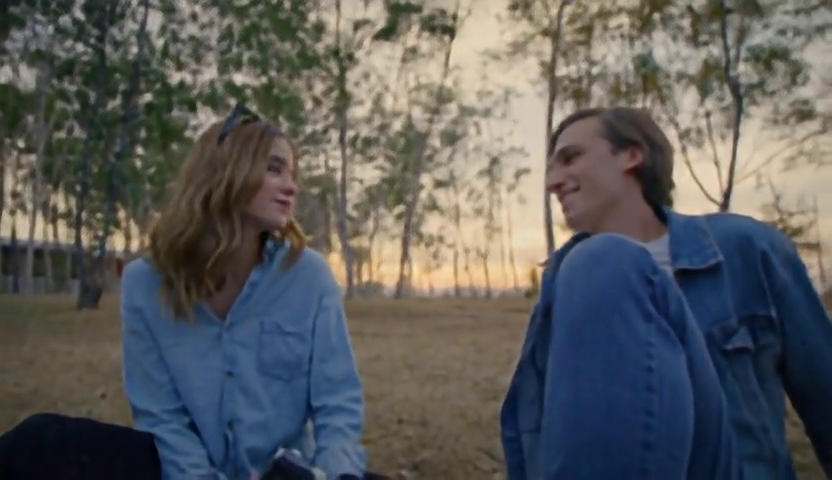} &
        \includegraphics[width=0.19\textwidth]{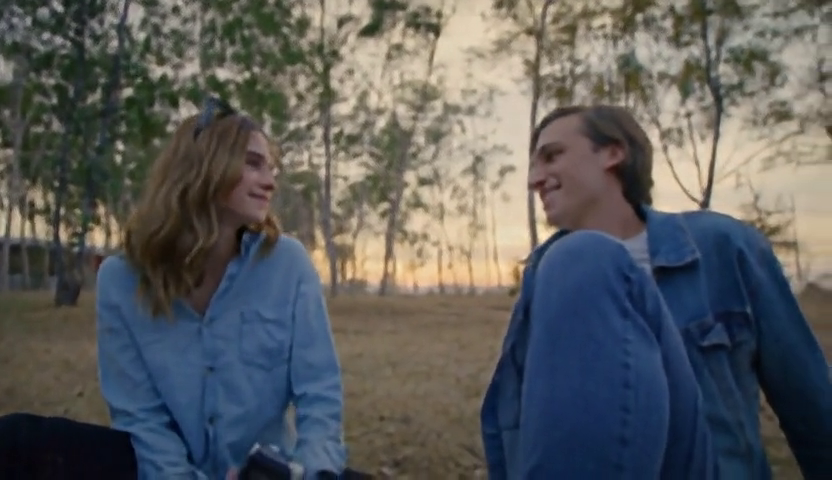} \\
        \rotatebox{90}{\,\,\,\,\,\,\,\,\,\,\small GT} &
        \includegraphics[width=0.19\textwidth]{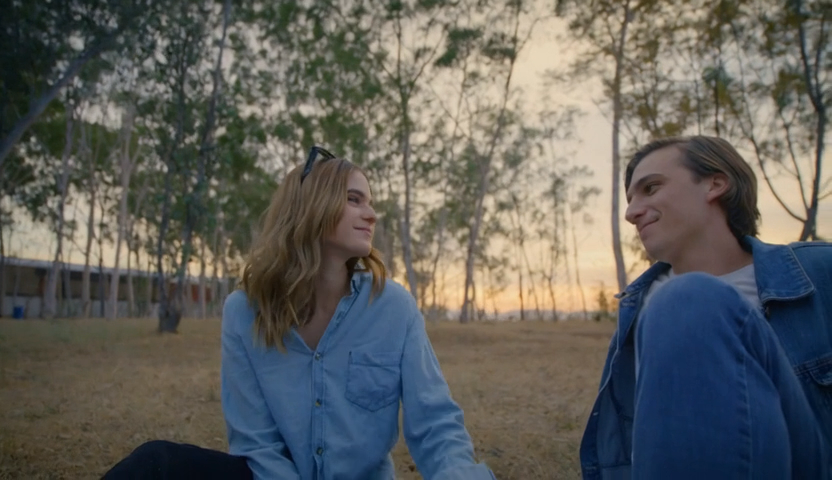} &
        \includegraphics[width=0.19\textwidth]{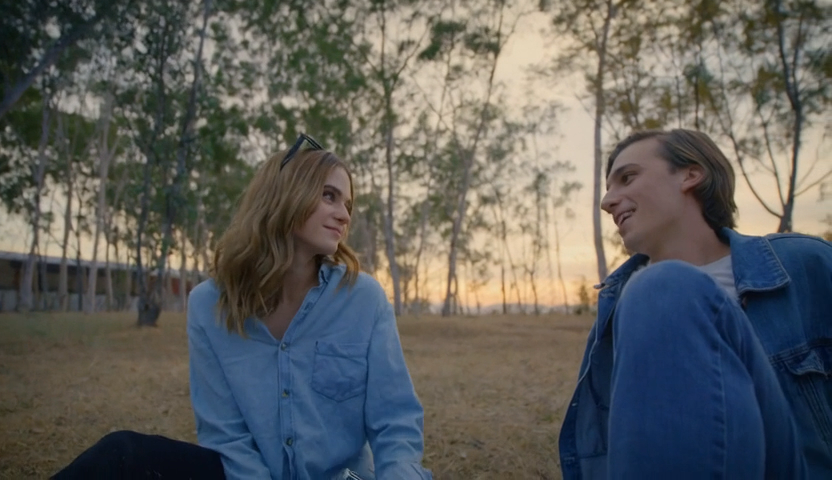} &
        \includegraphics[width=0.19\textwidth]{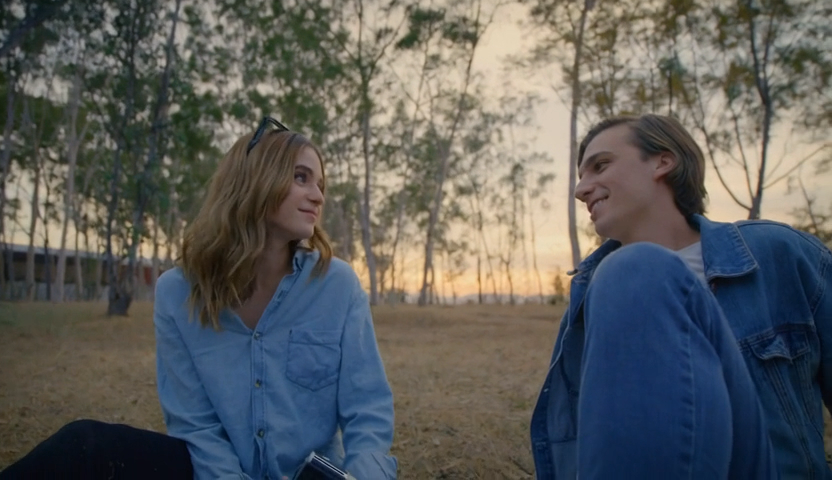} &
        \includegraphics[width=0.19\textwidth]{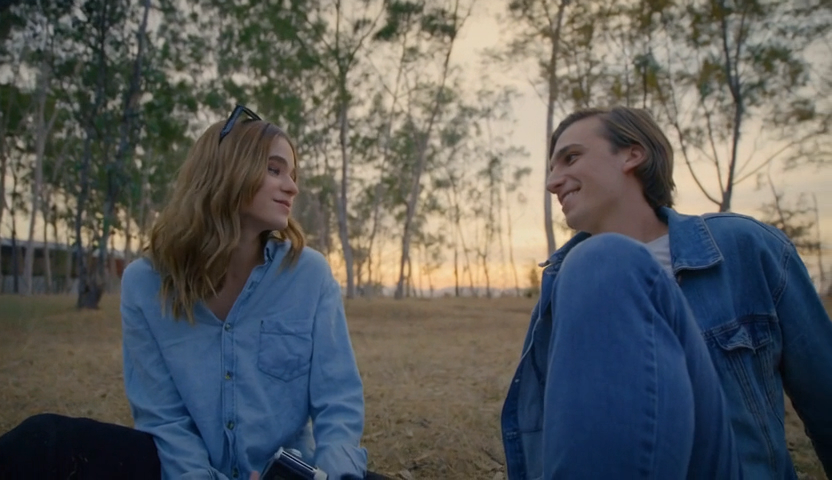} &
        \includegraphics[width=0.19\textwidth]{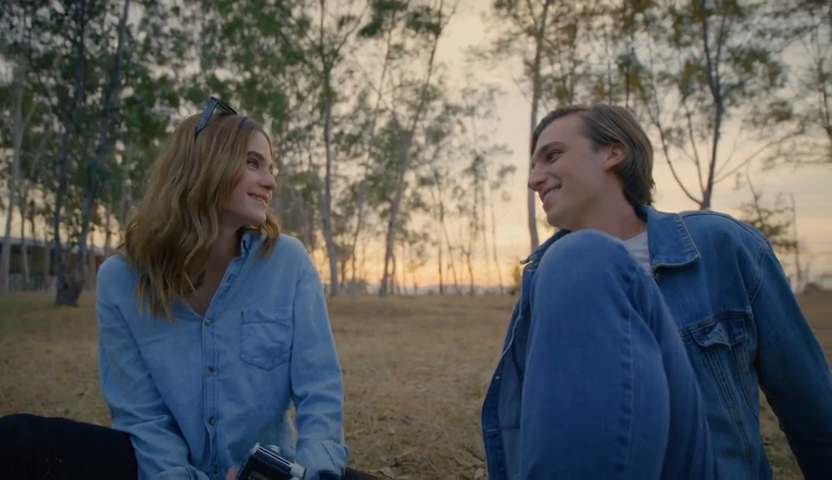} \\
    \end{tabular}
    }
    \caption{\textbf{Video Deblurring qualitative comparison.} Rows correspond to different methods; columns show consecutive frames covering the entire clip. InstantViR (both WanVAE \textbf{Ours} and LeanVAE \textbf{Ours$^\dag$} variants) restores fine structures consistently across time, outperforming slower diffusion-based baselines.}
    \label{fig:suppl_deblur}
\end{figure*}

\paragraph{Video Deblurring Comparison.} 
Figure~\ref{fig:suppl_deblur} presents comparisons on Gaussian deblurring. 
Our method effectively restores sharp details and outperforms optimization-based baselines, which often suffer from over-smoothing or severe temporal flickering. 
Notably, both our standard WanVAE-based model and the accelerated LeanVAE variant (Ours$^\dag$) maintain high reconstruction quality, generating sharp and stable results.

\begin{figure*}[h]
    \centering
    \setlength{\tabcolsep}{1pt} 
    \renewcommand{\arraystretch}{1.1}
    \resizebox{\linewidth}{!}{%
    \begin{tabular}{cccccc} 
        & \multicolumn{5}{c}{
        \begin{tikzpicture}[baseline]
            \draw[->, >=latex, line width=0.35mm] (0,0.1) -- (15.2cm,0.1) 
            node[right, xshift=2mm] at (15.2cm, 0.1) {Time};
        \end{tikzpicture}
      } \\
        \rotatebox{90}{\,\,\,\,\,\small LR Input} &
        \includegraphics[width=0.19\linewidth]{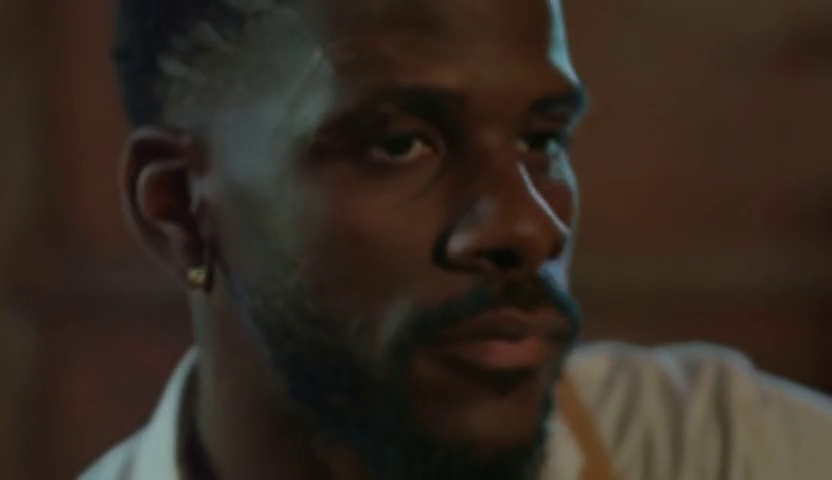} &
        \includegraphics[width=0.19\linewidth]{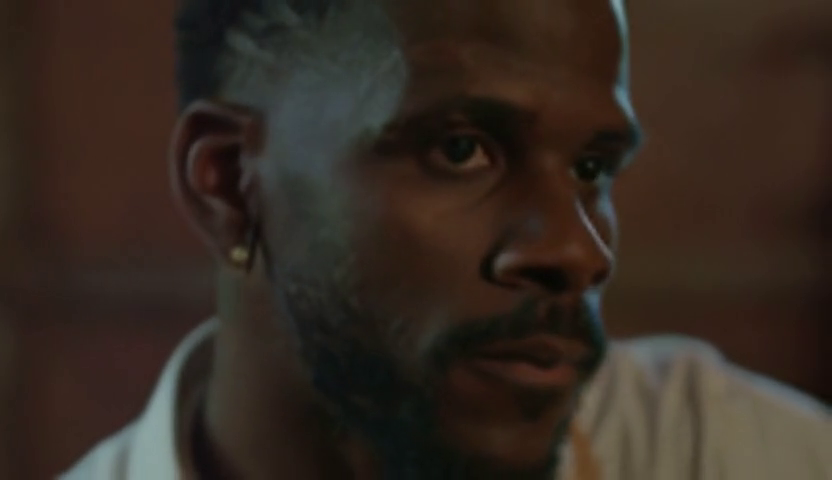} &
        \includegraphics[width=0.19\linewidth]{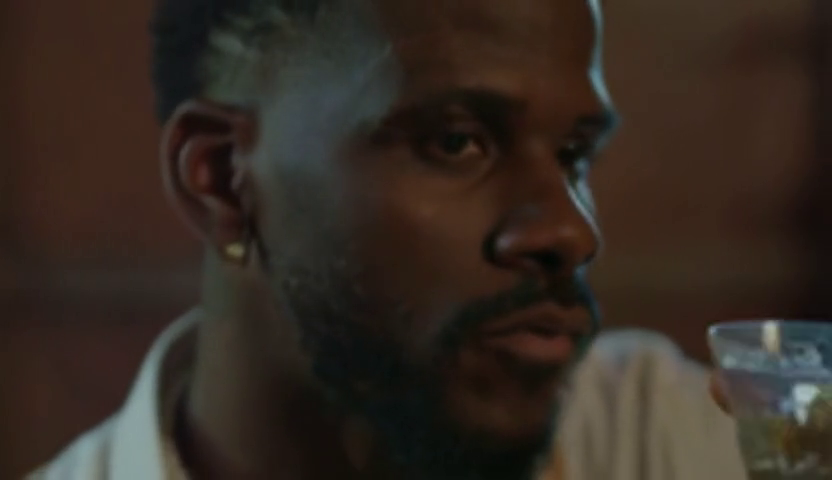} &
        \includegraphics[width=0.19\linewidth]{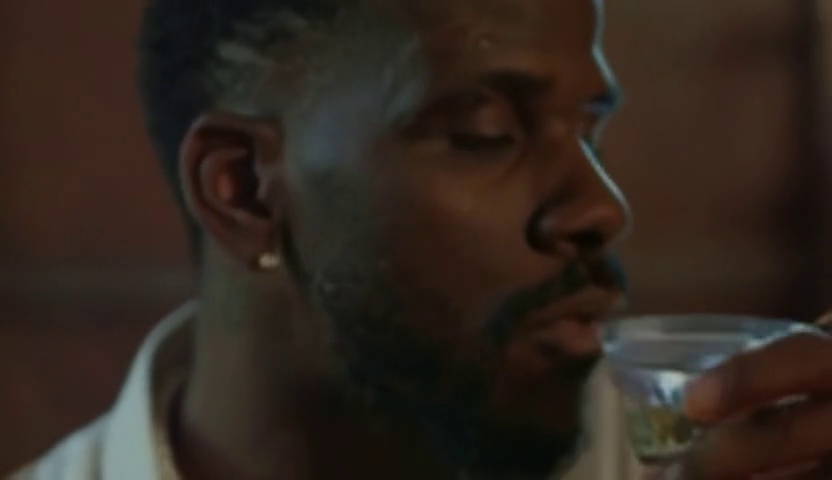} &
        \includegraphics[width=0.19\linewidth]{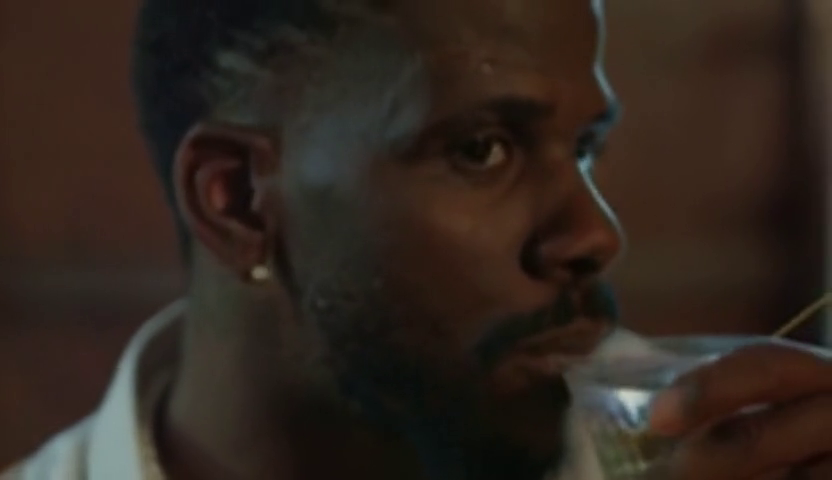} \\
        \rotatebox{90}{\,\,\,\,\,\,\,\,\small DPS~\cite{chung2022diffusion}} &
        \includegraphics[width=0.19\linewidth]{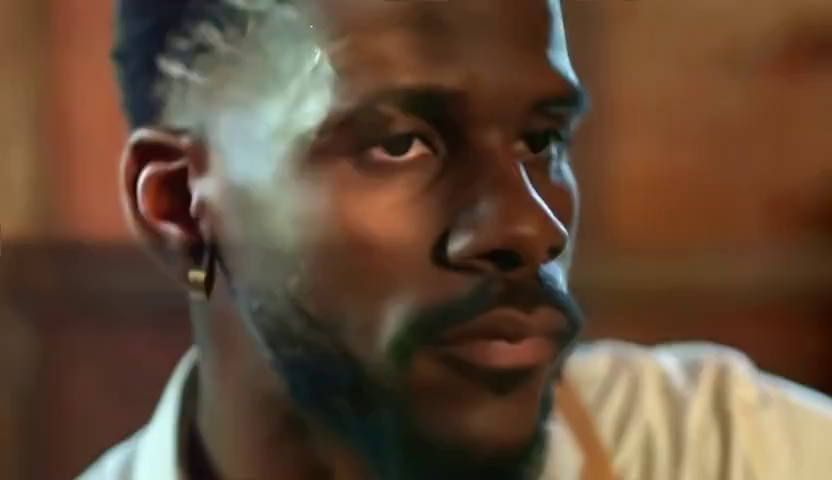} &
        \includegraphics[width=0.19\linewidth]{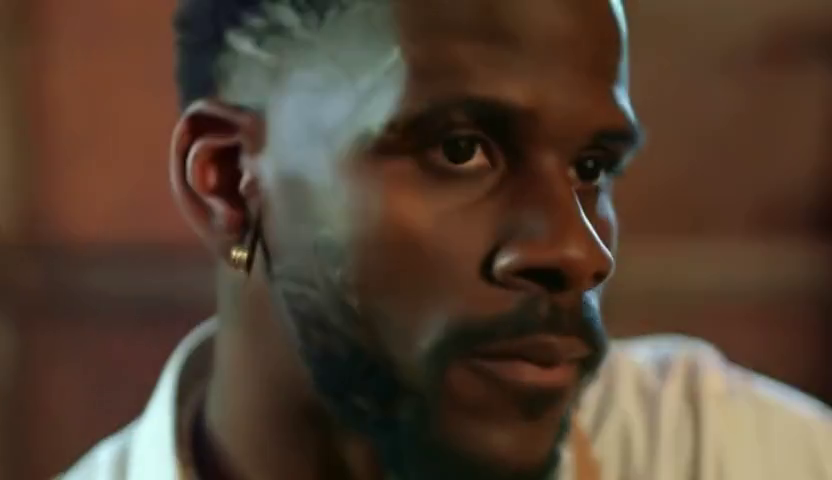} &
        \includegraphics[width=0.19\linewidth]{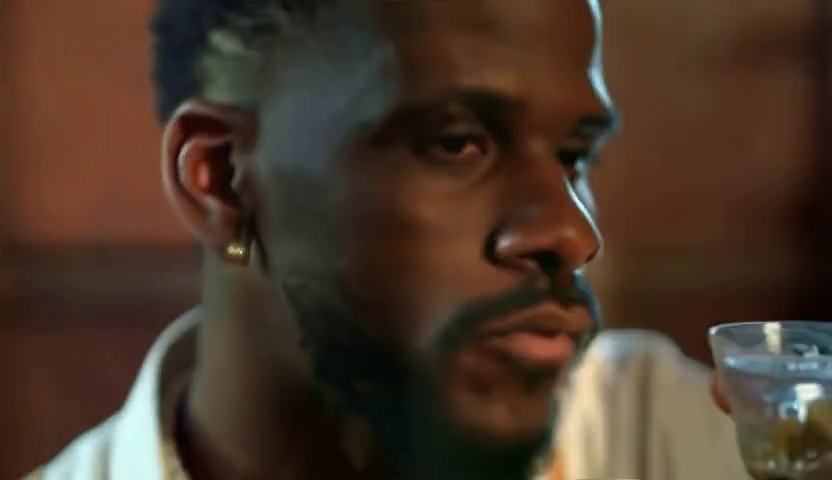} &
        \includegraphics[width=0.19\linewidth]{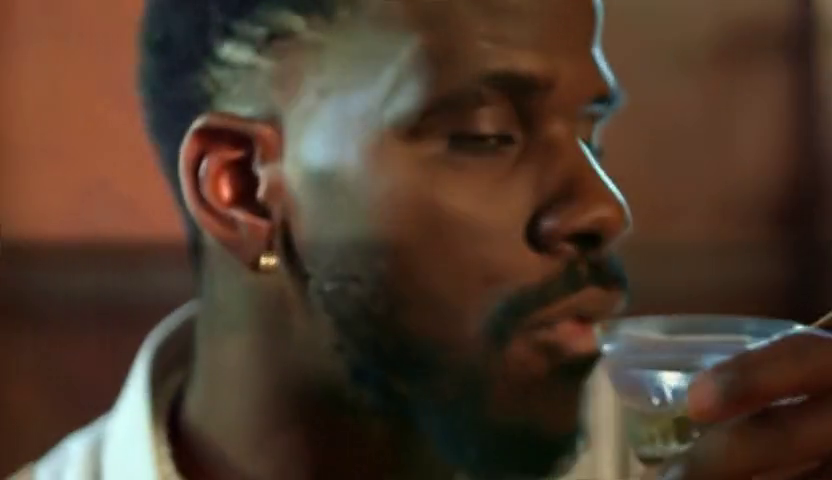} &
        \includegraphics[width=0.19\linewidth]{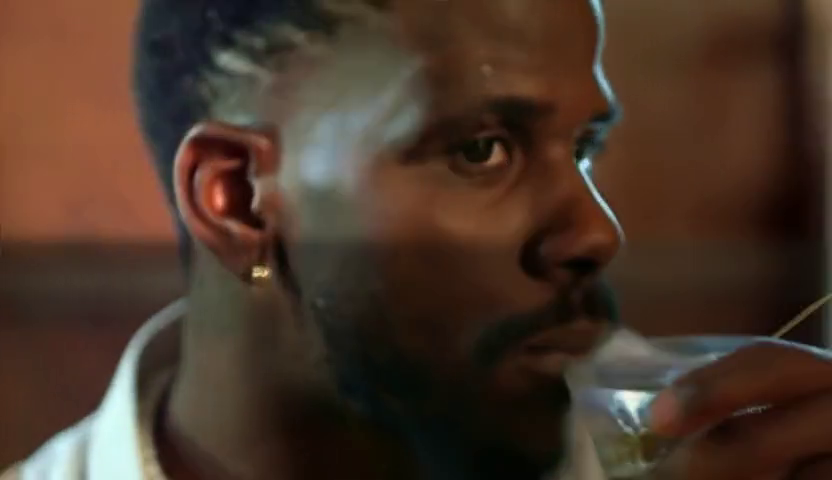} \\
        \rotatebox{90}{\,\,\,\,\,\,\,\,\small SVI~\cite{kwon2025solving}} &
        \includegraphics[width=0.19\linewidth]{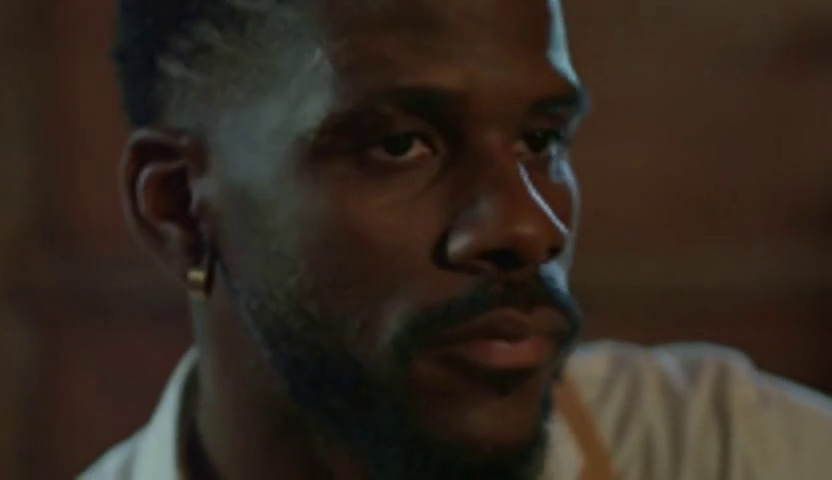} &
        \includegraphics[width=0.19\linewidth]{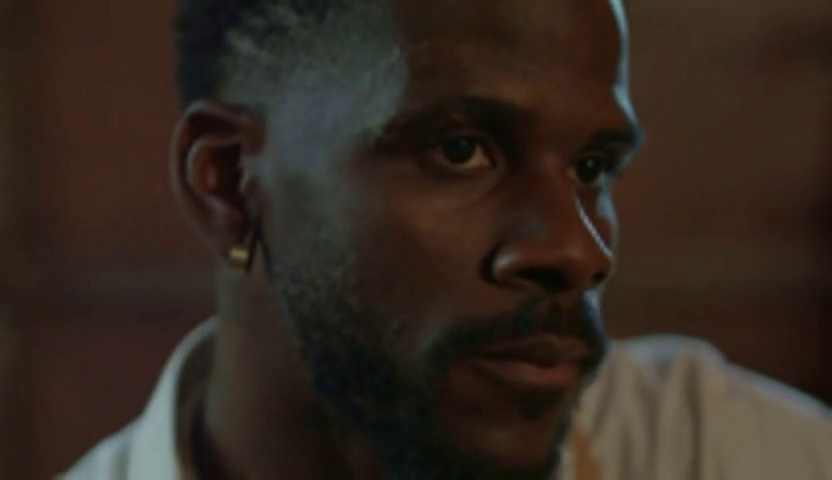} &
        \includegraphics[width=0.19\linewidth]{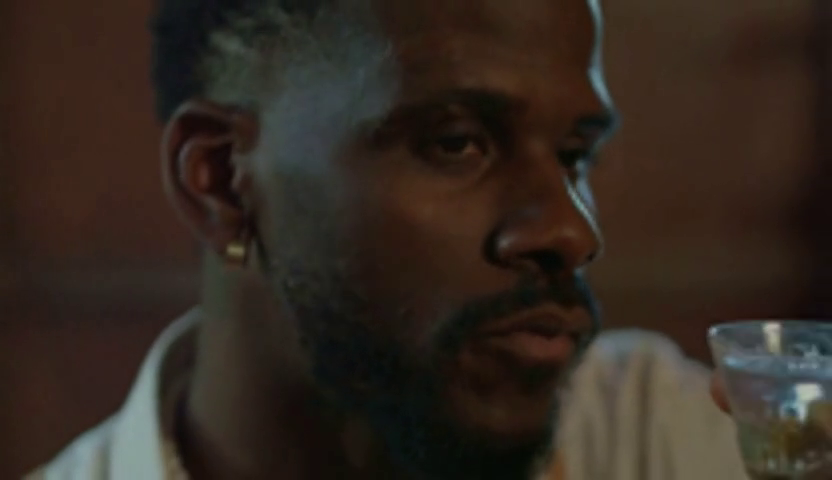} &
        \includegraphics[width=0.19\linewidth]{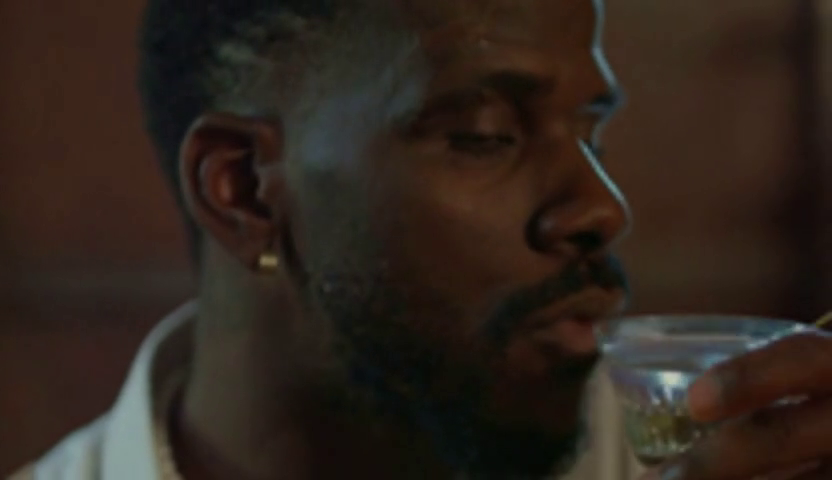} &
        \includegraphics[width=0.19\linewidth]{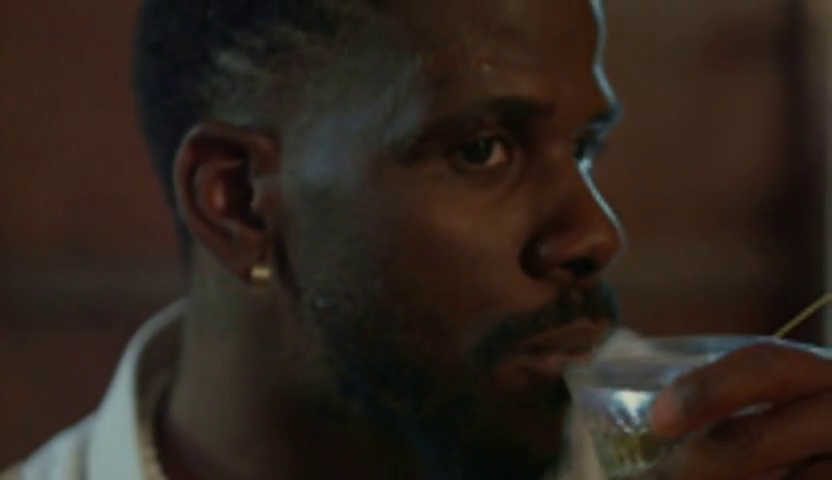} \\
        \rotatebox{90}{\small Vision-XL~\cite{kwon2024visionxlhighdefinitionvideo}} &
        \includegraphics[width=0.19\linewidth]{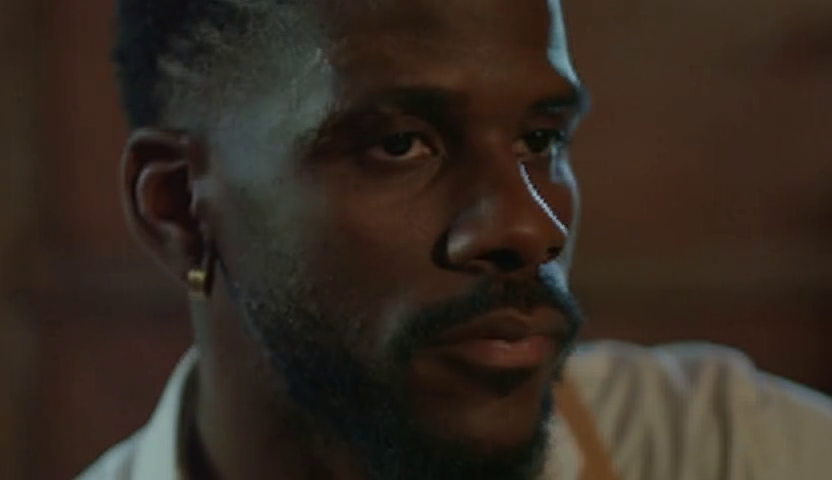} &
        \includegraphics[width=0.19\linewidth]{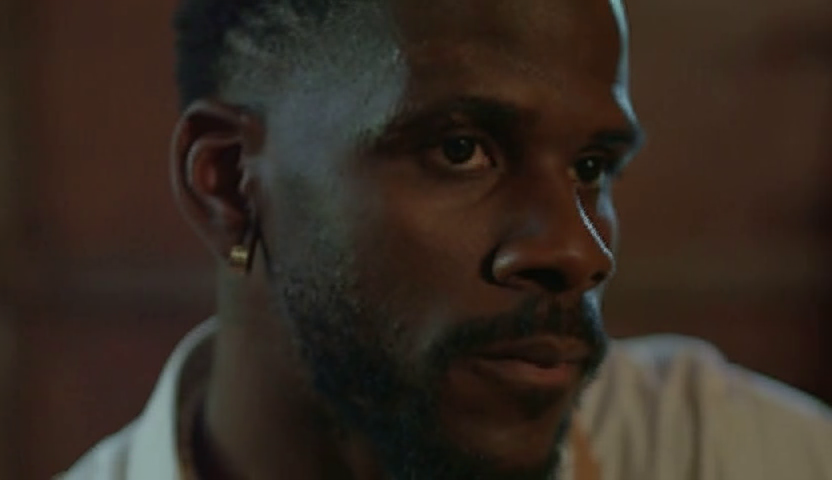} &
        \includegraphics[width=0.19\linewidth]{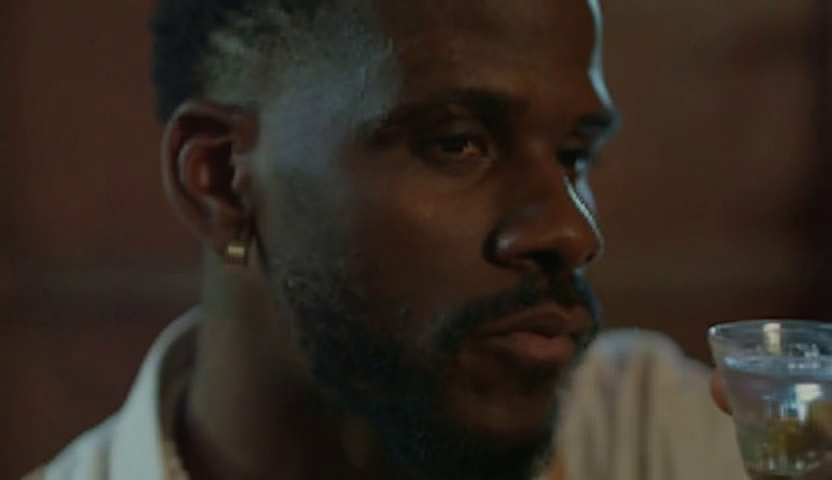} &
        \includegraphics[width=0.19\linewidth]{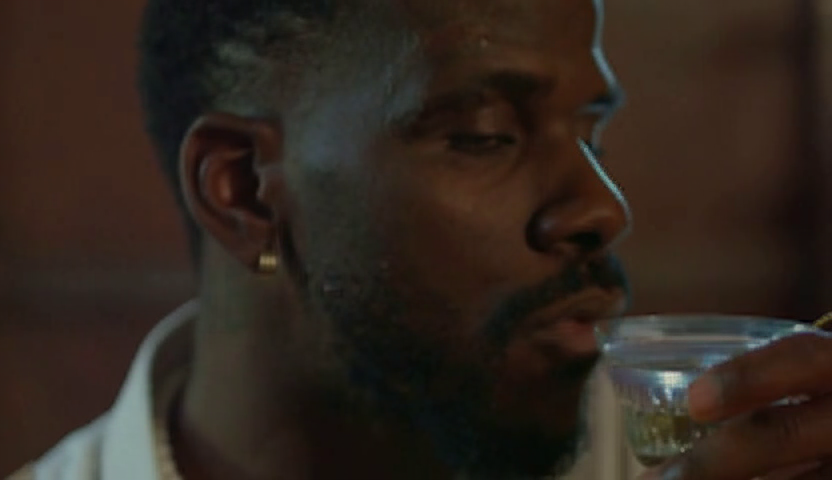} &
        \includegraphics[width=0.19\linewidth]{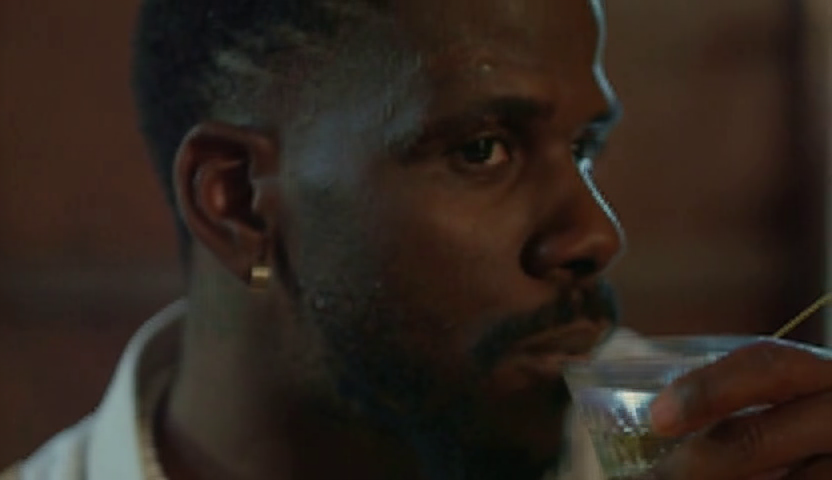} \\
        \rotatebox{90}{\,\,\,\,\,\,\,\,\,\,\small Ours} &
        \includegraphics[width=0.19\linewidth]{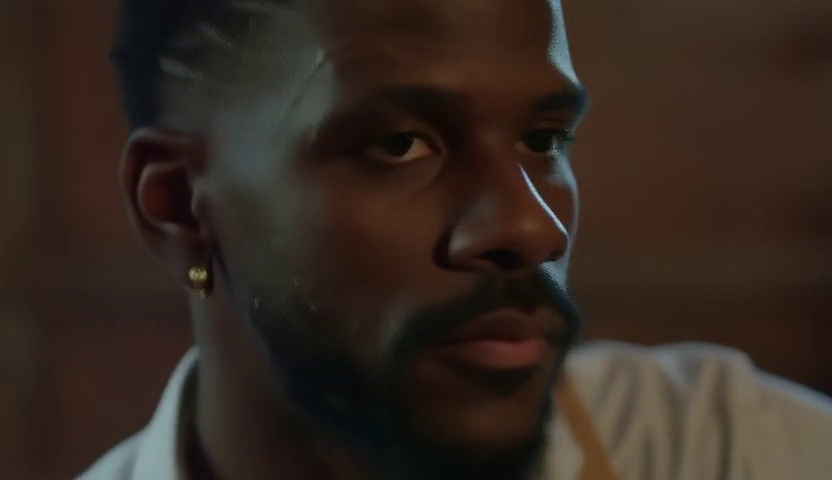} &
        \includegraphics[width=0.19\linewidth]{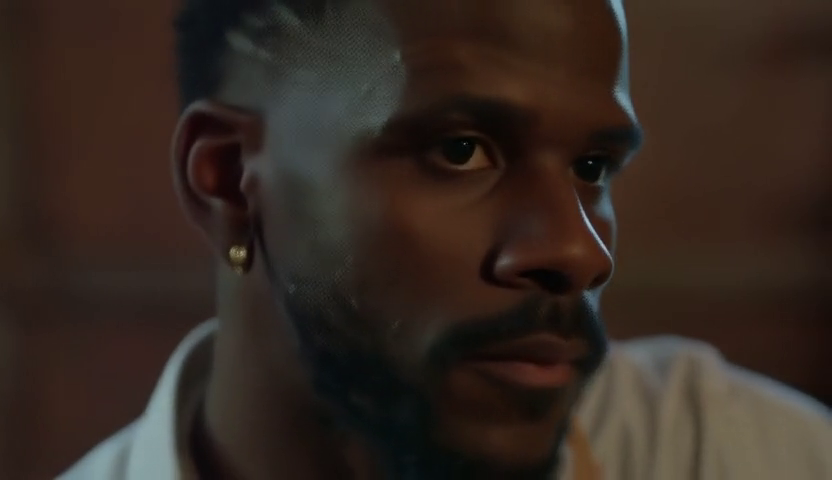} &
        \includegraphics[width=0.19\linewidth]{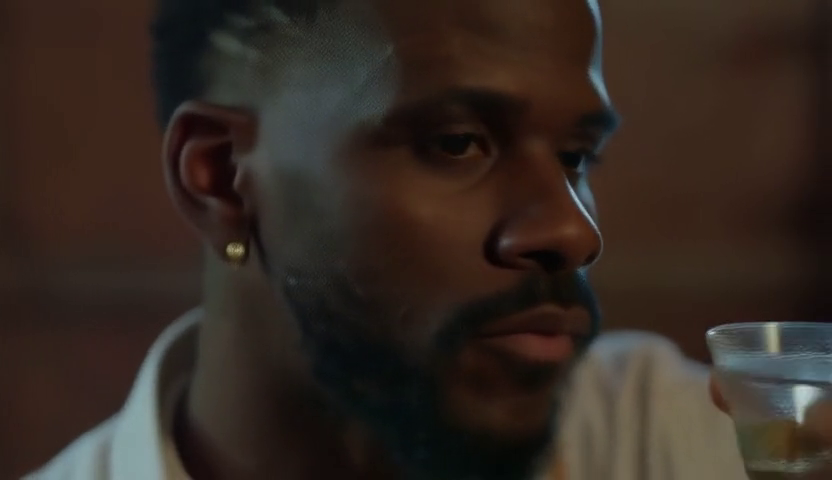} &
        \includegraphics[width=0.19\linewidth]{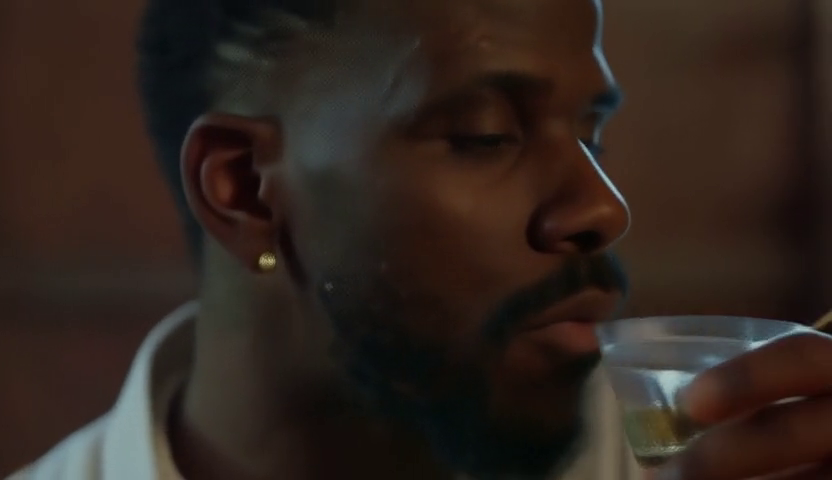} &
        \includegraphics[width=0.19\linewidth]{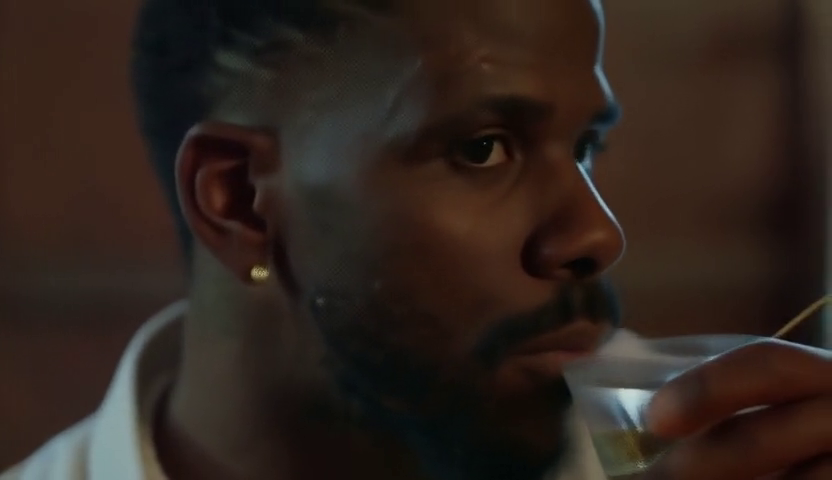} \\
        \rotatebox{90}{\,\,\,\,\,\,\,\,\,\,\small GT} &
        \includegraphics[width=0.19\linewidth]{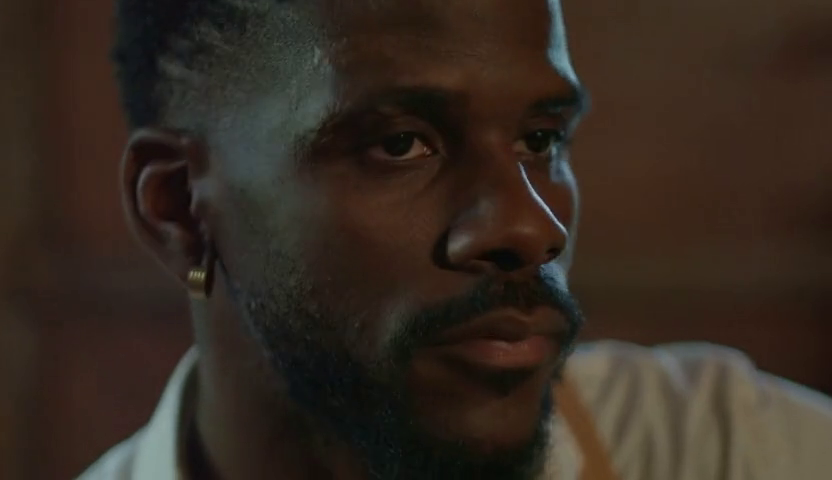} &
        \includegraphics[width=0.19\linewidth]{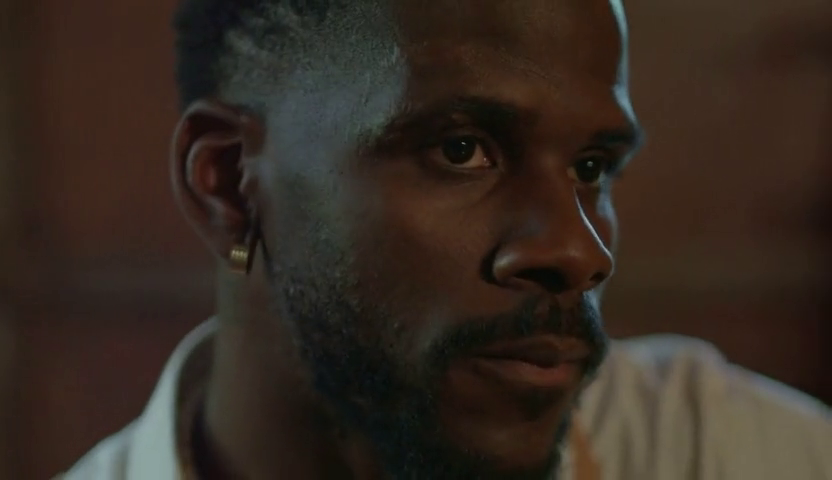} &
        \includegraphics[width=0.19\linewidth]{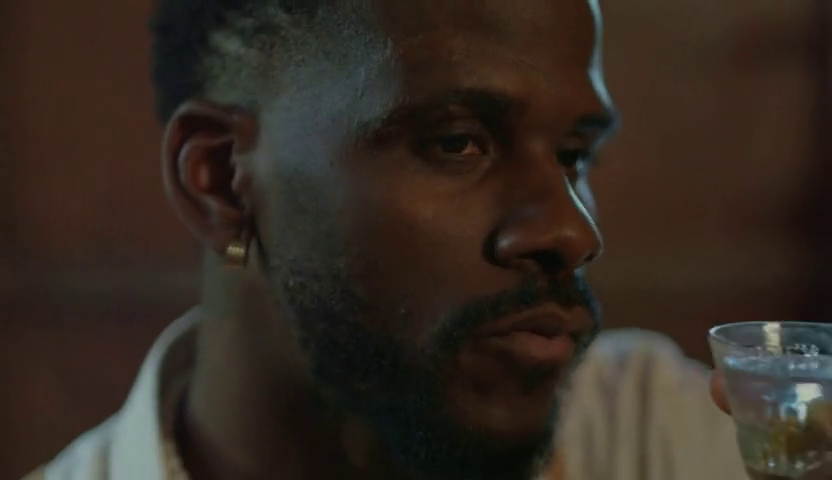} &
        \includegraphics[width=0.19\linewidth]{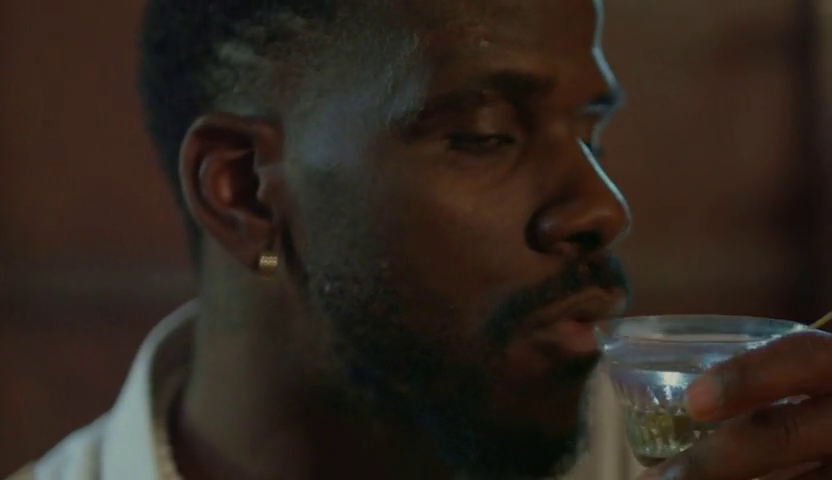} &
        \includegraphics[width=0.19\linewidth]{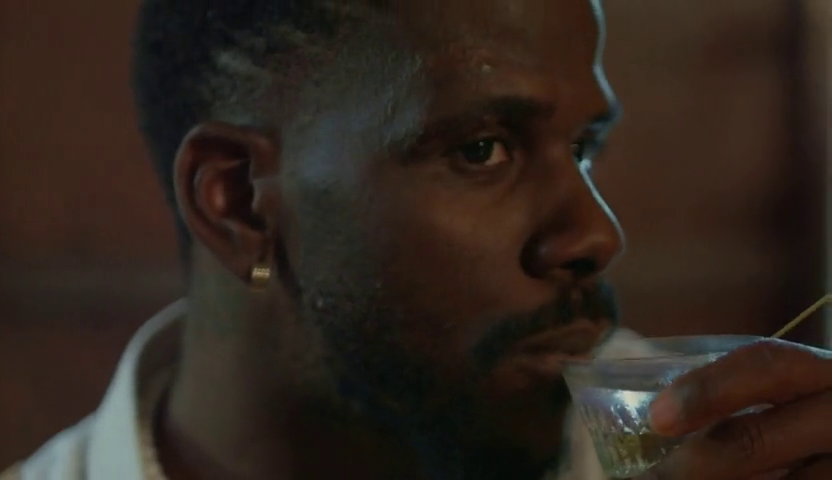} \\
    \end{tabular}%
    } 
    \caption{\textbf{Video Super-Resolution (4$\times$) qualitative comparison.} InstantViR restores temporally consistent structures, outperforming slower diffusion-based baselines in both sharpness and coherence.}
    \label{fig:suppl_sr}
\end{figure*}

\paragraph{Video Super-Resolution Comparison.}
Figure~\ref{fig:suppl_sr} showcases results for 4$\times$ video super-resolution.
While baselines like SVI and Vision-XL provide reasonable stability, they often lack high-frequency texture details. 
InstantViR successfully recovers fine structures and textures, producing a high-resolution output that closely matches the ground truth, demonstrating the power of our distilled video prior.

\begin{figure*}[h]
    \centering
    \setlength{\tabcolsep}{1pt}
    \begin{tabular}{cccc}
        \includegraphics[width=0.24\textwidth]{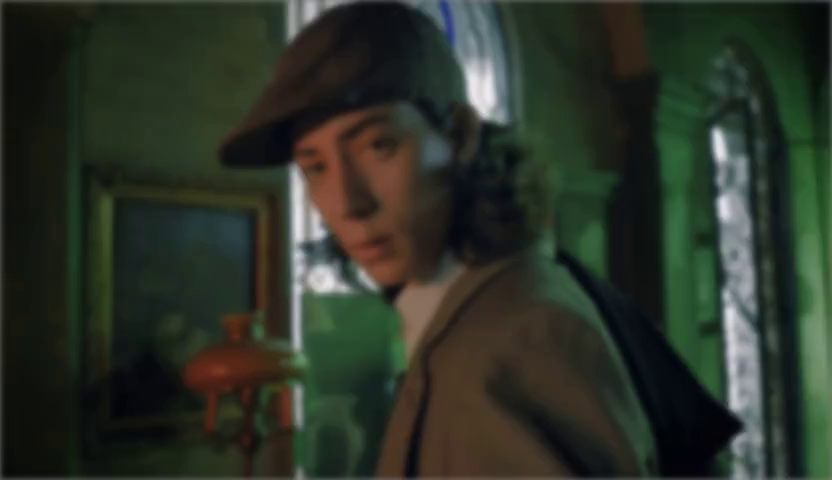} &
        \includegraphics[width=0.24\textwidth]{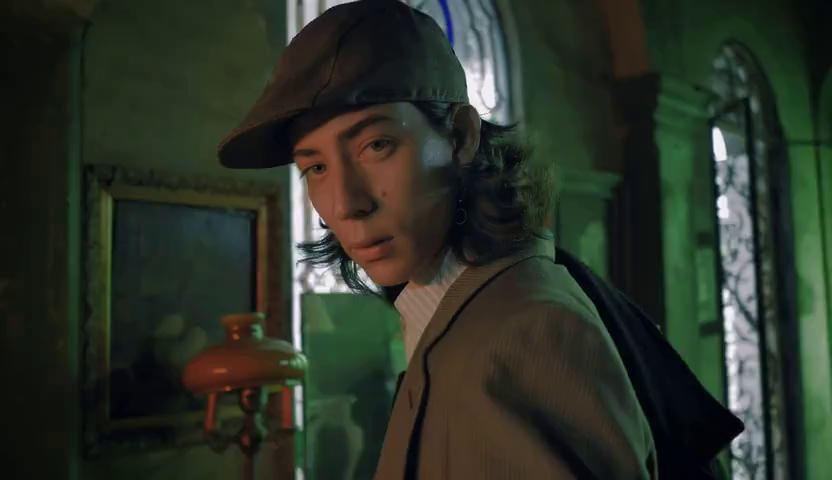} &
        \includegraphics[width=0.24\textwidth]{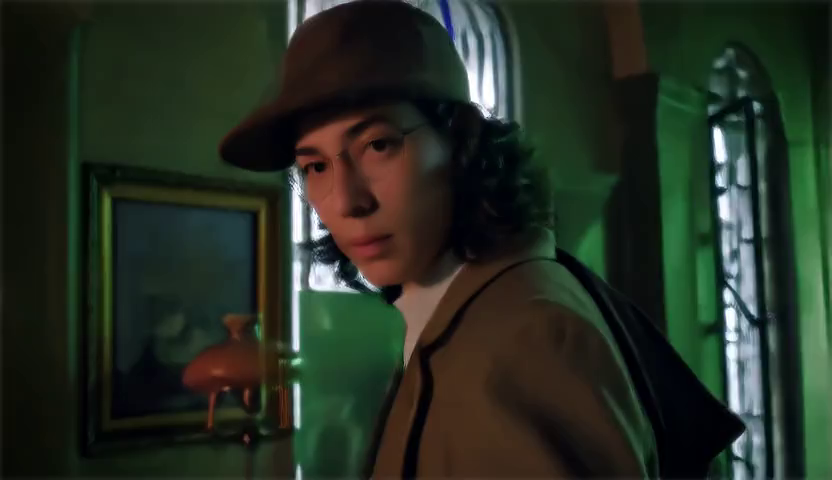} &
        \includegraphics[width=0.24\textwidth]{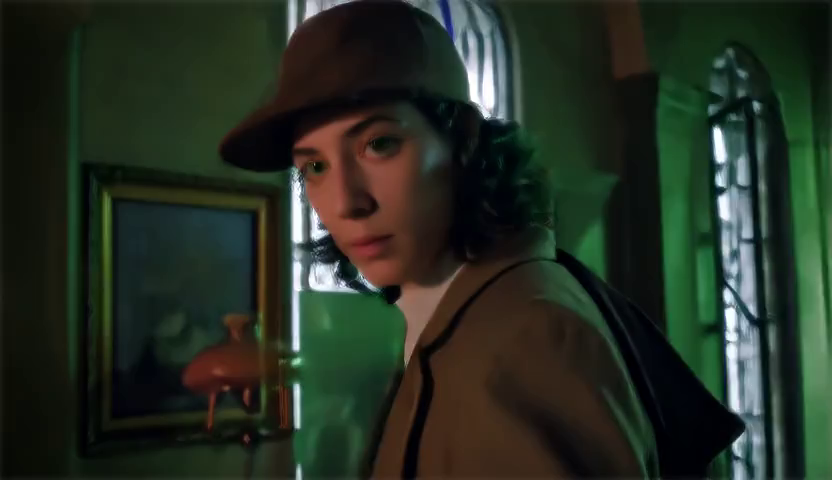} \\
        \small Blurred Input & \small Ground Truth & \small Prompt: ``glasses'' & \small Prompt: ``green eyes'' \\
    \end{tabular}
    \caption{\textbf{Text-Guided Video Deblurring.} InstantViR can generate diverse outcomes from the same blurred input based on text prompts, adding specific features like glasses or changing eye color while maintaining temporal coherence.}
    \label{fig:suppl_text_reg}
\end{figure*}

\paragraph{Text-Guided Deblurring.}
Figure~\ref{fig:suppl_text_reg} demonstrates the controllability of InstantViR. Given the same blurred measurement, our model can restore details according to different text prompts (e.g., adding ``glasses'' or changing eye color to ``green eyes''), generating semantically consistent and realistic content that aligns with the user's intent.

\end{document}